%% file: main.tex
\documentclass{article}

\usepackage{microtype}
\usepackage{graphicx}
\usepackage{booktabs} % for professional tables
\usepackage{multirow} % for tables with merged rows
\usepackage{comment}
\usepackage{subcaption}   % Using subcaption instead of subfigure
\usepackage{enumitem}

\usepackage{hyperref}

\usepackage[accepted]{icml2025}

\usepackage{amsmath}
\usepackage{amssymb}
\usepackage{mathtools}
\usepackage{amsthm}

\usepackage{tcolorbox}
\tcbuselibrary{listings, skins, breakable}
\usepackage{listings}

\newtcblisting{pseudocode}{
  colback=gray!5!white,
  colframe=gray!40!black,
  listing only,
  listing options={
    basicstyle=\ttfamily\footnotesize,
    breaklines=true,
    showstringspaces=false,
    columns=fullflexible
  },
  left=2mm, right=2mm, top=1mm, bottom=1mm,
  enhanced,
  sharp corners,
  boxrule=0.3pt,
  width=\columnwidth,      % ensures it fits in one column
  breakable
}

\theoremstyle{plain}
\newtheorem{theorem}{Theorem}[section]
\newtheorem{proposition}[theorem]{Proposition}
\newtheorem{lemma}[theorem]{Lemma}
\newtheorem{corollary}[theorem]{Corollary}
\theoremstyle{definition}
\newtheorem{definition}[theorem]{Definition}

\theoremstyle{remark}

\usepackage{tikz}
\usetikzlibrary{shapes.geometric, arrows}

\tikzstyle{startstop} = [rectangle, rounded corners, minimum width=3cm, minimum height=1cm,text centered, draw=black, fill=red!30]
\tikzstyle{process} = [rectangle, minimum width=3cm, minimum height=1cm, text centered, draw=black, fill=blue!30]
\tikzstyle{arrow} = [thick,->,>=stealth]
\tikzstyle{data} = [ellipse, minimum width=3cm, minimum height=1cm, text centered, draw=black, fill=green!30]
\tikzstyle{decision} = [diamond, minimum width=3cm, minimum height=1cm, text centered, draw=black, fill=yellow!30]

\input{macros}

\newif\ifshowcomments
\showcommentstrue  % Change to \showcommentsfalse to hide all comments

\usepackage[capitalize,noabbrev]{cleveref}

\icmltitlerunning{\STAMP: Inference Verification Despite Nondeterminism}

\begin{document}

\twocolumn[
\icmltitle{\STAMP: Inference Verification Despite Nondeterminism}


\icmlsetsymbol{equal}{*}

\begin{icmlauthorlist}
\icmlauthor{Adam Karvonen}{equal,mats}
\icmlauthor{Daniel Reuter}{equal,mats}
\icmlauthor{Roy Rinberg}{mats,harvard}
\icmlauthor{Luke Marks}{mats}
\icmlauthor{Adri\`a Garriga-Alonso}{far}
\icmlauthor{Keri Warr}{ant}
\end{icmlauthorlist}

\icmlaffiliation{far}{FAR AI}
\icmlaffiliation{mats}{ML 
Alignment and Theory Scholars (MATS)}
\icmlaffiliation{harvard}{Harvard University}
\icmlaffiliation{ant}{Anthropic}

\icmlcorrespondingauthor{Adam Karvonen}{adam.karvonen@gmail.com}

\icmlkeywords{Machine Learning, Verifiable Inference, LLM}

\vskip 0.3in
]

% \printAffiliationsAndNotice{ }  % commented out due to ICML bug with empty argument
\printAffiliationsAndNotice{\icmlEqualContribution} % otherwise use the standard text.
\input{sections/00-abstract}

\input{sections/01-intro}
\input{sections/02-related-work}
\input{sections/03-preliminaries}

\input{sections/04-theory}

\input{sections/05-experimental-design}
\input{sections/06-results}
\input{sections/07-discussion}
\input{sections/conclusions}
\newpage
\input{sections/ack}

\bibliography{citations}
\bibliographystyle{icml2025}

\appendix

\input{sections/appendix}
\input{sections/appendix-additional-results}

\end{document}

%% file: macros.tex
\newcommand{\ex}[2]{{\ifx&#1& \mathbb{E} \else \underset{#1}{\mathbb{E}} \fi \left[#2\right]}}
\newcommand{\pr}[2]{{\ifx&#1& \mathbb{P} \else \underset{#1}{\mathbb{P}} \fi \left[#2\right]}}
\newcommand{\var}[2]{{\ifx&#1& \mathrm{Var} \else \underset{#1}{\mathrm{Var}} \fi \left(#2\right)}}

\renewcommand{\epsilon}{\varepsilon}  % Epsilon redefined
\newcommand{\DiFR}{\textsc{D}i\textsc{FR}}  % DiFR main name
\newcommand{\STAMP}{\DiFR}  % STAMP algorithm (alias for DiFR)

\newtoggle{showcomments}
\togglefalse{showcomments} % to hide comments
\newcommand{\updated}[1]{\iftoggle{showcomments}{\textcolor{blue}{#1}}{#1}}

%% file: sections/00-abstract.tex
\begin{abstract}

As demand for LLM inference grows, it is becoming increasingly important that providers and their customers can verify that inference processes are performed correctly, without errors or tampering. However, re-running the same inference process twice often leads to different results due to benign numerical noise, making it difficult to distinguish legitimate variation from actual problems.
To address this problem, we introduce Token-DiFR (\textbf{Token}-\textbf{Di}vergence-\textbf{F}rom-\textbf{R}eference), a method for verifying inference outputs by comparing generated tokens against predictions made by a trusted reference implementation conditioned on the same random seed. 
Sampling seed synchronization tightly constrains valid outputs, leaving providers minimal room to deviate from correct inference, which allows output tokens themselves to serve as auditable evidence of correctness at zero additional cost to the provider. 
Token-DiFR reliably identifies sampling errors, simulated bugs, and model quantization, detecting 4-bit quantization with AUC $>$ 0.999 within 300 output tokens.
For applications requiring sample-efficient forward-pass verification, we additionally introduce Activation-DiFR, a scheme that uses random orthogonal projections to compress activations into compact fingerprints for subsequent verification. Activation-DiFR detects 4-bit quantization with AUC $>$ 0.999 using just 2 output tokens, while reducing communication overhead by 25–75\% relative to existing methods. 
We release an open-source integration with vLLM to accelerate practical deployment of verifiable inference.

\end{abstract}

%% file: sections/01-intro.tex
\section{Introduction}

Large language models (LLMs) have become essential infrastructure for numerous applications, yet verifying the integrity of the inference process remains a fundamental challenge. Users accessing models through third-party providers need assurance that the advertised configuration is faithfully executed, while providers themselves require quality assurance mechanisms to detect infrastructure bugs before they affect users at scale. Without robust verification methods, neither party can confidently ensure that inference outputs were computed according to the desired specification, \updated{or that they have not been subtly altered by malicious tampering.}

Recent incidents across the industry have demonstrated the practical importance of inference verification. Evaluations of providers serving the Kimi K2 model revealed dramatic quality variations \citep{moonshot2025k2verifier}, with some vendors achieving only 61-72\% similarity to a reference implementation as measured by coarse statistics about its tool calling behavior. Anthropic recently disclosed that a bug in their inference service went undetected for weeks, substantially degrading performance by systematically preventing the most likely tokens from being sampled. This resulted in unwanted behavior such as producing Thai or Chinese characters in response to English prompts, or producing obvious syntax errors in code \citep{anthropic2025postmortem}.

The most straightforward approach to verifying that a given inference output was computed correctly is to attempt to exactly reproduce it using the same input and a trusted implementation of the inference process. However, bitwise reproducibility is rarely achievable in practice. 
For one, floating point operations are non-associative; this means that factors such as kernel selection, batching schemes, and hardware-specific optimizations which alter the ordering subsequently alter the numerical results of these operations. One study found that running the same input through the same inference engine using Qwen-3-235B 1,000 times under greedy decoding produced 80 unique outputs \citep{he2025nondeterminism}. 

To address this challenge, we introduce \textbf{Token-DiFR} (Token-Divergence-From-Reference),\footnote{This methodology is concurrently studied for the concrete application of detecting steganography in LLM inference \cite{verifyingModelWeightExfil}} a verification method that can be applied post-hoc to unmodified sampling procedures. By conditioning on a shared sampling seed, Token-DiFR reduces inference to a nearly deterministic process. Unlike distributional verification methods that check whether outputs are statistically consistent with a reference model's distribution, allowing many valid generation paths and potential adversarial manipulation, Token-DiFR exploits the fact that seed synchronization leaves providers with almost no degrees of freedom. Empirically, over 98\% of tokens exactly match between provider and verifier (see Table \ref{tab:raw_scores_exact_match}), and when differences occur, there are typically only 2–3 alternative tokens. This tight specification means that the output tokens themselves serve as evidence of legitimate inference, auditable post-hoc with zero overhead for the provider and zero modifications to standard inference pipelines.

This near-determinism enables robust verification: Token-DiFR detects 4-bit model quantization with AUC $>$ 0.999 within just 300 output tokens (see Table~\ref{tab:auc_gumbel_margin_w99}). Thus, verifying outputs from a provider can provide high confidence in their entire inference process without needing to check every generation, as systematic issues will manifest consistently across randomly sampled outputs.

For applications requiring highly sample-efficient verification of the forward pass (setting aside the sampling process) activation-based fingerprinting methods offer complementary strengths: they do not require sampling seed synchronization and can detect misconfigurations using only a handful of tokens. We study prior approaches in this space and propose \textbf{Activation-DiFR}, a simple fingerprinting scheme that applies random orthogonal projections to hidden states, which can detect 4-bit quantization with AUC $>$ 0.999 with just 2 output tokens (see Table \ref{tab:auc_proj_k32_w99}). Across six model and configuration pairs, Activation-DiFR matches detection performance while reducing communication overhead by 25–75\% relative to existing protocols.

\paragraph{Core Contributions} Our work provides the following contributions:

\begin{itemize}
\item \textbf{Token-DiFR:} A trustless, zero-communication method that verifies correctness of the entire generation process directly from token outputs.
\item \textbf{Activation-DiFR:} An activation-based verification scheme with reduced communication overhead compared to prior fingerprinting methods.
\item \textbf{Empirical Validation:} Empirical evaluations demonstrating strong detection of quantization, sampling errors, and simulated inference bugs
\item \textbf{Open-Source Implementation:} A publicly available integration with vLLM enabling practical deployment of our verification methods today. Our implementation is available at: \url{github.com/adamkarvonen/difr}.
\updated{\item \textbf{Deployable verification for open-source models:} Token-DiFR can be applied immediately to any open-source model served by an inference provider by issuing temperature-0 (greedy) queries and replaying them with a trusted implementation, enabling practical spot-checking of inference correctness today, as we demonstrate in a small case study on third-party Llama~3.1~8B providers (Appendix~\ref{app:openrouter_case_study}).}
\end{itemize}

%% file: sections/02-related-work.tex
\section{Related Work}

Numerous methods have been proposed to verify the correctness of LLM inference performed by untrusted entities. These approaches can be categorized into cryptographic verifiable computing, activation-based validation, and sampling verification.

\subsection{Cryptographic Verifiable Computing}

Cryptographic verifiable computing allows one to verify that a computation was performed correctly on an untrusted provider using mathematical proofs. Zero-knowledge proofs (ZKPs) provide the strongest security guarantees but impose prohibitive computational overhead for LLM inference.

Recent work has explored applying ZKPs to neural network inference. \citet{chen2025zktorch} present a general-purpose compiler that converts ML models into zero-knowledge proofs using parallel proof accumulation, achieving proof generation in approximately 2,646 seconds per token for LLaMA-2-7B. The zkLLM system \citep{sun2024zkllm} uses interactive zero-knowledge proofs with sumcheck protocols, requiring 986 seconds for commitment generation and 803 seconds for proving per forward pass on LLaMA-2-13B. As TOPLOC notes, extrapolating these per-token costs across a 2000-token generation yields approximately 23 days for commitment generation and 18 days for proving, making current ZKP approaches impractical for production.

\subsection{Internal State Fingerprinting Methods}

Rather than attempting exact reproduction of inference outputs, several methods verify inference by comparing intermediate activation states. \citet{singh2025logic} propose LOGIC, which collects top-k log probabilities (k=20) at sampled positions, while \citet{ong2025toploclocalitysensitivehashing} propose TOPLOC, which captures the indices and values of top-k activation values (k=128) from the final hidden layer. TOPLOC encodes these indices and values as a polynomial to reduce communication costs and employs a single global threshold tuned to work across all models. However, this threshold-based approach proves unsuccessful for detecting minor deviations such as  quantization. Both methods enable verification through replay: the verifier re-executes inference using the provider's output sequence, recomputes activations, and compares against the provider's fingerprints.

\textbf{Limitations of Internal State Fingerprinting.} While these internal state fingerprinting methods provide strong integrity checks on internal computations, they fundamentally cannot verify the sampling process itself. Internal state fingerprinting only verifies that a sequence and its corresponding activations are \textit{consistent}—that there exists a forward pass producing this state—not that this sequence was actually generated by the model's sampling procedure. A malicious provider could generate arbitrary text, compute internal state fingerprints on that text, and achieve perfect scores despite never using legitimate sampling. This makes internal state fingerprinting a proof of \textit{reconstructability} rather than \textit{generation authenticity}.

Internal state fingerprinting also requires communication overhead, typically on the order of 5-20 bytes per token, as providers must transmit fingerprints alongside their outputs.

\subsection{Distributional Model Equality Tests}

\citet{gao2025modelequalitytestingmodel} study a related task they call \emph{Model Equality Testing}, framing it as a two-sample test between a trusted distribution $P$ and an API distribution $Q$ over prompt-completion pairs. They instantiate a Maximum Mean Discrepancy (MMD) test with string kernels, using a Hamming distance kernel on characters or tokens across full generations. Because this approach only requires samples from $P$ and $Q$, it can be applied to fully black-box APIs without access to logits or sampling internals.

\citet{zhu2025auditingblackboxllmapis} propose a Rank-based Uniformity Test (RUT) for auditing black-box APIs. For each prompt, their method draws one completion from the target API and many completions (typically 100) from a locally deployed reference model, then computes the log-rank of the target's sampled token within the reference model's probability distribution. Under the null hypothesis---where target and reference generate from the same conditional distribution---these rank statistics should be uniformly distributed. RUT requires logit access but not seed synchronization.

Both approaches verify that outputs are statistically consistent with a reference model's \emph{distribution}, rather than conditioning on a shared sampling seed. While many generation paths can produce statistically indistinguishable outputs, conditioning on a shared seed reduces inference to a near-deterministic process with far fewer degrees of freedom. We discuss the implications of this distinction in Section~\ref{sec:discussion}.

%% file: sections/03-preliminaries.tex
\section{Background}\label{sec:background}

\subsection{Nondeterminism in ML Inference}\label{sec:nondeterminism}

\updated{Reproducing numerical results across different hardware, software, or even identical systems is notoriously difficult in ML workloads. A major contributor is that floating point arithmetic is non-associative: $(a+b)+c \neq a+(b+c)$ in finite precision, so any change in reduction order can produce slightly different results \citep{he2025nondeterminism}. In LLM inference, this reduction order often depends on the effective batch size and kernel strategy, which depend on the instantaneous inference server workload, which is inherently variable. Additional variability comes from software and hardware differences such as CUDA version, GPU architecture, and kernel implementations, which change which primitives are available and how they are scheduled \citep{schmalbach2024temperature}. For mixture-of-experts models, routing and capacity constraints introduce further dependence on other tokens in the batch, since which experts a particular token uses can depend on what other sequences are being processed concurrently \citep{puigcerver2024sparse,schmalbach2024temperature}.}

\updated{\citet{he2025nondeterminism} show that much of this nondeterminism can be eliminated on fixed hardware by using batch-invariant deterministic kernels that produce identical results whenever the model, inference implementation, and device are held fixed. In such a setting, inference verification becomes simple: the verifier can re-run the model once and check exact equality of logits or tokens. However, real deployments often involve varied inference stacks, custom kernels, and multiple GPU types, and users may want to compare providers that do not share a single fixed inference implementation. In this broader setting, we expect statistical verification to remain necessary, as verifiers must tolerate benign numerical variation across acceptable configurations.}

\subsection{Inference Sampling Algorithms}\label{sec:sampling-algorithms}

Sampling operates in two stages: first, the model produces a logit vector over the vocabulary; second, a sampling algorithm converts these logits into a token. While the sampling stage itself is deterministic given a fixed random seed, numerical noise in computing the logits can cause different probability distributions, leading to different sampled tokens even with identical seeds.

Most modern inference engines, including vLLM \citep{vllm_gumbel_sampler}, use the Gumbel-Max trick \citep{gumbel1954,vieira2014gumbel,huijben2021review} for sampling because it is more efficient on GPUs than alternative approaches like the inverse probability transform (detailed in Appendix \ref{app:verification_algorithms}). See \cref{alg:inference-generation} for more details.

\begin{algorithm}[H]
\caption{Gumbel-Max Sampling}
\label{alg:inference-generation}
\begin{algorithmic}[1]
\REQUIRE Logits $\vec{\ell} \in \mathbb{R}^V$, temperature $T$, top-$k$ parameter $k$, top-$p$ parameter $p$, seed $\sigma$
\STATE Apply top-$k$ filtering: set $\ell_i \gets -\infty$ for all but the $k$ largest logits%\footnote{Top-$k$ retains only the $k$ tokens with highest logits. Top-$p$ (nucleus sampling) retains the smallest set of tokens whose cumulative softmax probability exceeds $p$.}
\STATE Apply top-$p$ filtering: set $\ell_i \gets -\infty$ for tokens outside the top-p nucleus.
\STATE $\vec{z} \gets \textsc{Gumbel}_\sigma(0,1)^{V}$
\STATE $t \gets \arg\max_{i \in \{1,\dots,V\}} \big(\ell_i + T \cdot z_i\big)$
\RETURN sampled token $t$
\end{algorithmic}
\end{algorithm}

\subsection{Inference Prefill Mode} \label{sec:inference-prefill}

During normal autoregressive generation, an LLM performs one prefill step for the input tokens and then multiple decode steps, with one per generated token. While prefill and decode perform the same mathematical operations, prefill often achieves 3-5x higher throughput in practice because it processes many tokens in parallel, making it easier to achieve full compute utilization \citep{erdil2025inferenceeconomicslanguagemodels}. The verifier can exploit this throughput advantage: once a provider generates an output sequence, the verifier simply feeds the entire sequence (both the original prompt and generated tokens) as one input sequence, obtaining logits and intermediate activations for all positions simultaneously.

%% file: sections/04-theory.tex
\begin{figure*}[htbp]
    
    \centering
    \begin{subfigure}{0.48\textwidth}
        \includegraphics[width=\textwidth]{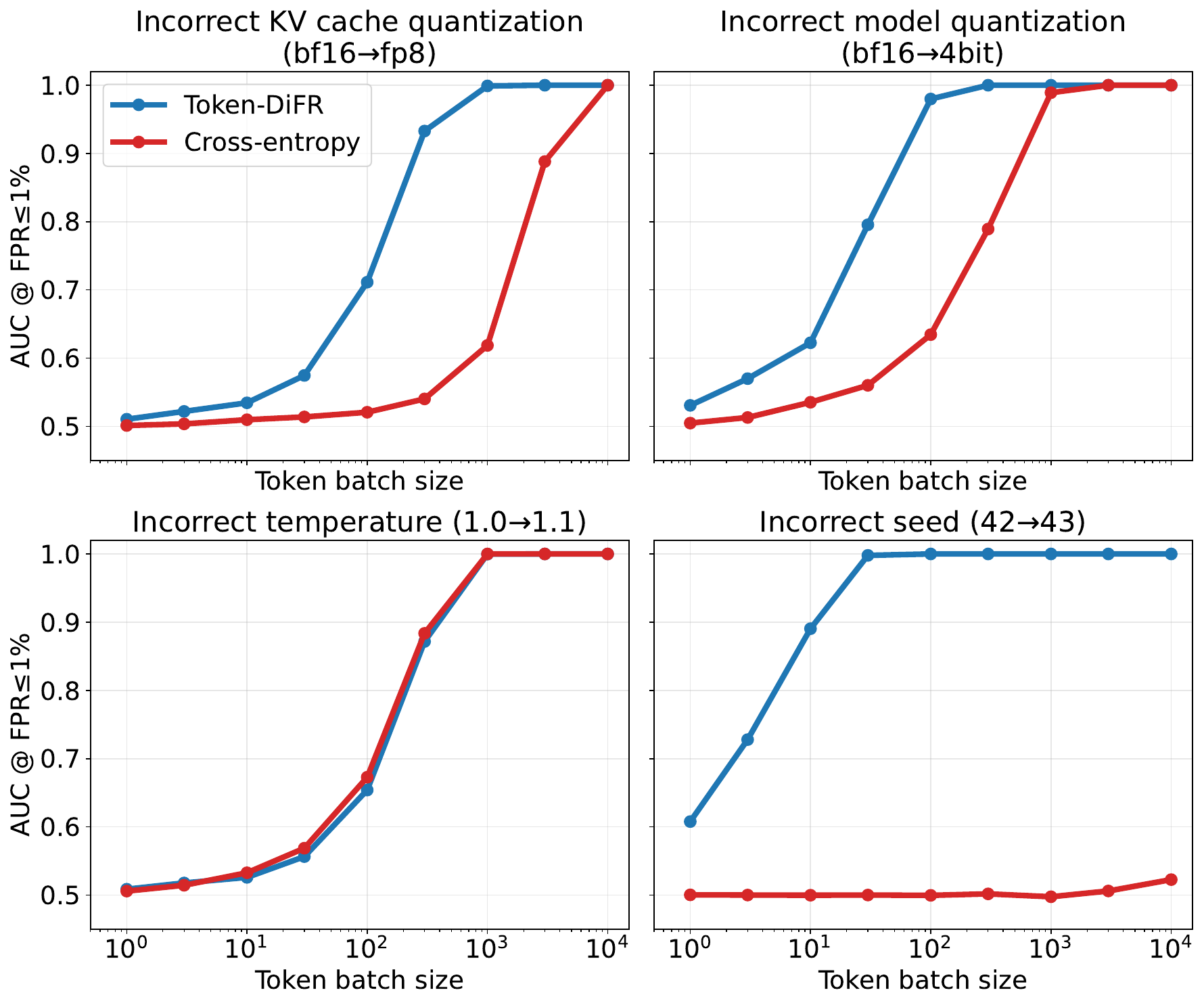}
        \caption{Llama 3.1 8B}
    \end{subfigure}
    \hfill
    \begin{subfigure}{0.48\textwidth}
        \includegraphics[width=\textwidth]{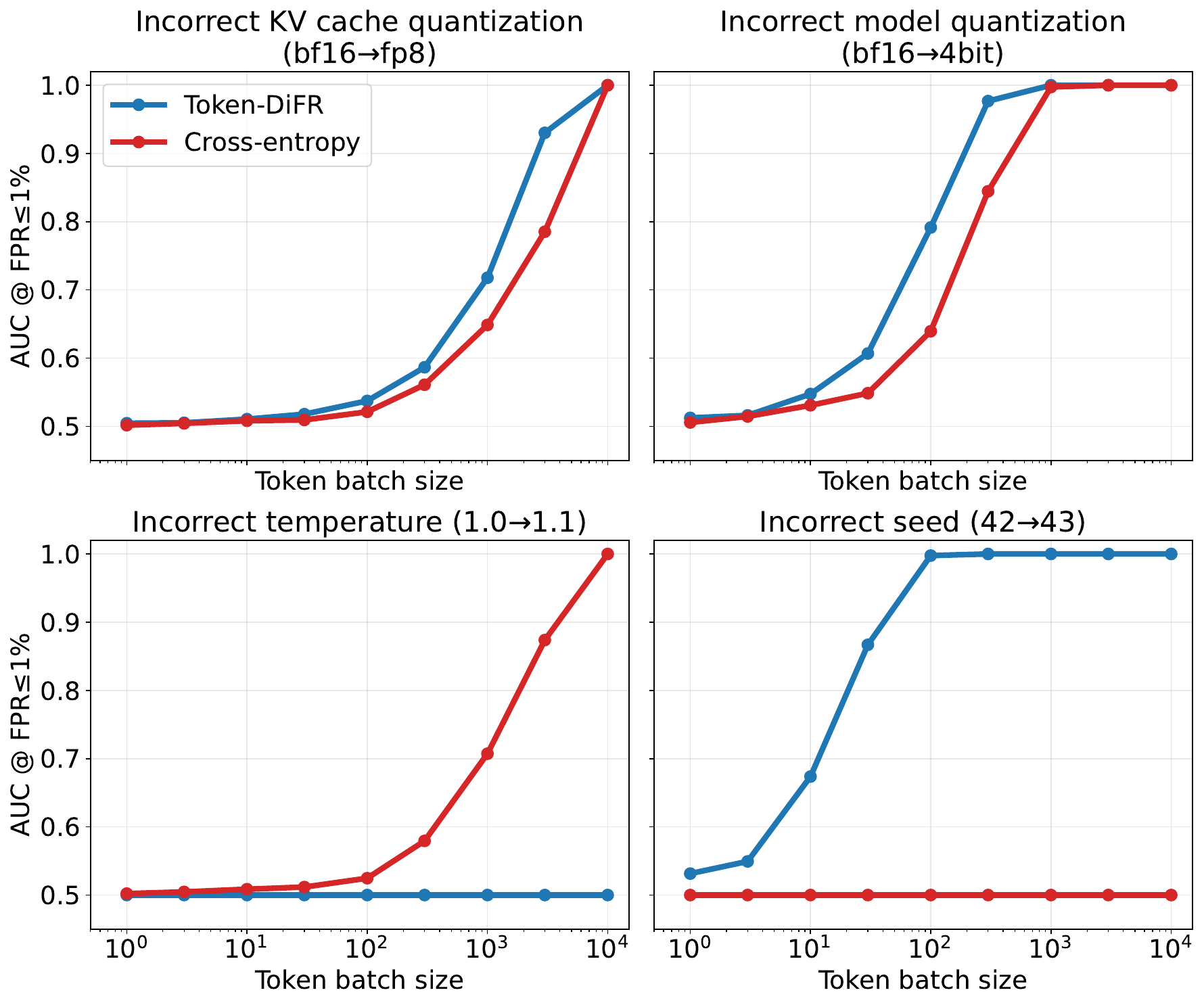}
        \caption{Qwen3-30B-A3B}
    \end{subfigure}

    \vspace{1em}
\caption{\textbf{Token-DiFR accurately detects major misconfigurations even under GPU-mismatched deployments.} We plot batch-level detection performance (AUC at 1\% FPR) as a function of the number of tokens when distinguishing the reference configuration from four misconfigurations (4-bit model quantization, FP8 KV-cache quantization, incorrect sampling seed, and temperature 1.1 vs.\ 1.0). For Llama 3.1 8B (left), Token-DiFR reliably detects all misconfigurations within a few thousand tokens. For Qwen3-30B-A3B (right), it still cleanly detects large deviations, but the smallest change (temperature +0.1) is harder to separate from the pooled honest baseline because benign differences between A100 and H200 deployments broaden the null distribution. In (matched) settings where the verifier and provider share the same H200 setup Qwen3-30B-A3B shows similarly strong performance to Llama 3.1 8B (see Appendix~\ref{app:qwen3-30b-a3b-results}).
}
    \label{fig:figure_1}
\end{figure*}

\section{Methods}
\label{sec:methods}

We present a statistical framework for verifiable LLM inference that is robust to nondeterminism. In Section \ref{sec:problem-setup}, we formalize the verification problem. Then we propose two approaches for measuring divergence-from-reference: (i) comparing token-level sampling outcomes reconstructed from a shared PRNG seed (Token-\DiFR) in Section \ref{sec:token-difr-def}, and (ii) comparing compressed internal activations (Activation-\DiFR) in Section \ref{sec:activation-difr}.

\subsection{Problem Formulation} \label{sec:problem-setup}

We consider two parties: a \textbf{provider} that computes inference outputs, and a \textbf{verifier} that seeks to confirm these outputs were produced using an inference process conforming to specification $\phi$. The specification $\phi$ fully parametrizes the expected behavior of the provider's inference process, including the model architecture and weights, numerical precision (e.g., bf16 vs. fp8), sampling hyperparameters (temperature, top-p, top-k), and the PRNG seed $\rho$.

The provider generates a batch of input--output pairs $\{(x_i, y_i^*)\}_{i=1}^B$, claiming that each output $y_i^*$ was sampled according to specification $\phi$ conditioned on input $x_i$. Informally, the verifier would like to conduct the following hypothesis test:

\begin{description}
\item[$\mathbf{H_0}$:] The provider followed specification $\phi$---that is, any discrepancies between $y^*$ and the verifier's recomputed predictions are driven purely by benign (acceptable) numerical noise.
\item[$\mathbf{H_1}$:] The provider deviated from specification $\phi$ in some way, e.g., wrong model/weights, incorrect quantization, different temperature or top-p values.
\end{description}

To accomplish this, the verifier runs a forward pass over each input--output pair $\{x_i, y_i^*\}$ using specification $\phi$ and computes a test statistic $S$ that measures the \DiFR. The batch is flagged as incorrectly computed if $S > \tau$, where $\tau$ is some pre-determined threshold, determined empirically.

\updated{To choose the threshold $\tau$ and configure the statistic $S$, we assume the verifier has access to a calibration set generated under specification $\phi$ on trusted hardware. This calibration set may come from a single ``reference'' deployment or from a pool of implementations that are all deemed acceptable (for example, different GPU types or tensor-parallel configurations). Pooling scores from these honest configurations defines the range of benign numerical variation allowed under $H_0$, and we then flag only those deployments whose statistics fall outside this pooled distribution.}

\updated{Operationally, we treat verification as a binary classification problem. For each token, the verifier observes the provider's sampled token $t^*$ and computes a scalar \emph{per-token score} $s(t^*)$ using a trusted implementation of the model and sampling procedure. Scores from many tokens are then aggregated into a batch-level statistic $S$ (for example, by averaging over a fixed batch size), and the verifier decides whether $S$ looks more like it came from a compliant configuration ($H_0$) or a misconfigured one ($H_1$). Single tokens typically do not carry enough signal for reliable decisions, so all of our experiments perform classification using aggregated scores over multiple tokens.}

\subsection{\updated{Token-\DiFR{}: Token-Level Divergence under Fixed Seeds}}
\label{sec:token-difr-def}

\updated{Although the DiFR framework can be instantiated with any sampling procedure, in this work we focus on the Gumbel-Max sampling procedure, where tokens are generated by adding Gumbel-distributed noise to logits and taking the $\arg\max$. Given a logit vector $\vec{\ell}$ and a temperature $T$, we draw Gumbel noise $\vec{g_\sigma}$ from a shared random seed $\sigma$ and form post-Gumbel scores}
\[
z_i = \vec{\ell}[i] + T \cdot \vec{g_\sigma}[i].
\]
\updated{Because the seed $\sigma$ is fixed, the noise vector $\vec{g_\sigma}$ is deterministic, so any disagreement between provider and verifier arises solely from numerical differences in the logits.%
\footnote{An additional advantage of this Gumbel-Max parameterization is that post-Gumbel logit differences are invariant to temperature rescaling, as temperature is applied to the Gumbel noise rather than the logits, which makes verification scores easily comparable across different temperature settings.}
}

\updated{Token-\DiFR{} is a per-token score that measures how much the provider's claimed token deviates from what the verifier would have sampled under the same seed. Let $t^{*}$ denote the provider's claimed token at a given position, and let}
\[
\hat{t} = \arg\max_i z_i
\]
\updated{denote the token that the verifier would sample using the same logits, temperature, and Gumbel noise. We define the \emph{post-Gumbel logit difference}}
\begin{equation}
\label{eq:post_gumbel_diff}
\delta_{\text{logit}}(t^{*}, \hat{t}, \sigma)
=
\big(\vec{\ell}[\hat{t}] + T \cdot \vec{g_\sigma}[\hat{t}]\big)
-
\big(\vec{\ell}[t^{*}] + T \cdot \vec{g_\sigma}[t^{*}]\big).
\end{equation}
\updated{If $t^{*} = \hat{t}$, then $\delta_{\text{logit}} = 0$; otherwise, $\delta_{\text{logit}}$ grows as the verifier's preferred token becomes more favored than the claimed token under the shared randomness. This scalar $\delta_{\text{logit}}$ is the basic Token-\DiFR{} score for a single token position.}

To handle cases where the claimed token $t^{*}$ is filtered out by the verifier due to top-$k$ or top-$p$ sampling, we treat its post-Gumbel score as $-\infty$, which yields $\delta_{\text{logit}} = +\infty$. In practice we clip extreme values to a maximum margin $\Delta_{\max}$ to reduce the influence of very rare outliers. Formally, the bounded Token-\DiFR{} margin used in our experiments is
\[
\mathrm{Token\text{-}DiFR}_{\text{margin}}(t^{*}, \hat{t}, \sigma)
=
\min\big(\delta_{\text{logit}}(t^{*}, \hat{t}, \sigma),\ \Delta_{\max}\big),
\]
where $\Delta_{\max}$ is a hyperparameter.

This can be computed with the following pseudocode:

\begin{pseudocode}
# logits: shape [vocab size V]
# c_token: int, claimed token t*
# T: float, sampling temperature
# max_logit_diff (Delta_max): float, clipping hyperparameter
# sigma: seed for Gumbel noise

set_random_seed(sigma)
gumbel = sample_gumbel(shape=[V])

z = logits + T * gumbel
v_token = z.argmax()          # verifier's token
diff = z[v_token] - z[c_token]

return min(diff, max_logit_diff)
\end{pseudocode}

\updated{Throughout the main text we use the clipped margin $\mathrm{Token\text{-}DiFR}_{\text{margin}}$ as our default per-token metric because it is simple, has only one hyperparameter ($\Delta_{\max}$), and performs well across a wide range of settings. In Appendix~\ref{app:token-difr-variants}, we describe several alternative Token-\DiFR{} transformations, such as likelihood-style scores that place more weight on rare large deviations. These variants can offer modest gains in some regimes such as rare bug detection.}

\textbf{Cross-entropy Baseline} 
A simple baseline for inference verification is computing the cross-entropy of a sampled token - namely, we define \textit{cross-entropy} as a baseline, which measures the negative log-likelihood of the claimed token under the verifier's softmax distribution. Notably, cross-entropy does not require synchronization of the RNG seed, making it a natural point of comparison.

\textbf{Implementation compatibility.} Our implementation of Token-\DiFR{} is fully compatible with vLLM \cite{kwon2023vllm} without requiring any modifications to the inference engine. The method only requires access to synchronized PRNG seeds, which vLLM provides as a standard sampling parameter.

The logit difference metric defined above is specific to Gumbel-Max sampling, but the broader Token-DiFR approach of measuring divergence from a reference implementation can be easily applied to other sampling methods. We present verification algorithms for alternative sampling approaches in Appendix~\ref{app:verification_algorithms}.

\subsection{Activation-\DiFR{}: Compressed Activation Comparisons}
\label{sec:activation-difr}

\updated{Token-\DiFR{} verifies that tokens were sampled from a claimed model, given only access to the tokens. Specifically, it verifies that a token was honestly sampled from the model's output distribution. However, discrete tokens are an inherently low-resolution signal compared to forward pass activations.} To provide further granularity and precision, we introduce Activation-\DiFR{}, which verifies inference by comparing internal model state from the provider against those computed by a trusted verifier. 
%Unlike token-level verification, activation-based verification directly inspects the forward pass computation.

Unlike Token-\DiFR{} which only takes tokens as input, Activation-\DiFR{} requires modifying the generation process to compute and save \textit{activation fingerprints}, which are then compared against during the verification process. However, Activation-\DiFR{} also does not require using the same sampling seed on the inference server and verification server, which can simplify the operational aspect of verifying many instances with a single verification process.
We break Activation-\DiFR{} apart into a 2 step process: Activation-fingerprint collection, and Activation-fingerprint matching.

For each token position, the provider and verifier compute activations $a, \hat{a} \in \mathbb{R}^D$. Using a shared random seed $s_{\text{proj}}$, both parties generate an identical random orthogonal projection matrix $P \in \mathbb{R}^{k \times D}$ with $k \ll D$, and compute:
\[
f = P_{\sigma'} a \quad \text{and} \quad \hat{f} = P_{\sigma'} \hat{a}
\]

\begin{algorithm}[H]
\caption{Activation-Fingerprint Collection}
\label{alg:afp-collect}
\begin{algorithmic}[1]
\STATE \textbf{Input:} context $c$, token $t$, model $\theta$, projection dim $k \ll D$, projection seed $\sigma'$
\STATE $a \gets \text{ForwardPassExtractActivations}(\theta, c, t)$ 
\STATE $f \gets P_{\sigma'} a$ 
\STATE \textbf{return} $f$
\end{algorithmic}
\end{algorithm}

% ---------- Algorithm 2: Activation-Fingerprint Matching ----------

Then we compute the $\ell_2$ distance between projected activations, where higher values indicate significant deviations in the underlying forward pass:
\[
d = \lVert f - \hat{f} \rVert_2
\]

\begin{algorithm}[H]
\caption{Activation-Fingerprint Matching}
\label{alg:afp-match}
\begin{algorithmic}[1]
\STATE \textbf{Input:} context $c$, token $t$, model $\theta$, projection dim $k \ll D$, projection seed $\sigma'$, reference fingerprint $f \in \mathbb{R}^k$
\STATE $\hat{a} \gets \text{ForwardPassExtractActivations}(\theta, c, t)$
\STATE $\hat{f} \gets P_{\sigma'} \hat{a}$
\STATE \textbf{return} $\lVert f - \hat{f} \rVert_{2}$
\end{algorithmic}
\end{algorithm}

The provider transmits only the compressed $k$-dimensional vector $f$. 
To further reduce communication costs, activations can be logged at only a subset of token positions—for instance, every $J$-th token. Both $k$ and $J$ can be tuned based according to verification confidence and bandwidth constraints. 
% We empirically select $k \in \{1, 2, 4, 8, 16, 32, 64\}$ to balance communication overhead and verification sensitivity. 

\textbf{Efficient compression via Johnson–Lindenstrauss lemma.}  Transmit full activation vectors to compare against would maximize verification signal; however, this would require sending $D$ scalars per token position (where $D \geq 4096$ in modern LLMs), creating prohibitive communication overhead. Instead, we leverage the Johnson–Lindenstrauss lemma \cite{johnson1984extensions}, which guarantees that random projections can compress high-dimensional vectors into lower dimensions while approximately preserving pairwise distances. 

\textbf{Why use activations?} Activation-\DiFR{} specifically targets the forward pass rather than the sampling process, and provides a method to validate activations were generated from a claimed model.
Although this means that it is unable to verify that tokens where legitimately sampled from a sampling process, the forward pass constitutes the vast majority of inference computation and therefore represents the main economic incentive for degraded inference. 
To generate valid activation fingerprints while using a degraded model (e.g., one with quantized weights), an adversary must still recompute the forward pass with the correct model, incurring the same computational and memory costs they sought to avoid.

%% file: sections/05-experimental-design.tex
\begin{figure*}[t]
    \centering
    \includegraphics[width=0.7\textwidth]{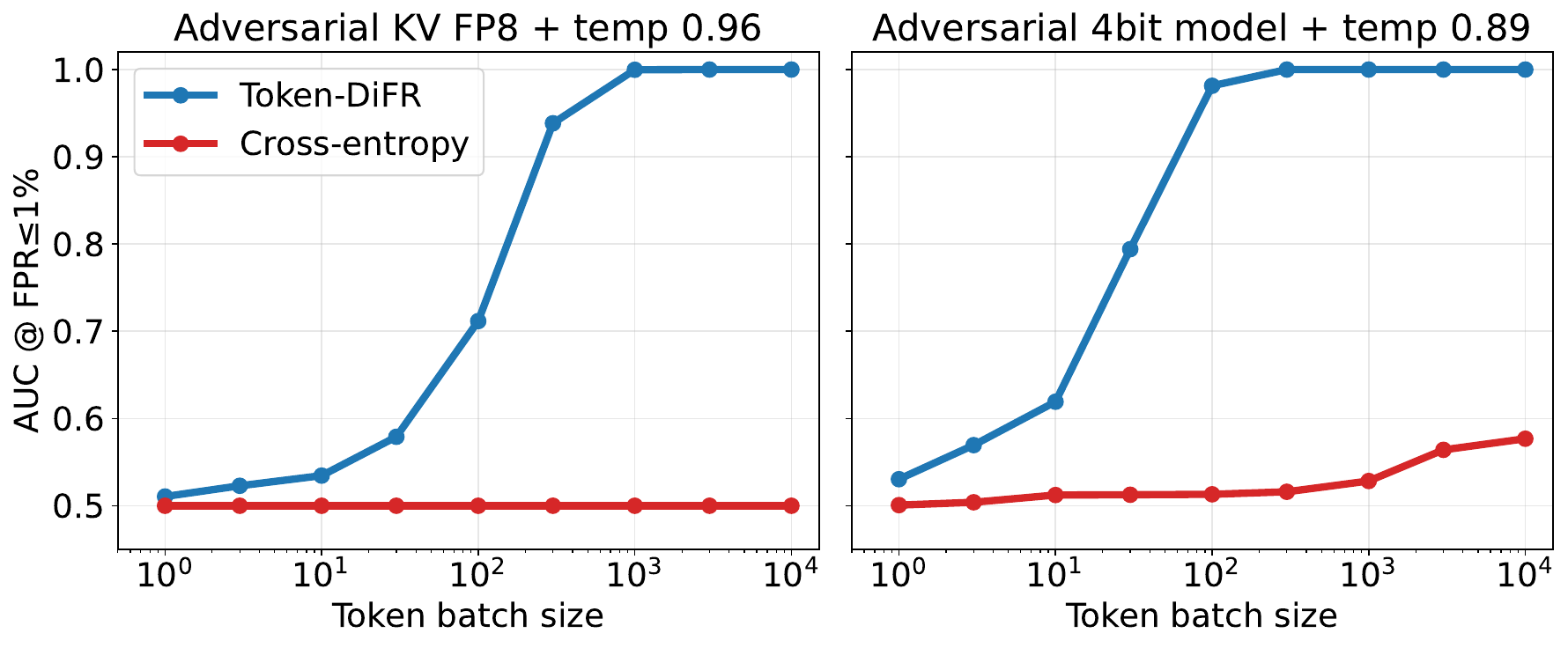}
    \caption{\textbf{Cross-entropy is vulnerable to simple adversarial manipulation.} Using Llama 3.1 8B, we consider 4-bit model quantization and FP8 KV-cache quantization and tune the sampling temperature of the misconfigured model until its mean cross-entropy matches the reference configuration. Under this attack, cross-entropy-based detectors fall to near-chance performance, while Token-DiFR maintains high detection accuracy across batch sizes.}
    \label{fig:adversarial_scaling}
\end{figure*}

\section{Experimental Design}\label{sec:experimental-design}

To evaluate Token-\DiFR{} and Activation-\DiFR{}, we collect tokens and statistics from inference performed with configurations from three categories: (i) Reference configurations: the configuration advertised by the inference provider, (ii) Correct-but-noisy configurations: inference is done with the reference configuration, but on different GPUs or with different kernels to induce benign noise, (iii) Incorrect configurations: Differences in the configuration that substantively change the inference computation, such as incorrect quantization. 

For all configurations, we sweep three models: Llama-3.1-8B-Instruct \citep{grattafiori2024llama3}, Qwen3-8B \citep{yang2025qwen3}, and Qwen3-30B-A3B \citep{yang2025qwen3}, generating up to 512 tokens per prompt from 2,000 UltraChat \citep{ultrachat_200k} prompts. These models cover a wide range of differences between models such as hidden dimensions, training distributions and broader architectural differences such as being dense or mixture-of-experts. We collect approximately 1 million output tokens per configuration, with 9 configurations per model.

\subsection{Inference Configurations}\label{sec:inference-configs}

\textbf{Reference configurations:} The configuration advertised by the inference provider. Inference performed with this configuration represents the behavior expected by the user. We use vLLM as the inference engine with bf16 precision for model weights and activations, temperature of 1.0 for sampling, top-$p$ of 0.95 nucleus sampling, top-$k$ of 50 filtering, and a fixed PRNG seed.

\textbf{Correct-but-noisy configurations.} We simulate benign numerical noise by generating outputs using setups that are algebraically equivalent to the reference configuration apart from the ordering of floating-point reductions. We do so by running inference using the reference configuration's parameters while varying the type of GPU, parallelism strategy and inference server: (i) H200 GPUs with 4-way tensor parallelism (TP-4), (ii) A100 GPUs with no tensor parallelism, (iii) A100 GPUs with TP-4, (iv) H200 GPUs with no tensor parallelism using HuggingFace as the inference implementation.

These variations introduce additional floating-point noise due to differences in GPU architecture, kernels, and communication patterns, and plausibly capture a naturalistic degree of floating-point noise in production settings. For the HuggingFace configuration, we use a custom logits processor to replicate vLLM's sampling procedure, allowing us to verify the forward pass implementation independently of the sampling code. Together with the reference configuration, these form the null distribution for calibrating our statistical tests.

\updated{Pooling scores from multiple honest implementations is scientifically useful but introduces deployment overhead. In practice, a simpler approach is to use controlled perturbations such as FP8 KV cache quantization on a single reference implementation to define a tolerated noise band.}

\paragraph{Incorrect configurations.} To evaluate whether our metrics can identify actual misconfigurations in the inference process, we manually introduce four separate misconfigurations, each of which modifies exactly one parameter of the reference configuration, leaving the rest unaltered: (i) FP8 KV cache quantization. (ii) 4 bit model quantization instead of bf16. (iii) Temperature of 1.1 instead of 1.0. (iv) Incorrect random seed usage. All incorrect configurations run on H200 GPUs with no tensor parallelism. Additional details are provided in Appendix~\ref{app:config_details}.

\paragraph{\updated{Sampling Bug}}
\updated{To evaluate Token-\DiFR{}’s ability to detect implementation bugs, we simulate sampling bugs by introducing controlled deviations from the intended sampling process. In the reference configuration, tokens are sampled according to the model’s softmax distribution at temperature $T$. In the buggy configurations, 1\% of the time, the sampler instead ignores the model’s logits and uniformly selects one of the top-$k$ tokens (for $k \in {2, 32}$). For the remaining 99\% of tokens, normal sampling proceeds unchanged. }

\subsection{Verification Methodology}\label{sec:verification-methodology}

\textbf{Token dataset generation.} We first generate output sequences using each of our reference and correct-but-noisy configurations (our ``honest'' configurations) and each incorrect configuration. We then implement verification by running a single forward pass over each sequence using the reference configuration, measuring the divergence-from-reference at each token position using Token-\DiFR{} and Activation-\DiFR{} (as described in Section~\ref{sec:methods}) as well as some baseline methods such as cross-entropy and TOPLOC. We partition our dataset into training (50\%) and test (50\%) sets at the token level.

\textbf{Feature engineering.} We first multiply features by -1 if need be to ensure that higher scores indicate greater divergence-from-reference. We then winsorize (clip values to a chosen percentile) the continuous scores computed by Token-\DiFR{}, Activation-\DiFR{}, and TOPLOC at the 99th, 99.9th, 99.99th, and 99.999th percentiles found in the training split of the honest set, evaluating each winsorized variant separately to investigate the effect this preprocessing step has on batch-level classification performance. For Token-\DiFR{} features, values are set to infinity when the inference provider's claimed token is filtered out by top-$p$ or top-$k$ at verification time. We exclude these infinite values when computing percentiles, and then clip the infinite values to the corresponding percentile threshold. 

\textbf{Detector calibration.} We randomly sample batches of tokens without replacement from each split within-configuration and compute the mean score per batch, sweeping from 1 to 10,000 tokens. This simulates settings in which the verifier has access to only one short output sequence, as well as settings in which the verifier has access to a large set of output sequences. We then train binary classifiers\footnote{Practitioners may prefer to use parametric models to more precisely estimate classifier quality in the low false positive rate regime.} on aggregated scores from the training set to discriminate between batches drawn from a particular set of honest configurations and batches drawn from a particular incorrect configuration, as defined in Section~\ref{sec:inference-configs}.

This search is computationally inexpensive, since each setting only requires fitting a one-dimensional classifier on pre-computed scores, but the optimal clipping value can vary across models and deployment environments depending on the magnitude of benign logit noise (e.g., across different GPUs or kernels). The best settings will generally differ across applications, such as detecting subtle quantization changes, catching major implementation bugs or using the wrong model, or detecting steganography \cite{verifyingModelWeightExfil}. For example, some settings might focus on high-percentile statistics of the score distribution or include several summary features instead of a single mean. We find that clipping at the 99.9th percentile generally works well for quantization detection, while more aggressive 99.999th-percentile clipping is better for rare bug detection. In our experiments, choosing an appropriate clipping threshold can substantially improve sample efficiency, reducing the number of tokens needed to reach a target AUC by more than a factor of five.

\textbf{Performance evaluation.} We report area under the curve (AUC) at 1\% false positive rate (FPR), reflecting deployment scenarios where false positives may incur significant costs. If any features were found to have lower divergence-from-reference in the incorrect set than in the calibration set, then we do not train any downstream classifiers using that feature, setting their AUC to 0.5 (equivalent to random chance). This mirrors realistic deployments: when the null distribution already spans many GPUs and implementations, a provider whose setup happens to match the verifier’s (and therefore looks \emph{more} consistent than average) should not be penalized, so we only treat configurations as suspicious when their scores are more divergent than the pooled honest baseline, not less.

% \begin{table}[t]
%   \centering
%   \small
%   \begin{tabular}{lcc}
%    % \toprule

%     & Llama-3.1-8B & Qwen-3-30B-A3B \\
%     \cmidrule(lr){2-2}\cmidrule(lr){3-3}
%     \midrule

%     FP8 KV Cache & 1000 & 10000 \\
%     4-bit Model quantization & 300 & 300 \\
%     Temperature & 1000 & N/A (0.769) \\
%     Seed & 30 & 100 \\
%     \bottomrule
%   \end{tabular}
%   \caption{Tokens required to achieve AUC \> 0.999 for detecting various misconfigurations using Token-DiFR. These results correspond to the same experiments as Figure~\ref{fig:figure_1}, but report complete AUC rather than AUC at 1\% FPR. For Qwen3-30B-A3B temperature detection (1.1 vs.\ 1.0), detection does not reach this threshold; the parenthetical value shows the maximum AUC achieved.}
%   \label{tokens_per_auc}
% \end{table}

%% file: sections/06-results.tex
\section{Results}
\label{sec:results}

We evaluate our proposed verification methods across a range of misconfigurations and deployment scenarios. Our experiments demonstrate that both Token-\DiFR{} and Activation-\DiFR{} provide reliable signals for detecting deviations from claimed inference configurations.

We evaluate Token-\DiFR{} and Activation-\DiFR{} across three models (Llama 3.1 8B, Qwen3-8B, and Qwen3-30B-A3B) and nine configurations per model, spanning correct-but-noisy environments and a range of misconfigurations. Our main findings are:

\begin{enumerate}
    \item When the verifier closely matches the provider's hardware and inference stack, Token-\DiFR{} is a strong default that reliably detects both large and small misconfigurations across all models. Unlike cross-entropy, it is robust to simple adversarial manipulations.
    \item When we deliberately pool together diverse “honest” implementations of the same model (different GPUs, parallelism setups, and inference engines), Token-DiFR continues to detect large problems (such as 4 bit quantization, FP8 KV cache quantization, or incorrect seeds) but sometimes struggles with the smallest deviations (temperature off by 0.1, k=2 sampling bug), especially for Qwen3-30B-A3B.
    \item Activation-\DiFR{} achieves near-saturated detection performance for quantized models using very small batch sizes, and Pareto-dominates TOPLOC in terms of communication cost versus detection accuracy.
\end{enumerate}

We now expand on each of these points.

\subsection{Token-\DiFR{} in Matched vs. Mismatched Environments}

\textbf{For Llama 3.1 8B and Qwen3-8B, Token-\DiFR{} maintains strong performance across all settings.} When we aggregate across GPUs and inference engines for these two models, we observe almost no degradation in performance relative to the matched setting. As shown in Figure~\ref{fig:adversarial_scaling}, Token-\DiFR{} also remains robust to a simple adversarial manipulation where an attacker tunes the sampling temperature to match the mean cross-entropy of the reference configuration, a setting in which cross-entropy based detectors fail. We obtain near-perfect AUC for all misconfigurations by aggregating over at most a few thousand tokens. Large deviations such as 4-bit quantization or an incorrect seed are detected with high AUC after 100–500 tokens, while more subtle changes such as fp8 KV cache quantization typically require on the order of 1,000–5,000 tokens to reliably distinguish from benign numerical noise.

Token-\DiFR{} requires many tokens for a simple reason: with minor misconfigurations, empirically we often observe that over 95\% of generated tokens exactly match the trusted configuration (see Section \ref{app:raw_score_distributions}). When a misconfigured model produces the identical token sequence, that individual sample is effectively legitimate, and detection must rely on subtle aggregate deviations from the expected distribution. Reliable detection therefore comes from aggregating evidence over many tokens rather than from any single generation.

\textbf{For Qwen3-30B-A3B, pooled honest environments make the smallest misconfigurations hard to detect with a single margin statistic.} For this model, we found that not all “honest” implementations behaved equally. As shown in Table~\ref{tab:raw_scores_margin}, the HuggingFace implementation yielded much larger benign Token-\DiFR{} gaps than the vLLM-based configurations, with high-percentile values on the same order as our FP8 KV cache quantization setting. This raises a conceptual point: different inference implementations can introduce real behavioral differences, and it is ultimately up to the end user whether to treat those differences as acceptable numerical noise or as deviations worth flagging. In our main analyses, we treat behavior comparable to FP8 KV cache quantization as too large to fold into the null, so we exclude the HuggingFace runs from the default pooled-vLLM results and present them separately in Appendix~\ref{app:qwen3-30b-a3b-results}.

In the resulting pooled vLLM-only setting, Token-\DiFR{} still cleanly detects large deviations such as 4-bit quantization and incorrect seeds, achieving a near-perfect AUC with a few thousand tokens, and it also reliably detects FP8 KV cache quantization. However, the smallest deviations we study for Qwen3-30B-A3B, namely a temperature shift from 1.0 to 1.1 and the simulated k = 2 bug, remain difficult to separate from the broadened null distribution. Token-\DiFR{} continues to detect the other misconfigurations, but fails to reliably distinguish these two at our target false positive rate, while cross-entropy does. This is perhaps unsurprising: increasing the temperature by 0.1 induces a small but systematic bias toward higher-entropy token distributions, which cross-entropy measures directly as a shift in average negative log-likelihood. However, in Appendix~\ref{app:rare-bug-pooling} we show that simple alternative aggregation schemes which emphasize rare large deviations restore strong Token-DiFR performance in this regime. For a more detailed breakdown of classification performance that separates GPU differences, reference-only runs, and fully pooled configurations, see Appendix~\ref{app:qwen3-30b-a3b-results}.

\textbf{Matched verifier and provider environments restore uniformly strong performance for Token-\DiFR{}.} When we restrict attention to matched environments, where the verifier uses the same GPU type, tensor-parallel configuration, and inference engine as the provider, Token-\DiFR{} is consistently strong for all three models. In this matched setting, even the smallest misconfigurations for Qwen3-30B-A3B are reliably detected with a few thousand tokens, as shown in the appendix (see Appendix Figure \ref{fig:qwen3_30B_h200_all_results}). 

\begin{figure*}[t]
    \centering
    \includegraphics[width=0.7\textwidth]{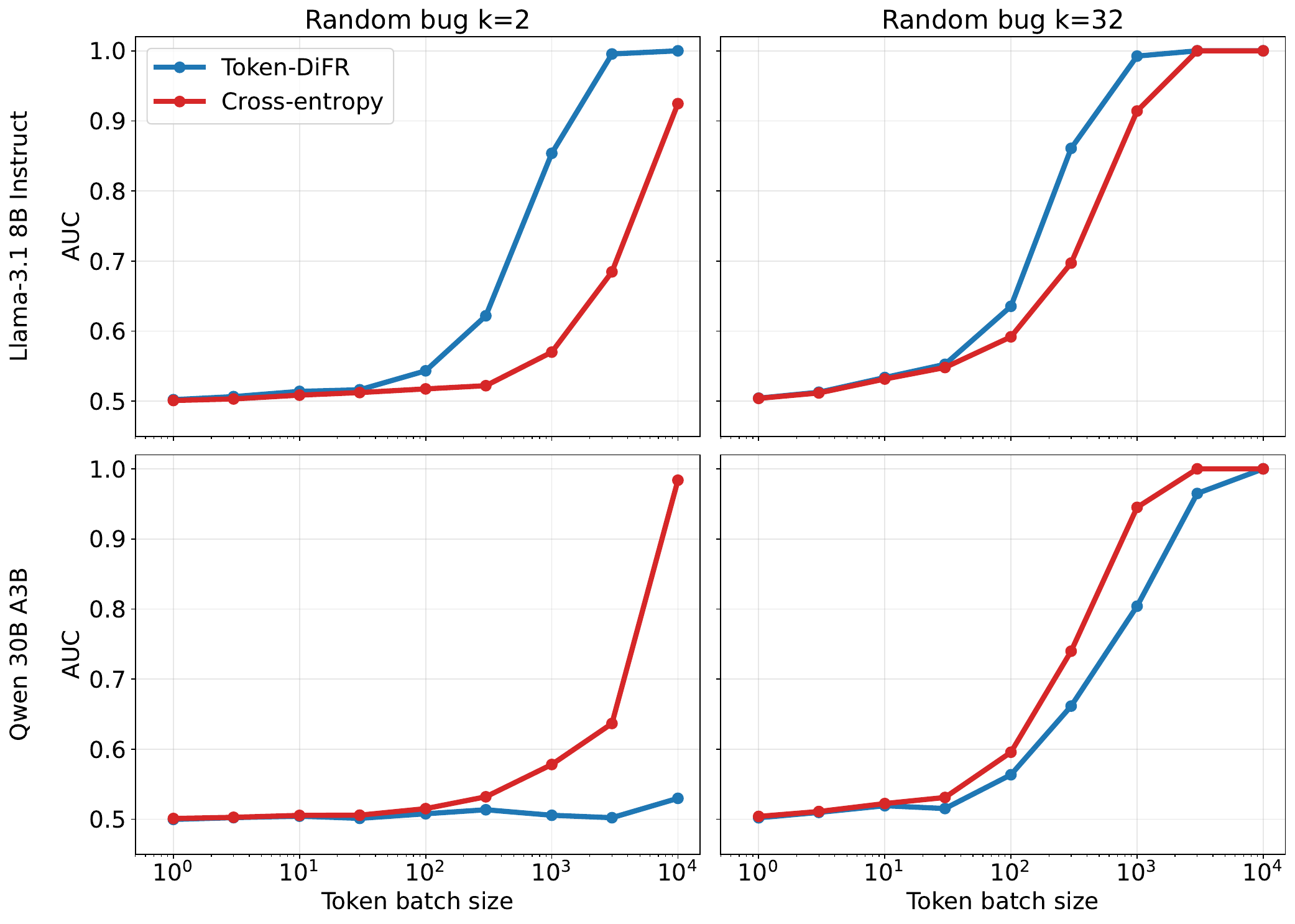}
    \caption{Detection of simulated sampling bugs for Llama 3.1 8B and Qwen3-30B-A3B. We introduce a bug that, with probability 1\% per token, ignores the model logits and instead samples uniformly from the top-$k$ tokens ($k \in \{2, 32\}$), and otherwise samples correctly. The curves show AUC at 1\% FPR as a function of batch size for cross-entropy and Token-DiFR variants. For Llama 3.1 8B, Token-DiFR detects both bug settings with modest batch sizes. For Qwen3-30B-A3B, the simple mean-pooled margin score underperforms cross-entropy in the $k=32$ case and fails to separate $k=2$ bugs from the pooled honest baseline. \updated{We pool per-token scores by taking their mean for consistency with our other figures, but in Appendix~\ref{app:rare-bug-pooling} we show that simple alternative aggregation schemes which emphasize rare large deviations restore strong Token-DiFR performance in this regime, and we recommend that practitioners consider monitoring multiple aggregation strategies in parallel.}}
    \label{fig:main_body_bug}
\end{figure*}

\begin{figure*}[t]
    \centering
    \includegraphics[width=0.7\textwidth]{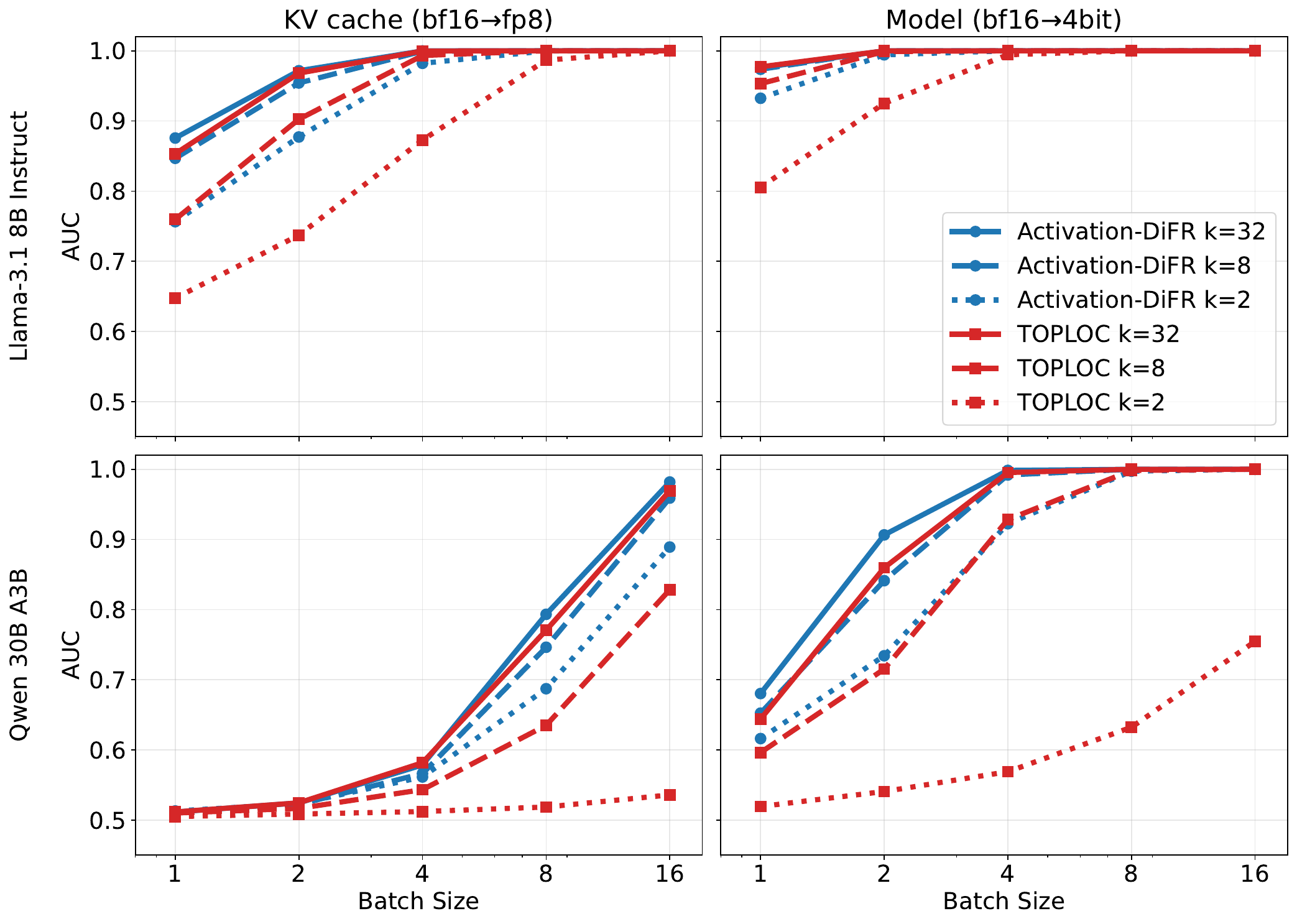}
    \caption{\textbf{Activation-DiFR detects quantization with small batches and outperforms TOPLOC at fixed communication cost.} We plot AUC at $<$ 1\% FPR as a function of batch size when distinguishing the reference configuration from 4-bit model quantization and FP8 KV-cache quantization, using activation fingerprints with projection dimension $k \in \{2, 8, 32\}$. Across both Llama 3.1 8B and Qwen3-30B-A3B, Activation-DiFR reaches near-saturated detection with small batches (for example, FP8 KV-cache quantization on Llama 3.1 8B is detected with roughly 4 tokens at $k=8$) and, at any fixed $k$, matches or exceeds TOPLOC’s detection accuracy. In the mismatched Qwen3-30B-A3B setting, classification still eventually saturates but requires larger batch sizes, similar to the increased difficulty observed for Token-DiFR under pooled honest environments.}
    \label{fig:main_body_activations}
\end{figure*}

\begin{table*}[t]
  \centering
  \small
  \begin{tabular}{lcccccc}
   % \toprule
    & \multicolumn{2}{c}{Llama 3.1 8B (H200)} & \multicolumn{2}{c}{Qwen3 30B A3B (A100)} & \multicolumn{2}{c}{Qwen3 30B A3B (H200)}  \\
    \cmidrule(lr){2-3}\cmidrule(lr){4-5}\cmidrule(lr){6-7}
    AUC $\ge \tau$ & Act-DiFR & TOPLOC & Act-DiFR & TOPLOC & Act-DiFR & TOPLOC \\
    \midrule
    $0.95$ & \textbf{0.125} & 0.219 & \textbf{1.438} & 7.25 & \textbf{0.156} & 0.312 \\
    $0.99$ & \textbf{0.219} & 0.406 & \textbf{2} & 14.5 & \textbf{0.219} & 0.438 \\
    $0.999$ & \textbf{0.312} & 0.625 & \textbf{7.25} & 32 & \textbf{0.281} & 0.562 \\
    $0.9999$ & \textbf{0.375} & 0.75 & \textbf{30} & 54 & \textbf{0.375} & 0.75 \\
    \bottomrule
  \end{tabular}
  \caption{\textbf{Communication cost} [bytes per token] (lower is better). required to detect FP8 KV-cache quantization at different target AUC levels ($<$ 1\% FPR) using Activation-\DiFR{} and TOPLOC. For each model and environment, we sweep the projection dimension $k$ and logging stride $J$ (number of tokens per 32-token window on which fingerprints are recorded) and report the \emph{minimum} average bytes-per-token overhead needed to reach AUC $\ge \tau$ (see Appendix~\ref{app:act_communication-cost} for full details). For Llama 3.1 8B and Qwen3-30B-A3B on matched H200 hardware, Activation-\DiFR{} attains very high accuracy (AUC $\ge 0.9999$) with $< 0.5$ bytes per token. In the most challenging setting, where Qwen3-30B-A3B is served on A100 GPUs and verified on H200 GPUs, Activation-\DiFR{} still substantially outperforms TOPLOC, requiring between $\sim$1.5 and 30 bytes per token to reach the same AUC targets, while TOPLOC consistently incurs higher communication cost. Bold entries highlight that Activation-\DiFR{} Pareto-dominates TOPLOC at fixed detection thresholds.}
  \label{tab:comm-cost-aucs}
\end{table*}

\subsection{Activation-\DiFR{} and Communication Efficiency}
\label{sec:results-activation}
\paragraph{Activation fingerprints saturate detection at small batch sizes, especially for larger misconfigurations.}
Token-level metrics such as Token-\DiFR{} and cross-entropy provide a discontinuous signal: whenever the provider and verifier sample the same token, there is no direct evidence about the quality of the forward pass. Activation-\DiFR{} instead compares compressed internal activations and yields a continuous signal even when tokens match.

Figure~\ref{fig:main_body_activations} shows that both TOPLOC and Activation-\DiFR{} achieve near-saturated detection of 4-bit model quantization with relatively few tokens. For Llama 3.1 8B and Qwen3-8B, a logistic regression classifier trained on Activation-\DiFR{} features reaches high AUC at FPR $= 1\%$ using only a handful of tokens (on the order of 1–4) when the projection dimension $k$ is sufficiently large. For Qwen3-30B-A3B, which exhibits more benign noise, Activation-\DiFR{} still detects 4-bit quantization within roughly 4 tokens, but detecting the more subtle FP8 KV cache quantization requires larger batch sizes: in our experiments, we typically need up to 16–20 tokens to achieve comparable detection performance. In other words, activations allow us to detect subtle forward-pass deviations much earlier than token-only methods, but the required batch size can grow for particularly noisy models and small misconfigurations.

\paragraph{Activation-\DiFR{} Pareto-dominates TOPLOC in communication cost versus detection accuracy.}
We compare activation-based methods in terms of the communication cost (bytes per token) required to reach a given detection threshold. In Appendix~\ref{app:act_communication-cost}, we formalize this trade-off in a 32-token setting by sweeping the projection dimension $k$ and the number of tokens $B$ per 32-token batch that carry activation fingerprints, treating $k \times B$ as the communication cost and computing Pareto frontiers over cost versus true positive rate at 1\% FPR. Across all three models, Figure~\ref{fig:pareto-frontier} and Table~\ref{tab:comm-cost-aucs} show that Activation-\DiFR{} lies on the Pareto frontier: for a fixed AUC target at FPR $= 1\%$, it achieves the same or higher detection accuracy than TOPLOC while using less communication (see also Figure~\ref{fig:cost-at-thresholds}). For easier tasks such as detecting FP8 KV-cache quantization in the 8B models or in Qwen3-30B-A3B on matched H200 hardware, Activation-\DiFR{} reaches AUC $\ge 0.9999$ with well under 1 byte per token of overhead and only a small number of tokens. For the more challenging FP8 KV-cache quantization on Qwen3-30B-A3B with an A100 provider and H200 verifier, achieving similarly high performance requires both more tokens (on the order of tens) and higher per-token overhead: in our experiments, Activation-\DiFR{} reaches AUC near 0.999 at 7.25 bytes per token, while TOPLOC requires 32 bytes per token to match this accuracy.

%% file: sections/07-discussion.tex
\section{Discussion}
\label{sec:discussion}

\subsection{Practical Recommendations} 

\textbf{Token-DiFR is a strong default metric, while cross-entropy is a practical fallback when RNG seed synchronization is not available.} 
When deterministic sampling is unavailable—for instance, when providers cannot or will not expose seed synchronization—cross-entropy provides a reasonable alternative. While it requires approximately 5-10$\times$ more samples to achieve equivalent detection performance for certain misconfiguration types such as quantization, it remains effective for detecting major sampling bugs, incorrect temperature configurations, and quantized models. 

\textbf{In practice, we expect practitioners to maintain a small family of simple one-dimensional detectors rather than rely on a single score.} Once logits are available, Token-DiFR margins, their clipped variants, likelihood-style transformations, and cross-entropy are all cheap to compute and easy to threshold. In our experiments, more aggressive clipping (at the 99.999th percentile) tends to work better for detecting rare sampling bugs, while milder clipping (at the 99.9th percentile) improves sensitivity to subtle but systematic changes such as quantization. Cross-entropy also remains a useful auxiliary signal, particularly for bug detection in settings like Qwen3-30B-A3B where small temperature shifts have strong effects on average log-likelihood. In realistic deployments, it is natural to train and monitor several such detectors in parallel, each tuned to a different class of potential failure modes.

\textbf{A simple practical method for calibrating benign noise.}
In our experiments we build a pooled honest baseline by combining configurations across multiple GPUs, parallelism strategies, and inference engines. This setting is useful for scientific understanding, but it introduces complexity which is not required for deployment. A simpler pattern is to fix a single reference implementation and then use controlled perturbations such as FP8 KV-cache quantization or a small temperature shift (for example, $T = 1.1$ instead of $T = 1.0$) to estimate the level of numerical variation one is willing to regard as benign. Configurations that induce larger shifts are flagged as misconfigured, without requiring the verifier to enumerate or explicitly calibrate every possible implementation the provider might be using.

\textbf{Cross-entropy (the baseline) has several conceptual limitations that practitioners should consider.} Most fundamentally, it discards the signal available in the sampling seed. When conditioning on a known seed, the token generation process becomes nearly deterministic, with very few potential tokens at any step (this near-determinism also means that seed-conditioned sampling verification may be useful for detecting steganographic channels). In contrast, there are many potential generations that could achieve an average cross-entropy value.

As demonstrated in Figure \ref{fig:adversarial_scaling}, a cross-entropy based classifier can be defeated by a simple attack where the adversary tunes the temperature to obtain the correct mean cross-entropy. Additionally, cross-entropy is significantly more sensitive to both the sampling configuration (temperature, top-$p$) and the entropy of the output distribution itself. For example, generations of sequential numbers will produce systematically lower cross-entropy values than generations of random numbers, even when both are correctly sampled, because sequential outputs have inherently lower entropy. This distribution-dependent behavior can complicate classification.

\textbf{Activation-\DiFR{} provides additional assurance for critical applications and does not require seed synchronization.} 
Token-\DiFR{} alone is sufficient for most verification scenarios, reliably detecting misconfigurations with batch sizes of 5,000-10,000 tokens without requiring any modifications to standard inference implementations or communication overhead. However, at these batch sizes, Token-\DiFR{} verifies the integrity of the batch as a whole rather than individual sequences within the batch—to verify each sequence independently would require smaller batch sizes or more tokens per sequence.

Activation-\DiFR{} provides much higher sensitivity to subtle deviations such as minor quantization errors, detecting issues with as few as 1 token. For applications requiring the highest confidence, we recommend combining Activation-\DiFR{} with Token-\DiFR{} to jointly verify both the sampling procedure and the integrity of the forward pass. In addition, unlike Token-\DiFR{}, Activation-\DiFR{} does not require RNG seed synchronization, making it applicable even when providers cannot expose sampling algorithms or seeds.

\subsection{Deployment Considerations}

\textbf{Token-DiFR and Activation-DiFR require access to model weights.} 
These methods enable model providers to verify their own inference traffic and allow users to verify open-source models, but cannot verify inference from providers serving closed-weights models. 

\updated{\textbf{Token-DiFR can be deployed today for spot-checking using temperature zero.} At temperature zero (greedy decoding), Token-DiFR requires no sampling synchronization, as the verifier simply recomputes the forward pass and checks whether the provider selected the argmax token at each position. This makes immediate deployment feasible for users who wish to spot-check providers by requesting greedy generations, as we illustrate in a small in-the-wild case study on third-party Llama~3.1~8B providers (Appendix~\ref{app:openrouter_case_study}). However, this would not work for typical production usage at non-zero temperatures.}

\updated{In practice, temperature-zero spot checking would successfully detect honest misconfigurations (e.g., a provider accidentally using an incorrect tokenization format). However, it is vulnerable to selective cheating, as a malicious provider could serve high-quality inference only at temperature zero while using degraded models for non-zero temperature requests. Therefore, verifying production usage at non-zero temperatures requires access to the provider's sampling procedure and the ability to synchronize random seeds.}

\textbf{Practical deployment requires knowledge of the provider's sampling procedure.} 
For open-source inference frameworks like vLLM, this requirement is straightforward—the sampling code is publicly available and can be directly replicated by verifiers. The core sampling logic typically consists of only a few dozen lines of code, making verification implementation tractable. Providers must also expose an API parameter allowing clients to specify per-request sampling seeds with a known noise generation procedure. For example, vLLM uses a per-request \texttt{torch.Generator} passed to \texttt{exponential\_()} for Gumbel sampling. 

\textbf{We recommend that open-source inference providers standardize on a common sampling algorithm.}
Open source providers can enable verification either by sharing their sampling procedure with verifiers or by adopting a standardized implementation (such as vLLM's sampling implementation). If providers use custom sampling procedures, verification requires implementing a custom verification function per provider. Given that sampling logic is typically not a performance bottleneck, standardization imposes minimal overhead while enabling straightforward verification across providers. Alternatively, providers could offer a standard sampling mode alongside their custom implementations to support verification use cases without constraining their default behavior.

\paragraph{Practical Deployment via Random Sampling}
While our framework requires the verifier to regenerate outputs, this does not impose prohibitive costs in practice. Not every user needs to verify every generation, as random sampling allows sporadic spot-checks to detect systemic issues. Since systematic misconfigurations affect outputs consistently, relatively small sample sizes suffice to provide high confidence about overall inference quality. Modern LLMs generate tokens at costs of just a few dollars per million tokens, making verification overhead negligible relative to total inference volume. The amount of verification needed is a tradeoff of cost versus confidence.

\subsection{Comparison with Distributional Verification Methods}

Distribution-based verification methods such as MMD~\citep{gao2025modelequalitytestingmodel} and RUT~\citep{zhu2025auditingblackboxllmapis} verify that outputs are statistically consistent with a reference model's distribution, allowing many valid generation paths that could produce indistinguishable statistics. This design choice has several implications.

First, because these methods check aggregate distributional properties rather than individual tokens, they require substantial sampling; for example, RUT uses 100 reference generations per prompt. In contrast, Token-DiFR requires only a single forward pass per sequence.

Second, distribution-based metrics are vulnerable to adversarial manipulation. As shown in Figure~\ref{fig:adversarial_scaling}, cross-entropy-based detection can be defeated by tuning the sampling temperature to match expected statistics. Token-DiFR conditions on a shared sampling seed, reducing inference to a near-deterministic process where over 98\% of tokens exactly match between provider and verifier (Table~\ref{tab:raw_scores_exact_match}). This tight specification leaves little room for such manipulation.

Third, distribution-based methods struggle with subtle misconfigurations. \citet{zhu2025auditingblackboxllmapis} report that RUT fails to reliably distinguish 8-bit quantized models from unquantized baselines, whereas Token-DiFR detects such differences within a few thousand tokens. While Token-DiFR requires seed synchronization to audit production traffic at non-zero temperatures, it can be applied without seed synchronization in the temperature-zero auditing setting (see Appendix~\ref{app:openrouter_case_study}).

\subsection{Limitations}

\textbf{We have not evaluated our approach on other sampling procedures such as speculative decoding.} 
Our algorithms and experiments focus on verifying sampling from a single LLM. In practice, providers may employ additional procedures like speculative decoding to reduce inference latency or constrained decoding to follow output formats. We do not evaluate speculative decoding scenarios, as vLLM version 0.10.0 does not yet support it, and the wide variety of speculative decoding implementations would each require slightly modified verification algorithms and may require recording additional metadata. However, our general approach of replaying inference with shared randomness remains viable. For a proposed algorithm for verifying the speculative decoding method from \citet{chen2023acceleratinglargelanguagemodel}, see Appendix~\ref{sec:speculative-decoding-verification}.

%% file: sections/conclusions.tex
\section{Conclusion}

Our results demonstrate that trustless, zero-communication verification of LLM inference is practically achievable and highly effective in detecting even subtle misconfigurations. As the industry continues to scale, we advocate for inference providers to adopt standardized reference implementations and consistent sampling algorithms, enabling rigorous verification by any third-party user. Doing so will greatly enhance transparency, trust, and reliability in LLM deployments, benefiting providers and end-users alike.

\clearpage

%% file: sections/ack.tex
\section{Acknowledgments}

The majority of this work was conducted while AK, DR, RR, and LM were participating in the ML Alignment and Theory Scholars (MATS) program, whose support we gratefully acknowledge.

%% file: sections/appendix.tex
\clearpage

\section{Token-DiFR Score Transformations}\label{app:token-difr-variants}

In the main text, we use the margin-based Token-\DiFR{} score as our default verification metric. For completeness, we also explored a small number of simple score transformations on top of the raw logit gaps. Empirically, these variants yield at most modest improvements, mostly in the bug detection setting, and generally do not justify the added complexity or hyperparameters, so we do not treat them as primary metrics in the body of the paper. We nevertheless describe them here and include them in the remaining appendix plots.

\textbf{Likelihood-based variant.}
The clipped logit difference can also be converted into a likelihood-style score by introducing a simple noise model for the logits.
We treat the logit gap $\Delta$ between the claimed token and the verifier's best token as being modified by Gaussian noise with standard deviation $\sigma$, and map it to a standardized score $z = -\Delta / \sigma$.
We then define a likelihood score
\[
\mathcal{L}(t^*) = \log \Phi(z),
\]
where $\Phi$ is the standard normal cumulative distribution function.
The parameter $\sigma$ can be estimated from a small calibration set or swept over a short grid as a single additional hyperparameter for improved classification performance.

\textbf{Exact-match variant.}
We also consider a binary exact-match metric that assigns a score of $1$ if the provider and verifier sample the same token under a shared seed and $0$ otherwise.
This variant is hyperparameter-free, but it produces an inherently worse classifier as it discards significant amounts of information contained in the logit differences.

\section{Additional Verification Algorithms}\label{app:verification_algorithms}

\subsection{Monte Carlo Simulation}

For sampling algorithms where analytical likelihood computation is difficult or intractable—which could potentially include certain speculative decoding schemes—Monte Carlo simulation provides a fully general verification approach. The  idea is to explicitly model the effect of numerical noise on the logits through repeated simulation of the sampling process.

Given the verifier's computed logits $\ell \in \mathbb{R}^V$ and a claimed token $t^*$, we sample Gaussian noise $\xi \sim \mathcal{N}(0, \sigma^2 I)$ and run the provider's sampling algorithm on perturbed logits $\ell + \xi$ for $N$ trials (e.g., $N = 1000$). The verification score is simply the empirical frequency:
\[
\mathcal{L}_{\text{MC}}(t^*) = \log\left(\frac{1}{N}\sum_{i=1}^N \mathbb{1}[\text{sample}(\ell + \xi_i) = t^*] + \varepsilon\right)
\]
where $\varepsilon > 0$ ensures numerical stability. To reduce computational cost, we apply noise only to the top-$k$ logits (we use $k=100$ in practice) and set remaining logits to $-\infty$.

While Monte Carlo simulation is somewhat more computationally expensive than analytical methods and has limited resolution for distinguishing between low-probability events, it offers a simple and general approach that works with any sampling algorithm and is still relatively cheap in practice. Empirically, we find it often achieves performance close to our analytical likelihood scores.

\begin{algorithm}[h]
\caption{Inverse-Probability-Transform (IPT) Sampler}
\label{alg:ipt}
\begin{algorithmic}[1]
\REQUIRE Seed $s$, probability vector $\mathbf{p}\in[0,1]^V$ with $\sum_{t=1}^V p_t=1$
\STATE $\mathbf{C} \gets \textsc{CumulativeSum}(\mathbf{p})$ \COMMENT{Convert PDF $\mathbf{p}$ to CDF $\mathbf{C}$}
\STATE $u \gets \textsc{Uniform}(s)$
\RETURN $\min\{t\in\{1,\dots,V\}: C_t>u\}$\COMMENT{binary search: find smallest $t$ s.t. $C_t>u$}
\end{algorithmic}
\end{algorithm}

\subsection{Inverse Probability Transform Sampling}\label{sec:IPT}

Algorithm~\ref{alg:ipt} presents the Inverse Probability Transform (IPT) sampling procedure, a common alternative to Gumbel-Max sampling for token generation, used in some inference stacks like HuggingFace \cite{wolf2020huggingfacestransformersstateoftheartnatural}. Given a probability vector $\mathbf{p} \in [0,1]^V$ (obtained by applying temperature scaling, top-$k$ and top-$p$ filters, and softmax to the logit vector), IPT sampling draws a uniform random variable $u \sim \text{Uniform}(0,1)$ and returns the smallest token index $t$ such that the cumulative distribution function satisfies $C_t > u$.

\textbf{IPT-likelihood score for IPT Verification} 
We propose Token-IPT-\DiFR{}, a method for estimating the IPT-likelihood score. For a claimed token index \(g \in \{1,\dots,V\}\) produced by IPT with shared seed \(s\), the claimed token's interval on the CDF is
\[
I_g \;=\; (C_{g-1},\, C_g] \quad\text{with width}\quad w_g \;=\; C_g - C_{g-1}.
\]
To account for numerical noise in the computed probabilities, we model $u$ as corrupted by Gaussian noise with standard deviation $\sigma$. We place a Gaussian kernel on the uniform draw and evaluate the probability mass that falls inside \(I_g\). Let
\[
Z \sim \mathcal{N}(u,\, \sigma^2),
\]
where \(u\) is the shared uniform draw and \(\sigma>0\) is a calibration parameter that captures effective noise in the verifier–provider pipeline. The \emph{Token-IPT-\DiFR{}} computes
\[
\mathcal{L}_{\text{CG}}(g \mid u, \mathbf{C}, \sigma)
\;=\;
\log\!\Big( \Phi\!\big(\tfrac{C_g - u}{\sigma}\big) - \Phi\!\big(\tfrac{C_{g-1} - u}{\sigma}\big) + \varepsilon \Big),
\]
where \(\Phi\) is the standard normal CDF and \(\varepsilon>0\) is a small constant for numerical stability.

Because temperature is applied to scale the logits, the $\sigma$ or scores may have to be scaled as well to enable use of the same classifier at multiple temperature values.

\section{Implementation Details for Incorrect Configurations}
\label{app:config_details}

For reproducibility, we provide the exact vLLM arguments used to implement each misconfiguration. Temperature, seed, and top-$p$ are standard \texttt{SamplingParams} arguments in vLLM. The quantization configurations are specified during model initialization with the following arguments:

\begin{itemize}
    \item \textbf{4-bit quantization:} \texttt{\{"quantization": "bitsandbytes"\}}
    \item \textbf{FP8 model quantization:} \texttt{\{"quantization": "fp8"\}}
    \item \textbf{FP8 KV cache quantization:} \texttt{\{"kv\_cache\_dtype": "fp8", "calculate\_kv\_scales": True\}}
\end{itemize}

For Qwen3-30B-A3B, we used the pre-quantized model \texttt{RedHatAI/Qwen3-30B-A3B-quantized.w4a16} for our 4-bit quantization as bitsandbytes quantization was not supported for this model.

\subsection{Non-Zero Baseline Detection Rates}\label{app:baseline-detection}

Our experimental design intentionally mimics real-world deployment: the trusted distribution combines outputs from multiple GPU types and configurations (H200 with TP-4, A100 with and without TP-4, and H200 with HuggingFace), while each untrusted provider uses a single GPU setup. This reflects realistic scenarios where verifiers may aggregate trusted outputs from multiple sources, but each individual provider operates with a fixed hardware configuration.

This setup introduces a detection artifact: metrics like cross-entropy show small but consistent differences across GPU types and parallelism strategies. As a result, classifiers can achieve slightly elevated detection rates even for configurations that should be indistinguishable from the reference (e.g., classifying between different random seeds with the cross-entropy metric). The classifier exploits hardware-induced biases rather than actual configuration differences.

As seen in Figure \ref{fig:figure_1}, these baseline effects are minor and do not prevent detection of substantive misconfigurations, which show much higher detection rates.

\clearpage
\section{In-the-wild audit of public Llama 3.1 8B providers}
\label{app:openrouter_case_study}

\begin{figure*}[htbp]
    
    \centering
    \begin{subfigure}{0.48\textwidth}
        \includegraphics[width=\textwidth]{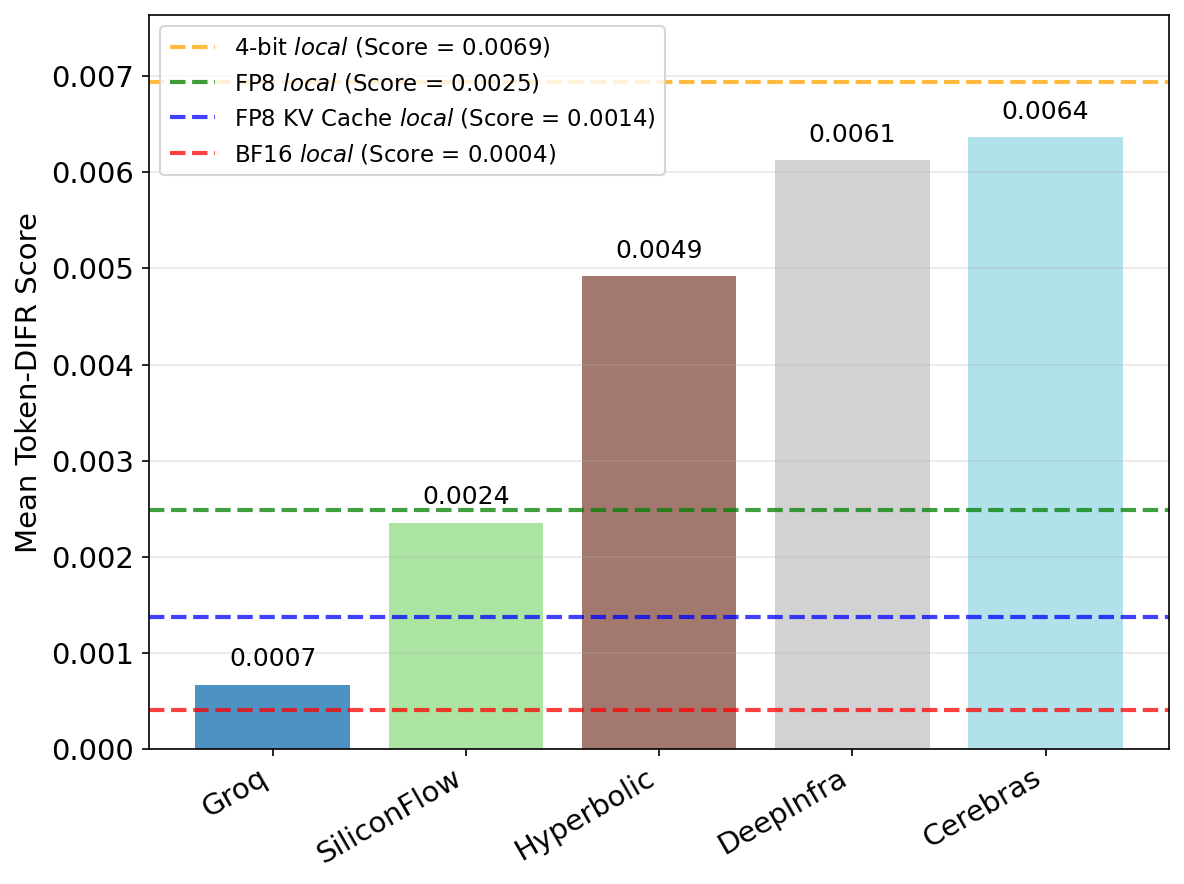}
        \caption{Token-DiFR Scores}
    \end{subfigure}
    \hfill
    \begin{subfigure}{0.48\textwidth}
        \includegraphics[width=\textwidth]{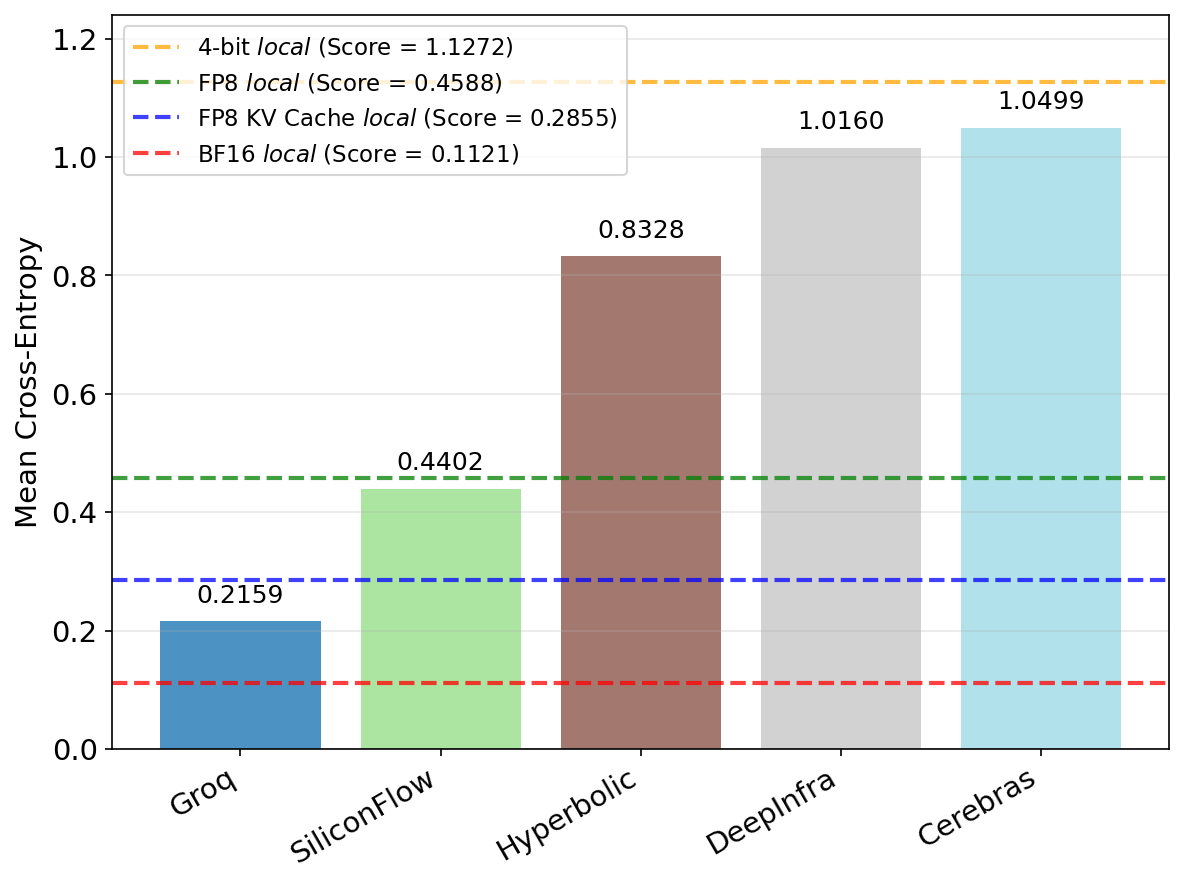}
        \caption{Cross-Entropy Scores}
    \end{subfigure}

    \vspace{1em}
\caption{\textbf{Token-DiFR and cross-entropy can audit in-the-wild Llama~3.1~8B deployments at temperature zero.}
    Low scores indicate tight adherence to the chosen reference specification, while higher scores indicate divergence, which can stem from either genuinely degraded inference (for example, heavy quantization) or benign specification differences such as alternative tokenizers or chat-template formats. 
    In this non-adversarial, temperature-zero setting, Token-DiFR and cross-entropy behave similarly and are equally simple to compute without seed synchronization.}
    \label{fig:openrouter_audit}
\end{figure*}

Existing work on RUT \cite{zhu2025auditingblackboxllmapis} and MMD \cite{gao2025modelequalitytestingmodel} evaluate existing open-source inference services; in similar a vein, to evaluate the usefulness of Token-\DiFR{}{},
we ran a small case study on several third-party providers that serve Llama~3.1~8B behind a public API.\footnote{Accessed in November~2025; implementations and configurations may have changed since then.} 

\paragraph{Setup.}
We first calibrated four local reference configurations of \texttt{meta-llama/Llama-3.1-8B-Instruct} under vLLM: a bf16 baseline, an fp8-quantized baseline, an fp8 KV-cache quantized baseline, and a 4-bit-quantized baseline. For each configuration, we sampled 2{,}000 prompts from our evaluation distribution, generated up to 500 tokens per prompt at temperature zero (greedy decoding), and computed per-token Token-DiFR scores by replaying the same generations under the bf16 reference and measuring the clipped post-Gumbel logit margin described in Section~\ref{sec:token-difr-def}. This yields a mean Token-DiFR score for each configuration, which we treat as a reference band for bf16-like, fp8-like, fp8-KV-like, and 4-bit-like behavior.

We then queried five public Llama~3.1~8B deployments via the API at temperature zero, using the same 2{,}000 prompts and generation length. For each provider, we treated its greedy outputs as the claimed tokens and computed Token-DiFR scores against our local bf16 reference in exactly the same way as for the controlled baselines, aggregating over approximately $5 \times 10^5$ output tokens per provider.

 \paragraph{Findings.}
Scores vary substantially across providers. Figure~\ref{fig:openrouter_audit} shows that Groq's Token-DiFR and cross-entropy scores fall between our bf16 and fp8-KV reference bands, which is consistent with their advertised near-bf16 configuration and gives us high confidence that they are serving something close to the intended specification. Silicon Flow lies near the fp8 reference band, which is consistent with their advertised fp8 configuration. The remaining providers (Hyperbolic, Cerebras, and DeepInfra) exhibit higher scores and therefore diverge more from our bf16 reference.

\textbf{Low Token-DiFR and cross-entropy scores indicate tight adherence to the chosen reference specification, while higher scores indicate systematic differences that may arise from either genuinely degraded inference or benign specification choices such as alternative tokenizers or chat templates.} For Cerebras and DeepInfra, manual inspection indicates that both use an earlier Llama chat-template format with an older system prompt, which likely accounts for a substantial fraction of the observed divergence, although we do not attempt to separate this from other implementation or configuration differences. We emphasize that higher scores in this case study should be interpreted as evidence of a different effective specification rather than as a direct claim about overall model quality.

\textbf{In the temperature-zero setting, both Token-DiFR and cross-entropy are simple to compute without seed synchronization and behave similarly in practice}, making them practical tools for spot-checking open-weight deployments and gaining confidence that advertised configurations are being followed. Unlike the non-zero-temperature setting considered in our adversarial experiments, it is harder to cheat cross-entropy with simple tricks such as temperature tuning, since greedy decoding largely fixes the output distribution. In principle, a malicious provider could still search for low cross-entropy generations (for example, by running beam search or reranking multiple candidates), but this would incur significant additional inference cost and latency, making such attacks less attractive in typical deployment scenarios.
\clearpage

\section{Alternative pooling strategies for rare bug detection}
\label{app:rare-bug-pooling}

\begin{figure*}[t]
    \centering
    \includegraphics[width=0.7\textwidth]{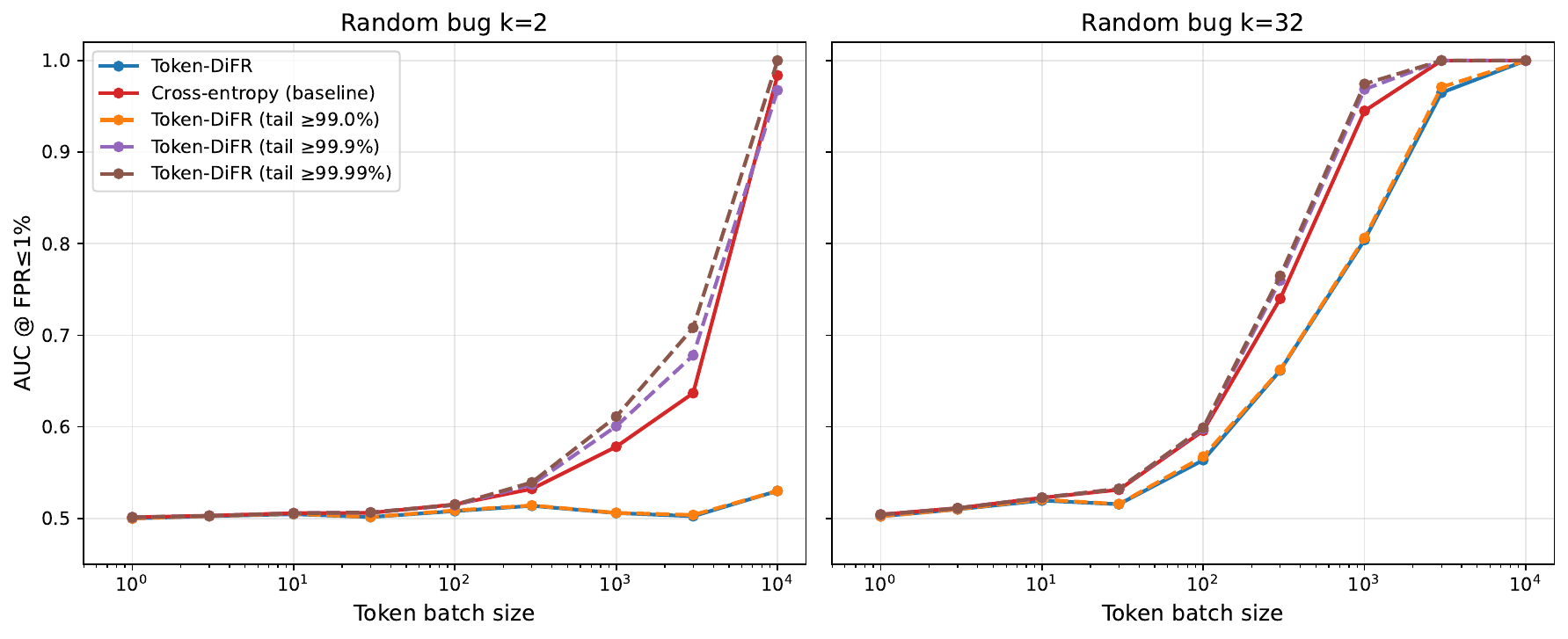}
    \caption{\textbf{Effect of tail-focused clipping on rare bug detection.} For the Qwen3-30B-A3B $k=2$ sampling bug, we compare the original mean-pooled Token-DiFR margin to a tail-focused variant that clips scores at the 99.999th percentile and zeros scores below the 99.99th percentile. The tail-focused pooling restores strong performance, reaching AUC~$=1.0$ at $10{,}000$ tokens.}
    \label{fig:rare_bug_demo}
\end{figure*}

In the main text, our default Token-DiFR detector aggregates per-token scores by taking the mean of clipped margins across all tokens in a batch. This works well for systematic misconfigurations such as quantization or temperature shifts, but rare bug regimes, where only a small fraction of tokens are affected, are particularly sensitive to how per-token scores are pooled.

To illustrate this, we consider a simple tail-focused pooling scheme applied to the simulated bug experiments in Figure~\ref{fig:main_body_bug}. Given a set of per-token Token-DiFR margin scores $\{s_j\}$ for a batch, we first clip scores at the 99.999th percentile and then set all scores below the 99.99th percentile to zero. The final batch statistic is the mean of these transformed scores. This clipping suppresses benign fluctuations while concentrating weight on rare large deviations.

On Qwen3-30B-A3B with the $k=2$ sampling bug, this simple transformation of the same per-token Token-DiFR scores is sufficient to recover strong bug detection performance: at FPR $= 1\%$, the detector reaches AUC~$= 1.0$ with $10{,}000$ tokens.

Practically, once a verifier has access to per-token scores, it is cheap to compute several pooled statistics in parallel (for example, mean, clipped mean, tail-focused pooling, or likelihood-style transforms) and either threshold them separately or combine them with a lightweight classifier. We therefore view Token-DiFR as a family of detectors parameterized both by the per-token score and the pooling rule, and recommend that practitioners select or monitor multiple aggregation strategies based on the failure modes they care about.

\clearpage
\section{Extension to Speculative Decoding} \label{sec:speculative-decoding-verification}

Speculative decoding \cite{leviathan2023fastinferencetransformersspeculative, chen2023acceleratinglargelanguagemodel, zhang2025learningharmonizedrepresentationsspeculative, li2025eagle3scalinginferenceacceleration} has emerged as a technique to accelerate LLM inference by using a smaller ``draft'' model to propose candidate tokens, which are then verified in parallel by the larger ``target'' model. This approach can achieve substantial speedups (often $2{-}3\times$) while preserving the exact output distribution of standard autoregressive sampling from the target model.

In the main body of this work, we focus our verification methodology on standard autoregressive sampling (Section~\ref{sec:token-difr-def}), where a single model generates one token at a time. This choice reflects two practical considerations: 
\begin{enumerate}[topsep=1pt,itemsep=1pt]
    \item There are a variety of proposed speculative decoding algorithms, each of which may require a slightly modified verification algorithm.
    \item At the time of writing, vLLM version 0.10 does not natively support speculative decoding.
\end{enumerate}

Nevertheless, we recognize that speculative decoding represents an important class of real-world inference strategies. In this appendix, we demonstrate that our verification framework naturally extends to handle speculative decoding with minimal modifications. We sketch the necessary adaptations using the canonical algorithm from \cite{leviathan2023fastinferencetransformersspeculative, chen2023acceleratinglargelanguagemodel} as a concrete example.

\subsection{Background: A Speculative Decoding Algorithm}

While many subsequent methods have been proposed since the original proposal, in the speculative decoding work introduced by  \cite{leviathan2023fastinferencetransformersspeculative, chen2023acceleratinglargelanguagemodel}, a draft model generates $K$ candidate tokens, which the target model then verifies in parallel. In the original work, for each candidate, the target model either accepts it, rejects and resamples from an adjusted distribution, or (if all candidates are accepted) samples one additional bonus token. This process maintains exact distributional equivalence to standard autoregressive sampling from the target model alone.

\subsection{Verification Strategy}

The key observation enabling verification is that \textit{each sampled token is a function of the random seed and the model logits from previous tokens}---precisely the quantities available to the verifier. Unlike standard autoregressive sampling, however, this particular version of speculative decoding uses three distinct sampling modes:

\begin{enumerate}[topsep=1pt,itemsep=1pt]
    \item \textbf{Accept}: Token accepted from draft proposal  
    \item \textbf{Reject}: Token resampled from adjusted target distribution $\max(q - p, 0)$
    \item \textbf{Bonus}: Final token sampled when all draft tokens accepted
\end{enumerate}

If the inference provider is using this given method of speculative decoding,then to enable verification, the inference server must record which mode was used for each output token. Given this metadata, the verifier:

\begin{enumerate}[topsep=1pt,itemsep=1pt]
    \item Performs a single forward pass with both the draft and target models over the full input + output sequence to obtain $p_i$ and $q_i$ at all positions
    \item For each token, computes the probability of observing that token under the recorded mode and seed (via analytical calculation or Monte Carlo simulation, using logit noise estimates for the draft and target models)
\end{enumerate}

This approach preserves the efficiency advantages of our verification framework: the verifier performs only prefill operations (no sequential decode), and verification cost remains dominated by a single target model forward pass (draft models are typically small).

\subsection{Generality and Limitations}

The approach outlined here applies directly to a single speculative decoding algorithm. However, many variants exist (e.g. \cite{zhang2025learningharmonizedrepresentationsspeculative, li2025eagle3scalinginferenceacceleration}), each with different acceptance criteria and resampling rules. Extending our verification framework to these methods would require identifying the sampling modes and recording which mode was used at each position. 

However, our general technique of replaying inference with a reference implementation remains applicable regardless of the underlying sampling algorithm. The primary engineering challenge is ensuring that the inference server records sufficient metadata to make each token's sampling process reproducible by the verifier.

\section{Activation Fingerprint Communication Cost Analysis}
\label{app:act_communication-cost}

To analyze the trade-off between communication overhead and detection accuracy, we examine the operational scenario where a verifier must classify batches of 32 tokens as either correctly or incorrectly generated. We limit ourselves to the setting of detecting fp8 KV Cache quantization. For each token position within a batch, the verifier can choose to collect activation fingerprints at varying granularities, parameterized by $k$—the number of features extracted per token.

\subsection{Experimental Setup}

In this setting, the minimum communication cost occurs when using $k=1$ on only a single token within each 32-token batch, yielding a cost of 1 byte per 32 tokens. At the other extreme, maximum verification fidelity is achieved by using $k=64$ on all 32 tokens, incurring a cost of $64 \times 32 = 2{,}048$ bytes per 32 tokens. Between these endpoints, we evaluate all combinations of $k \in \{1, 2, 4, 8, 16, 32, 64\}$ and batch sizes $B \in \{1, 2, \ldots, 32\}$, computing the communication cost as $k \times B$ bytes per 32 tokens.

For each $(k, B)$ configuration, we train a classifier on the training split using the methodology described in Section~\ref{sec:verification-methodology} and evaluate its true positive rate (TPR) at a fixed false positive rate of 1\% on the held-out test set. This yields a collection of $(k \times B, \text{TPR})$ points representing different operating points along the accuracy-cost spectrum.

\subsection{Pareto Frontier Analysis}

\begin{figure*}[t]
    \centering
    \begin{subfigure}{0.48\textwidth}
        \includegraphics[width=\textwidth]{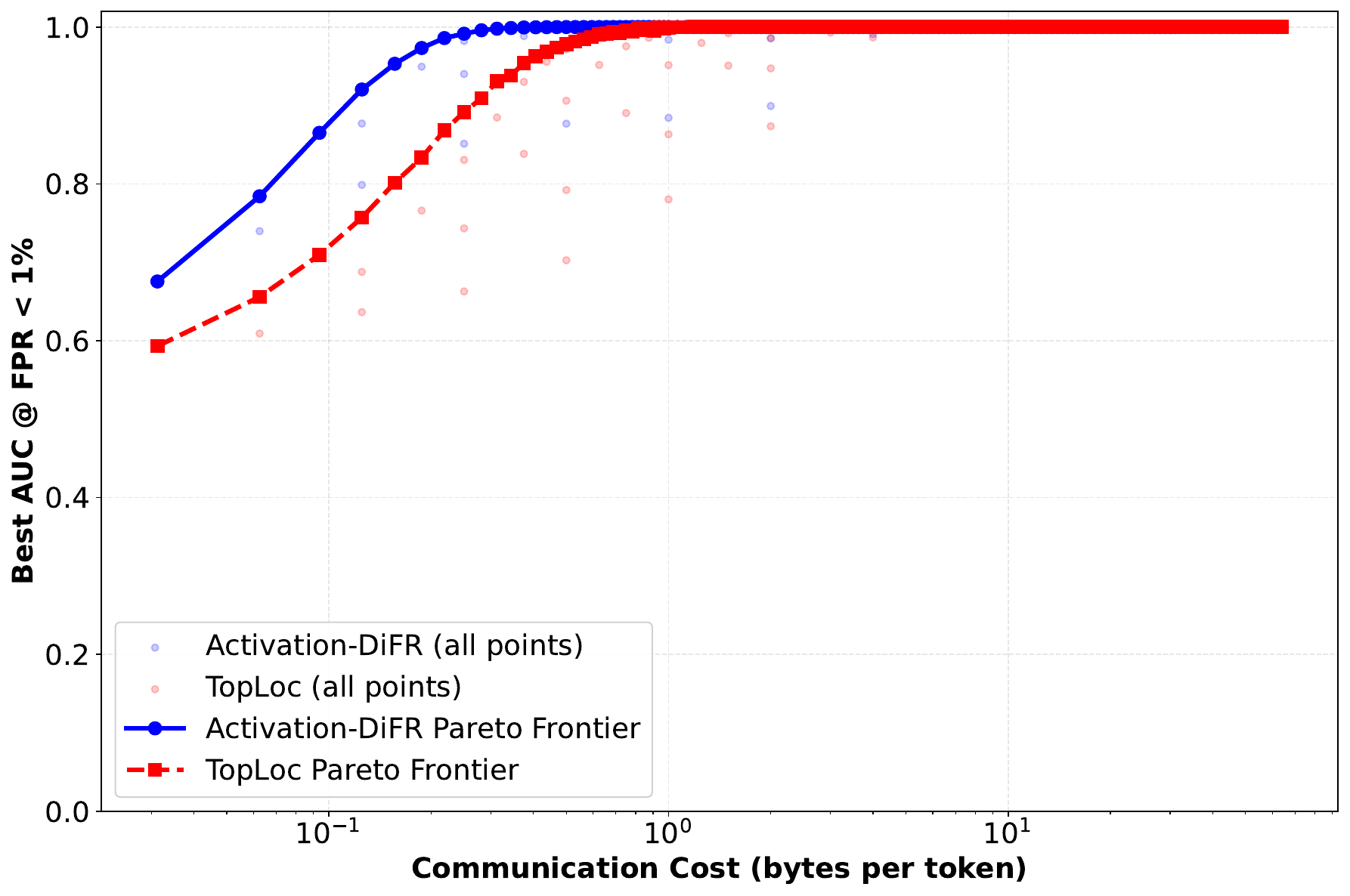}
        \caption{Qwen3-8B}
    \end{subfigure}
    \hfill
    \begin{subfigure}{0.48\textwidth}
        \includegraphics[width=\textwidth]{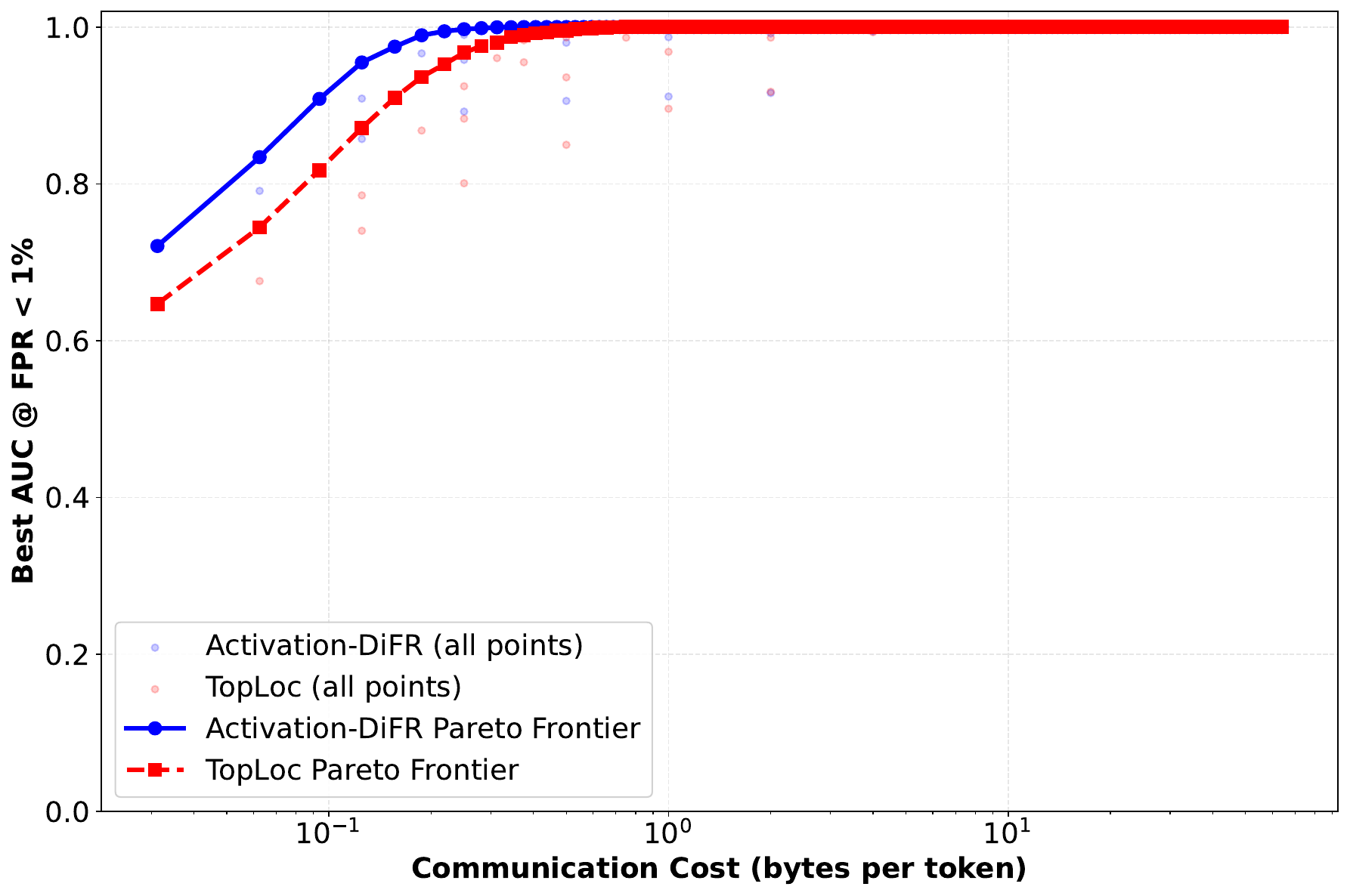}
        \caption{Llama-3.1-8B-Instruct}
    \end{subfigure}
    \caption{\textbf{Pareto frontiers for communication cost vs. detection accuracy for Qwen3-8B and Llama 3.1 8B.} Each point represents a $(k, B)$ configuration’s communication cost (x-axis, log scale) and AUC at FPR=1\%. Solid blue lines show Activation DiFR’s Pareto frontier; dashed red lines show TOPLOC’s frontier. Activation DiFR dominates TOPLOC across all cost levels and model architectures on the fp8 KV cache quantization detection task.}
    \label{fig:pareto-frontier}
\end{figure*}

We compute the Pareto frontier by identifying, for each communication cost level, the configuration achieving maximum TPR. Configurations not on this frontier are strictly dominated—there exists an alternative with equal or lower cost and strictly better detection performance. Figure~\ref{fig:pareto-frontier} presents the Pareto frontiers for Activation-DiFR and TOPLOC across three model architectures, demonstrating that Activation-DiFR consistently dominates TOPLOC across all cost levels.

\subsection{Threshold-Based Cost Comparison}

\begin{figure*}[t]
    \centering
    \begin{subfigure}{0.48\textwidth}
        \includegraphics[width=\textwidth]{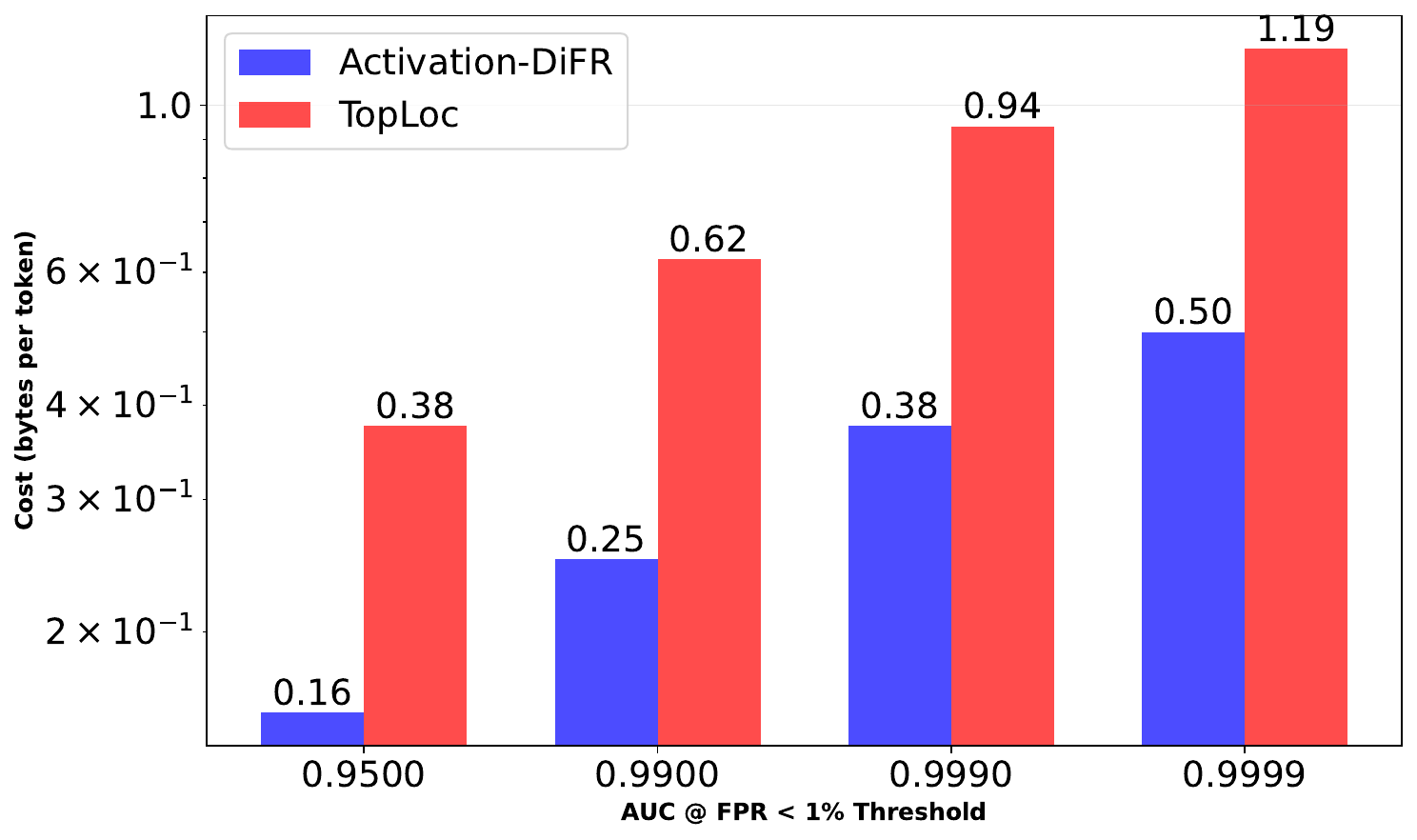}
        \caption{Qwen3-8B}
    \end{subfigure}
    \hfill
    \begin{subfigure}{0.48\textwidth}
        \includegraphics[width=\textwidth]{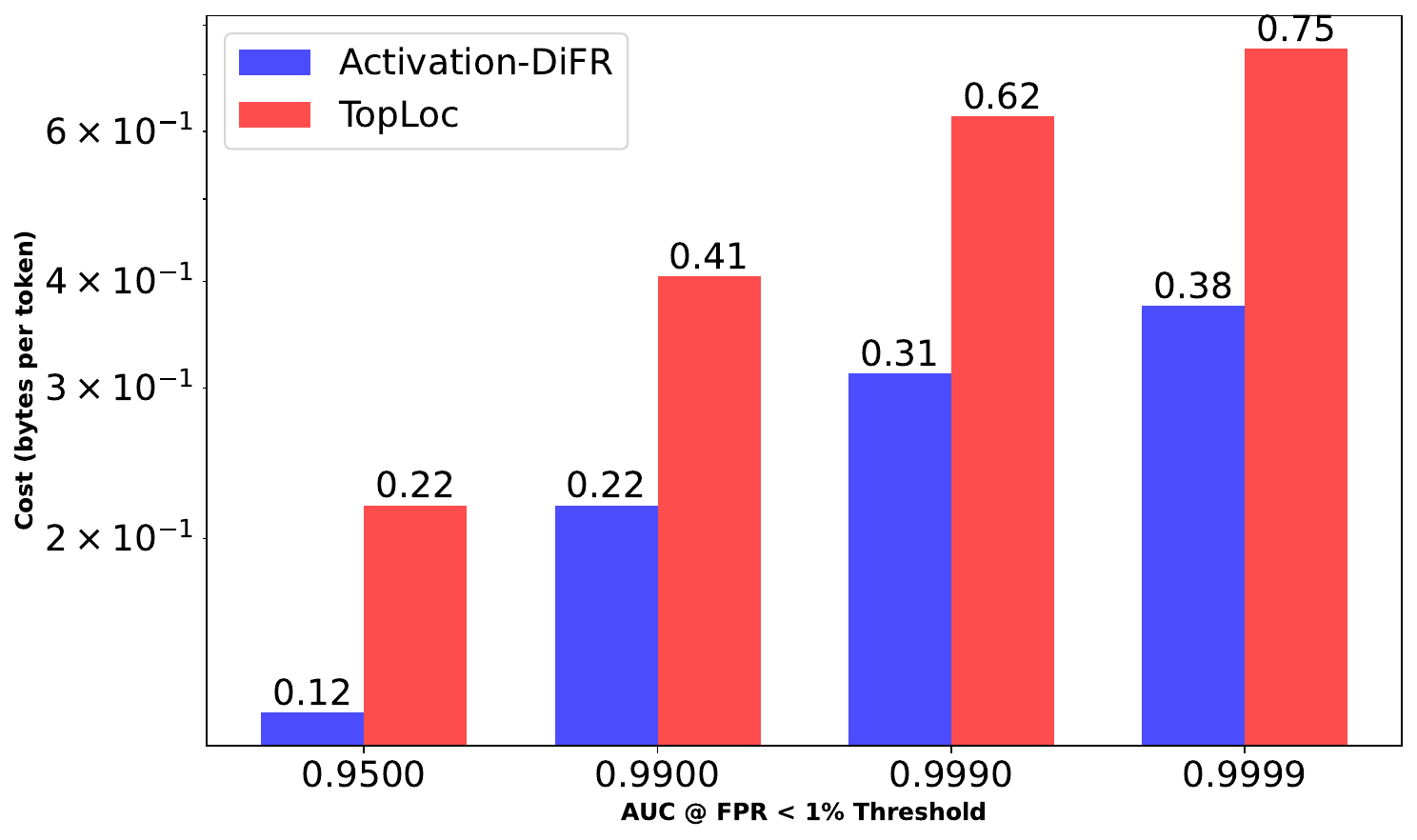}
        \caption{Llama-3.1-8B-Instruct}
    \end{subfigure}
    \caption{\textbf{Communication cost required to achieve AUC thresholds for Llama 3.1 8B and Qwen3-8B.} For each AUC threshold (x-axis) and method, we report the minimum communication cost (y-axis) required to achieve at least that AUC at FPR=1\%. Percentage labels indicate Activation DiFR’s relative cost reduction compared to TOPLOC. Activation DiFR consistently achieves target AUC at 25–50\% lower communication overhead across all models and thresholds on the fp8 KV cache quantization detection task.}
    \label{fig:cost-at-thresholds}
\end{figure*}

To quantify the communication savings more precisely, we measure the minimum cost required to achieve specific AUC thresholds: 0.95, 0.99, 0.999, and 0.9999. For each threshold $\tau$ and each method (Activation-DiFR or TOPLOC), we identify the lowest communication cost at which $\text{AUC} \geq \tau$. Figure~\ref{fig:cost-at-thresholds} presents these costs as bar charts, with communication costs annotated above each threshold. Across all three models and all thresholds examined, Activation-DiFR achieves the target AUC at 25--75\% lower communication cost compared to TOPLOC.

\begin{figure*}[t]
    \centering
    \begin{subfigure}{0.48\textwidth}
        \includegraphics[width=\textwidth]{images/comms_frontier/tokens_derived_new.Qwen_Qwen3_8B_comms_pareto_frontier_auc_fpr0p01.pdf}
        \caption{Qwen3-30B-A3B (Matched H200 GPUs)}
    \end{subfigure}
    \hfill
    \begin{subfigure}{0.48\textwidth}
        \includegraphics[width=\textwidth]{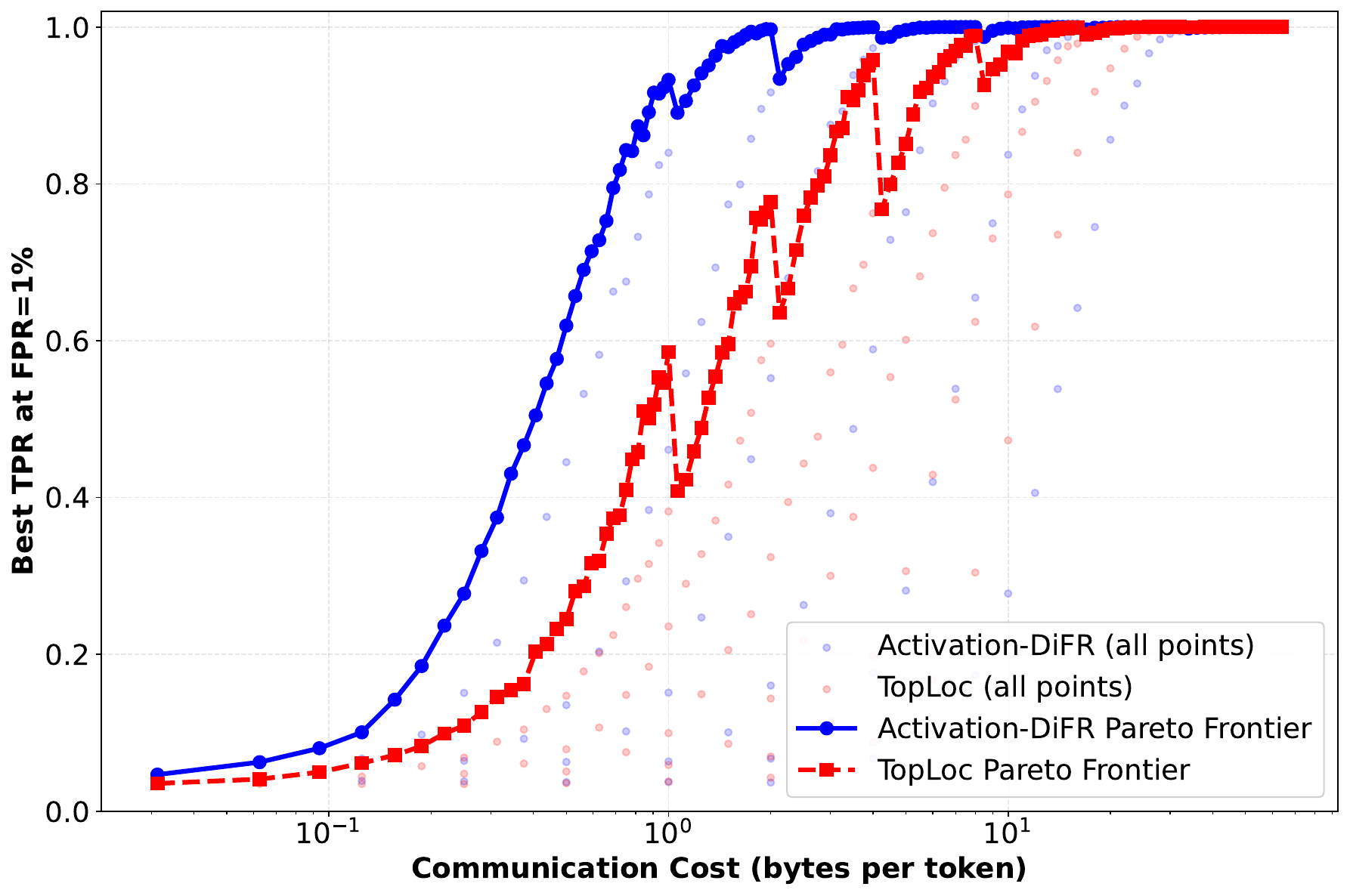}
        \caption{Qwen3-30B-A3B (A100 vs H200 GPU)}
    \end{subfigure}
    \caption{\textbf{Pareto frontiers for communication cost vs. detection accuracy for Qwen3-30B-A3B, with matching GPUs between provider and verifier (H200s) and mismatched GPUs (A100 vs H200).} Each point represents a $(k, B)$ configuration’s communication cost (x-axis, log scale) and AUC at FPR=1\%. Solid blue lines show Activation DiFR’s Pareto frontier; dashed red lines show TOPLOC’s frontier. Activation DiFR dominates TOPLOC across all cost levels and model architectures on the fp8 KV cache quantization detection task.}
    \label{fig:pareto-frontier-qwen3-30b}
\end{figure*}

\begin{figure*}[t]
    \centering
    \begin{subfigure}{0.48\textwidth}
        \includegraphics[width=\textwidth]{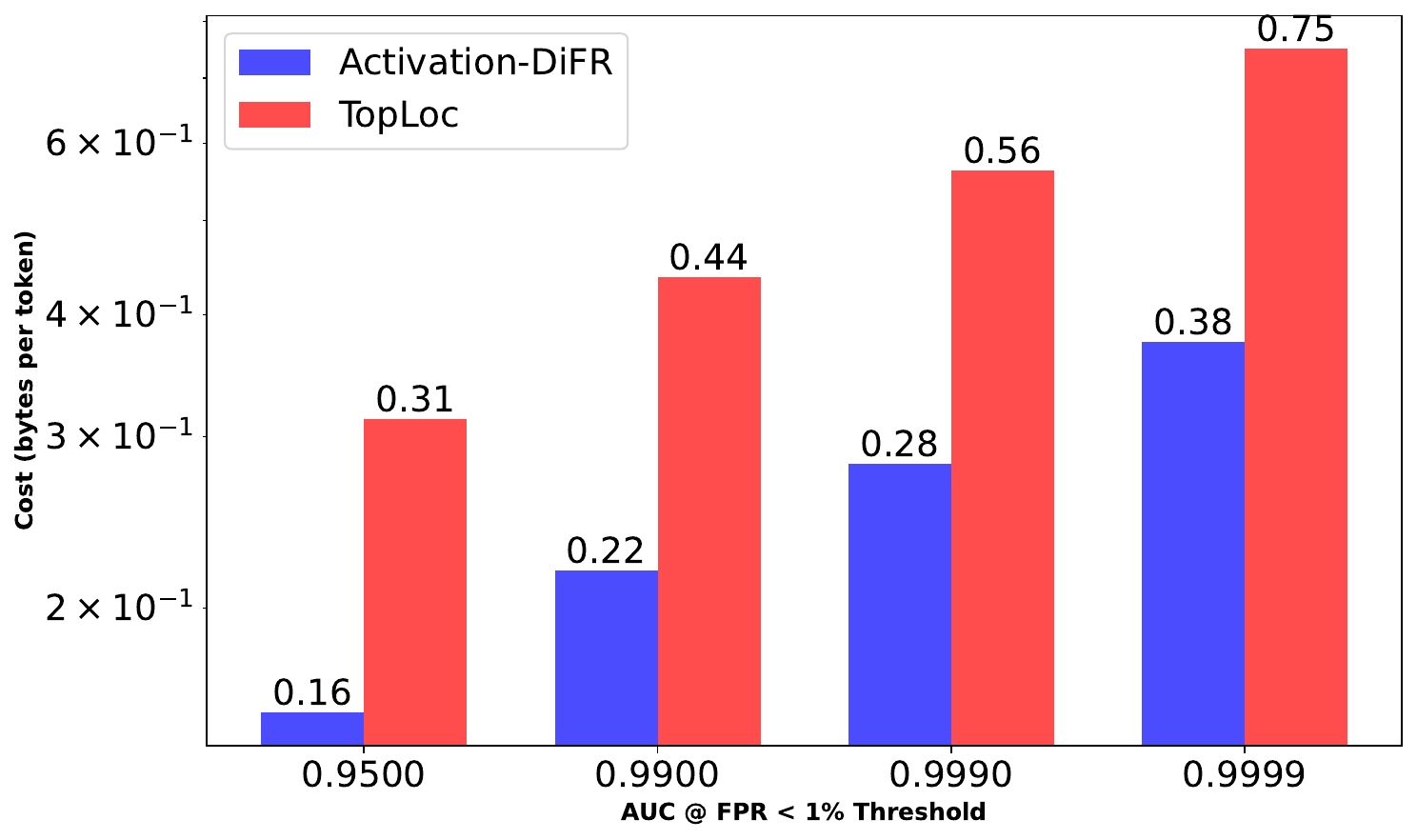}
        \caption{Qwen3-30B-A3B (Matched H200 GPUs)}
    \end{subfigure}
    \hfill
    \begin{subfigure}{0.48\textwidth}
        \includegraphics[width=\textwidth]{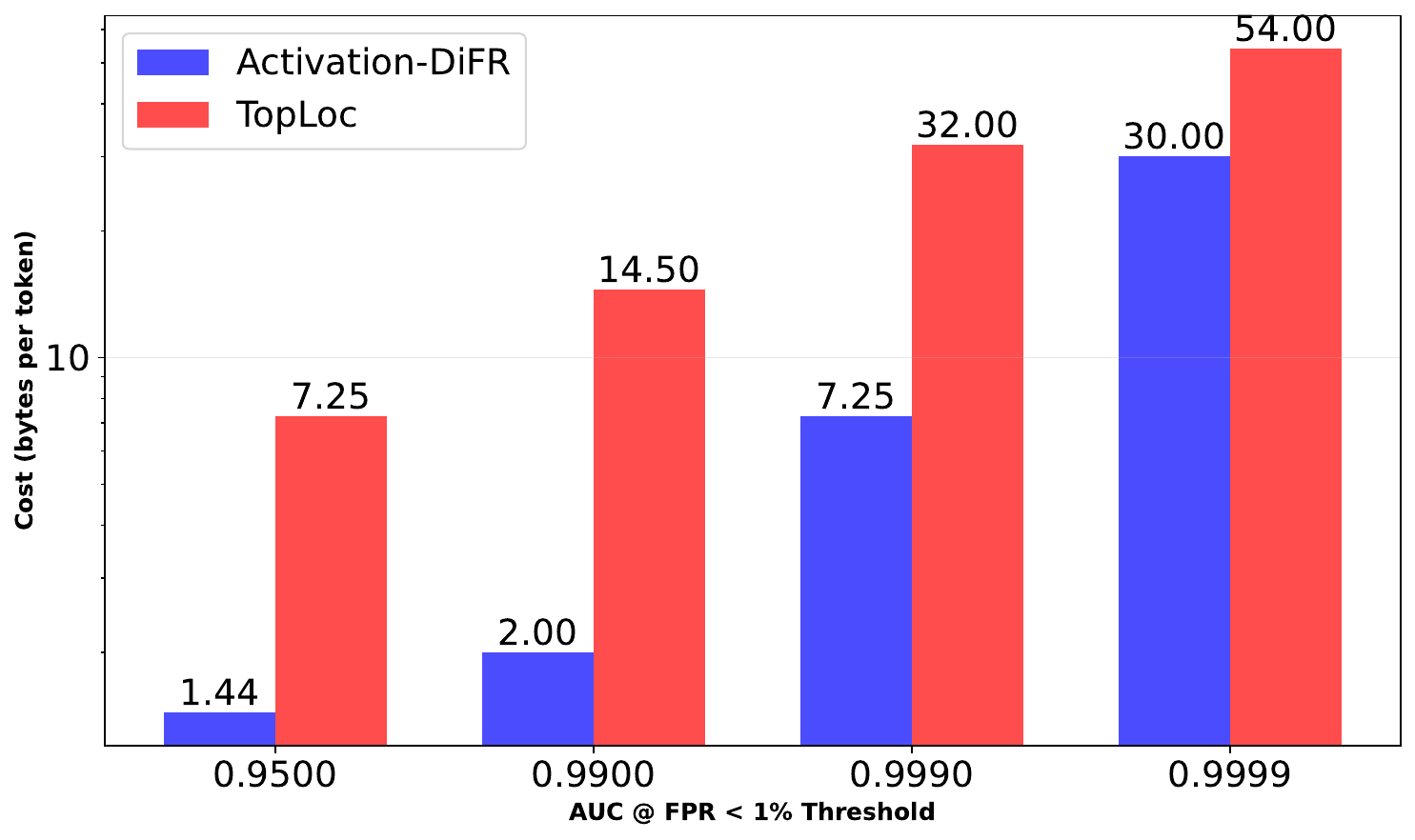}
        \caption{Qwen3-30B-A3B (A100 vs H200 GPU)}
    \end{subfigure}
    \caption{\textbf{Pareto frontiers for communication cost vs. detection accuracy for Qwen3-30B-A3B, with matching GPUs between provider and verifier (H200s) and mismatched GPUs (A100 vs H200).} For each AUC threshold (x-axis) and method, we report the minimum communication cost (y-axis) required to achieve at least that AUC at FPR=1\%. Percentage labels indicate Activation DiFR’s relative cost reduction compared to TOPLOC. Activation DiFR consistently achieves target AUC at 25–50\% lower communication overhead across all models and thresholds on the fp8 KV cache quantization detection task.}
    \label{fig:cost-at-thresholds-qwen3-30b}
\end{figure*}

\clearpage
\section{Misconfiguration Detection AUC Results}

We include both AUC and AUC @ FPR $<$ 1\% for our quantization and incorrect sampling detection with Qwen3-30B-A3B, which is our most challenging model. These values underlie the abstract's 'AUC $>$ 0.999' claims.

\begin{table}[h!]
  \centering
  \small
  \begin{tabular}{lrrrrrrrr}
    \toprule
    Batch Size & \multicolumn{2}{c}{Incorrect Seed} & \multicolumn{2}{c}{Temperature (1.0→1.1)} & \multicolumn{2}{c}{4-bit Model Quant.} & \multicolumn{2}{c}{FP8 KV Cache Quant.} \\
     & AUC & AUC $<$ FPR 1\% & AUC & AUC $<$ FPR 1\% & AUC & AUC $<$ FPR 1\% & AUC & AUC $<$ FPR 1\% \\
    \cmidrule(lr){2-3}\cmidrule(lr){4-5}\cmidrule(lr){6-7}\cmidrule(lr){8-9}
    \midrule
    1 & 0.5694 & 0.5314 & 0.5006 & 0.5000 & 0.5284 & 0.5124 & 0.5111 & 0.5045 \\
    3 & 0.6762 & 0.5493 & 0.5016 & 0.5000 & 0.5802 & 0.5160 & 0.5320 & 0.5051 \\
    10 & 0.8708 & 0.6737 & 0.5048 & 0.5000 & 0.7065 & 0.5474 & 0.5930 & 0.5105 \\
    30 & 0.9834 & 0.8668 & 0.5149 & 0.5000 & 0.8668 & 0.6067 & 0.6947 & 0.5178 \\
    100 & 0.9999 & 0.9976 & 0.5284 & 0.4994 & 0.9776 & 0.7915 & 0.8317 & 0.5372 \\
    300 & 1.0000 & 1.0000 & 0.5561 & 0.4977 & 0.9993 & 0.9768 & 0.9218 & 0.5866 \\
    1000 & 1.0000 & 1.0000 & 0.6235 & 0.4975 & 1.0000 & 1.0000 & 0.9786 & 0.7177 \\
    3000 & 1.0000 & 1.0000 & 0.7016 & 0.4975 & 1.0000 & 1.0000 & 0.9976 & 0.9302 \\
    10000 & 1.0000 & 1.0000 & 0.7688 & 0.4975 & 1.0000 & 1.0000 & 1.0000 & 1.0000 \\
    \bottomrule
  \end{tabular}
  \caption{AUC values for Token-DiFR score across different batch sizes and incorrect configurations for Qwen-3-30B-A3B.}
  \label{tab:auc_gumbel_margin_w99}
\end{table}

\begin{table}[h!]
  \centering
  \small
  \begin{tabular}{lrrrrrrrr}
    \toprule
    Batch Size & \multicolumn{2}{c}{Incorrect Seed} & \multicolumn{2}{c}{Temperature (1.0→1.1)} & \multicolumn{2}{c}{4-bit Model Quant.} & \multicolumn{2}{c}{FP8 KV Cache Quant.} \\
     & AUC & AUC $<$ FPR 1\% & AUC & AUC $<$ FPR 1\% & AUC & AUC $<$ FPR 1\% & AUC & AUC $<$ FPR 1\% \\
    \cmidrule(lr){2-3}\cmidrule(lr){4-5}\cmidrule(lr){6-7}\cmidrule(lr){8-9}
    \midrule
    1 & 0.8475 & 0.5075 & 0.8521 & 0.5076 & 0.9911 & 0.6658 & 0.9206 & 0.5115 \\
    2 & 0.8577 & 0.5074 & 0.8621 & 0.5076 & 0.9997 & 0.9854 & 0.9507 & 0.5706 \\
    4 & 0.8663 & 0.5072 & 0.8707 & 0.5075 & 1.0000 & 1.0000 & 0.9836 & 0.6999 \\
    8 & 0.8693 & 0.5068 & 0.8751 & 0.5075 & 1.0000 & 1.0000 & 0.9982 & 0.9294 \\
    16 & 0.8669 & 0.5060 & 0.8756 & 0.5074 & 1.0000 & 1.0000 & 1.0000 & 0.9984 \\
    \bottomrule
  \end{tabular}
  \caption{AUC values for Activation-DiFR score across different batch sizes and incorrect configurations for Qwen-3-30B-A3B.}
  \label{tab:auc_proj_k32_w99}
\end{table}

\clearpage
\section{Raw Score Distributions}
\label{app:raw_score_distributions}

This appendix provides comprehensive statistics of the raw score distributions for each model, configuration, GPU type, and metric. The tables report the count of samples, share of infinite values (where applicable), mean, standard deviation, and key percentiles (90th, 99th, 99.9th, 99.99th, and 99.999th). These statistics characterize the verification score distributions under both correct and incorrect configurations, providing insight into the separability of different misconfiguration types. All metrics are defined in Section~\ref{sec:token-difr-def}.

\clearpage

\begin{table*}[p]
\centering
{\large\textbf{Token-DiFR (margin) raw score distributions}}

\vspace{0.5em}

\scriptsize
\setlength{\tabcolsep}{3pt}
\begin{tabular}{lllllrrrrrrr}
\toprule
Model & Config & GPU & Count & $\infty$ & Mean & Std & p90 & p99 & p99.9 & p99.99 & p99.999 \\
\midrule
Qwen3-30B-A3B & reference & a100 & 629294 & 0.0011 & 0.02 & 0.25 & 0.00 & 0.61 & 4.02 & 6.53 & 9.13 \\
Qwen3-30B-A3B & reference & h200 & 627756 & 0.0001 & 0.00 & 0.02 & 0.00 & 0.00 & 0.01 & 0.91 & 2.42 \\
Qwen3-30B-A3B & correct\_tp4 & a100 & 629072 & 0.0012 & 0.00 & 0.06 & 0.00 & 0.10 & 0.75 & 2.58 & 4.97 \\
Qwen3-30B-A3B & correct\_tp4 & h200 & 630195 & 0.0011 & 0.00 & 0.06 & 0.00 & 0.10 & 0.72 & 2.47 & 4.75 \\
Qwen3-30B-A3B & correct\_hf & h200 & 806271 & 0.0024 & 0.06 & 0.41 & 0.00 & 2.26 & 5.24 & 7.48 & 9.52 \\
Qwen3-30B-A3B & incorrect\_adv\_kv\_fp8\_temp\_0\_96 & h200 & 604449 & 0.0031 & 0.02 & 0.14 & 0.00 & 0.56 & 1.97 & 3.94 & 5.89 \\
Qwen3-30B-A3B & incorrect\_config\_random\_bug\_k\_2 & h200 & 625908 & 0.0031 & 0.00 & 0.12 & 0.00 & 0.00 & 1.86 & 5.03 & 7.64 \\
Qwen3-30B-A3B & incorrect\_config\_random\_bug\_k\_32 & h200 & 621929 & 0.0094 & 0.00 & 0.06 & 0.00 & 0.00 & 0.12 & 3.15 & 6.79 \\
Qwen3-30B-A3B & incorrect\_kv\_cache\_quant\_fp8 & h200 & 603591 & 0.0040 & 0.02 & 0.13 & 0.00 & 0.52 & 1.80 & 3.62 & 6.04 \\
Qwen3-30B-A3B & incorrect\_model\_quant\_4bit & h200 & 559981 & 0.0114 & 0.05 & 0.26 & 0.00 & 1.35 & 3.07 & 4.88 & 7.88 \\
Qwen3-30B-A3B & incorrect\_model\_quant\_fp8 & h200 & 661475 & 0.0027 & 0.01 & 0.11 & 0.00 & 0.41 & 1.58 & 3.47 & 5.12 \\
Qwen3-30B-A3B & incorrect\_seed\_43 & h200 & 665261 & 0.0001 & 0.27 & 0.85 & 0.93 & 4.27 & 6.70 & 9.05 & 11.19 \\
Qwen3-30B-A3B & incorrect\_temperature\_1\_1 & h200 & 615796 & 0.0040 & 0.00 & 0.02 & 0.00 & 0.00 & 0.21 & 0.51 & 1.42 \\
Qwen3-30B-A3B & incorrect\_top\_p\_0\_85 & h200 & 628161 & 0.0000 & 0.02 & 0.21 & 0.00 & 0.76 & 3.13 & 5.45 & 8.10 \\
Qwen3-8B & reference & a100 & 786831 & 0.0005 & 0.00 & 0.04 & 0.00 & 0.00 & 0.22 & 2.05 & 4.49 \\
Qwen3-8B & reference & h200 & 784915 & 0.0005 & 0.00 & 0.04 & 0.00 & 0.00 & 0.21 & 1.89 & 4.28 \\
Qwen3-8B & correct\_tp4 & a100 & 788186 & 0.0006 & 0.00 & 0.04 & 0.00 & 0.00 & 0.23 & 1.94 & 4.28 \\
Qwen3-8B & correct\_tp4 & h200 & 785547 & 0.0005 & 0.00 & 0.04 & 0.00 & 0.00 & 0.23 & 1.97 & 4.50 \\
Qwen3-8B & correct\_hf & h200 & 783799 & 0.0006 & 0.00 & 0.04 & 0.00 & 0.00 & 0.21 & 1.82 & 4.38 \\
Qwen3-8B & incorrect\_adv\_kv\_fp8\_temp\_0\_96 & h200 & 774505 & 0.0018 & 0.01 & 0.11 & 0.00 & 0.37 & 1.63 & 3.83 & 5.75 \\
Qwen3-8B & incorrect\_adv\_model\_4bit\_temp\_0\_89 & h200 & 708037 & 0.0051 & 0.06 & 0.30 & 0.00 & 1.55 & 3.53 & 5.54 & 7.59 \\
Qwen3-8B & incorrect\_config\_random\_bug\_k\_2 & h200 & 783008 & 0.0029 & 0.01 & 0.14 & 0.00 & 0.01 & 2.37 & 5.13 & 7.47 \\
Qwen3-8B & incorrect\_config\_random\_bug\_k\_32 & h200 & 777720 & 0.0097 & 0.00 & 0.08 & 0.00 & 0.00 & 0.28 & 4.28 & 6.84 \\
Qwen3-8B & incorrect\_kv\_cache\_quant\_fp8 & h200 & 764520 & 0.0028 & 0.01 & 0.10 & 0.00 & 0.34 & 1.33 & 3.58 & 5.80 \\
Qwen3-8B & incorrect\_model\_quant\_4bit & h200 & 710621 & 0.0085 & 0.05 & 0.27 & 0.00 & 1.36 & 3.11 & 5.25 & 7.09 \\
Qwen3-8B & incorrect\_model\_quant\_fp8 & h200 & 778203 & 0.0021 & 0.01 & 0.09 & 0.00 & 0.30 & 1.24 & 3.38 & 5.53 \\
Qwen3-8B & incorrect\_seed\_43 & h200 & 738399 & 0.0005 & 0.38 & 1.01 & 1.56 & 4.77 & 7.29 & 9.71 & 11.87 \\
Qwen3-8B & incorrect\_temperature\_1\_1 & h200 & 781435 & 0.0059 & 0.00 & 0.02 & 0.00 & 0.06 & 0.28 & 0.43 & 1.28 \\
Qwen3-8B & incorrect\_top\_p\_0\_85 & h200 & 783913 & 0.0000 & 0.03 & 0.24 & 0.00 & 1.04 & 3.36 & 5.64 & 8.35 \\
Llama-3.1-8B & reference & a100 & 398030 & 0.0004 & 0.00 & 0.04 & 0.00 & 0.00 & 0.12 & 1.77 & 4.33 \\
Llama-3.1-8B & reference & h200 & 400206 & 0.0004 & 0.00 & 0.04 & 0.00 & 0.00 & 0.12 & 1.93 & 4.28 \\
Llama-3.1-8B & correct\_tp4 & a100 & 401402 & 0.0005 & 0.00 & 0.04 & 0.00 & 0.00 & 0.15 & 1.91 & 4.53 \\
Llama-3.1-8B & correct\_tp4 & h200 & 402934 & 0.0005 & 0.00 & 0.04 & 0.00 & 0.00 & 0.14 & 1.83 & 3.97 \\
Llama-3.1-8B & correct\_hf & h200 & 398700 & 0.0005 & 0.00 & 0.03 & 0.00 & 0.00 & 0.12 & 1.64 & 3.57 \\
Llama-3.1-8B & incorrect\_adv\_kv\_fp8\_temp\_0\_96 & h200 & 399515 & 0.0016 & 0.01 & 0.10 & 0.00 & 0.28 & 1.48 & 3.47 & 5.05 \\
Llama-3.1-8B & incorrect\_adv\_model\_4bit\_temp\_0\_89 & h200 & 382343 & 0.0039 & 0.05 & 0.25 & 0.00 & 1.23 & 3.12 & 5.32 & 7.28 \\
Llama-3.1-8B & incorrect\_config\_random\_bug\_k\_2 & h200 & 399210 & 0.0028 & 0.01 & 0.14 & 0.00 & 0.00 & 2.45 & 5.11 & 6.62 \\
Llama-3.1-8B & incorrect\_config\_random\_bug\_k\_32 & h200 & 396691 & 0.0091 & 0.00 & 0.13 & 0.00 & 0.00 & 1.39 & 6.27 & 7.57 \\
Llama-3.1-8B & incorrect\_kv\_cache\_quant\_fp8 & h200 & 399409 & 0.0029 & 0.01 & 0.08 & 0.00 & 0.24 & 1.11 & 3.01 & 4.91 \\
Llama-3.1-8B & incorrect\_model\_quant\_4bit & h200 & 381999 & 0.0092 & 0.04 & 0.21 & 0.00 & 1.04 & 2.69 & 4.55 & 5.91 \\
Llama-3.1-8B & incorrect\_model\_quant\_fp8 & h200 & 399564 & 0.0025 & 0.01 & 0.08 & 0.00 & 0.24 & 1.03 & 3.24 & 5.33 \\
Llama-3.1-8B & incorrect\_seed\_43 & h200 & 392316 & 0.0004 & 0.51 & 1.23 & 2.16 & 5.58 & 8.02 & 10.23 & 11.84 \\
Llama-3.1-8B & incorrect\_temperature\_1\_1 & h200 & 388448 & 0.0085 & 0.00 & 0.02 & 0.00 & 0.09 & 0.30 & 0.45 & 1.45 \\
Llama-3.1-8B & incorrect\_top\_p\_0\_85 & h200 & 408071 & 0.0000 & 0.03 & 0.25 & 0.00 & 1.12 & 3.48 & 5.75 & 8.27 \\
\bottomrule
\end{tabular}
\caption{Summary statistics for the sampled token margin between the verifier's most probable token and the claimed token. The $\infty$ column indicates the fraction of tokens where the claimed token was filtered out by top-$k$ or top-$p$ sampling.}
\label{tab:raw_scores_margin}
\end{table*}

\clearpage

\begin{table*}[p]
\centering
{\large\textbf{Cross-entropy raw score distributions}}

\vspace{0.5em}

\scriptsize
\setlength{\tabcolsep}{3pt}
\begin{tabular}{lllllrrrrrrr}
\toprule
Model & Config & GPU & Count & $\infty$ & Mean & Std & p90 & p99 & p99.9 & p99.99 & p99.999 \\
\midrule
Qwen3-30B-A3B & reference & a100 & 629294 & 0.0011 & 0.27 & 0.58 & 0.97 & 2.81 & 3.94 & 4.97 & 5.45 \\
Qwen3-30B-A3B & reference & h200 & 627756 & 0.0001 & 0.27 & 0.59 & 0.97 & 2.81 & 3.97 & 5.05 & 5.50 \\
Qwen3-30B-A3B & correct\_tp4 & a100 & 629072 & 0.0012 & 0.27 & 0.59 & 0.97 & 2.81 & 3.98 & 4.96 & 5.41 \\
Qwen3-30B-A3B & correct\_tp4 & h200 & 630195 & 0.0011 & 0.27 & 0.58 & 0.97 & 2.81 & 3.93 & 4.89 & 5.38 \\
Qwen3-30B-A3B & correct\_hf & h200 & 806271 & 0.0024 & 0.27 & 0.59 & 0.96 & 2.81 & 4.06 & 5.09 & 5.50 \\
Qwen3-30B-A3B & incorrect\_adv\_kv\_fp8\_temp\_0\_96 & h200 & 604449 & 0.0031 & 0.27 & 0.58 & 0.97 & 2.79 & 3.89 & 4.80 & 5.40 \\
Qwen3-30B-A3B & incorrect\_config\_random\_bug\_k\_2 & h200 & 625908 & 0.0031 & 0.28 & 0.59 & 0.97 & 2.81 & 3.98 & 5.05 & 5.50 \\
Qwen3-30B-A3B & incorrect\_config\_random\_bug\_k\_32 & h200 & 621929 & 0.0094 & 0.28 & 0.59 & 0.97 & 2.81 & 4.00 & 5.07 & 5.52 \\
Qwen3-30B-A3B & incorrect\_kv\_cache\_quant\_fp8 & h200 & 603591 & 0.0040 & 0.28 & 0.60 & 0.97 & 2.83 & 3.98 & 4.97 & 5.47 \\
Qwen3-30B-A3B & incorrect\_model\_quant\_4bit & h200 & 559981 & 0.0114 & 0.30 & 0.64 & 1.12 & 2.97 & 4.09 & 5.08 & 5.51 \\
Qwen3-30B-A3B & incorrect\_model\_quant\_fp8 & h200 & 661475 & 0.0027 & 0.27 & 0.59 & 0.97 & 2.81 & 3.92 & 4.88 & 5.45 \\
Qwen3-30B-A3B & incorrect\_seed\_43 & h200 & 665261 & 0.0001 & 0.27 & 0.59 & 0.97 & 2.81 & 4.01 & 4.99 & 5.52 \\
Qwen3-30B-A3B & incorrect\_temperature\_1\_1 & h200 & 615796 & 0.0040 & 0.29 & 0.62 & 1.02 & 2.94 & 4.14 & 5.09 & 5.46 \\
Qwen3-30B-A3B & incorrect\_top\_p\_0\_85 & h200 & 628161 & 0.0000 & 0.23 & 0.46 & 0.83 & 2.12 & 3.01 & 3.92 & 4.40 \\
Qwen3-8B & reference & a100 & 786831 & 0.0005 & 0.38 & 0.69 & 1.30 & 3.17 & 4.35 & 5.21 & 5.53 \\
Qwen3-8B & reference & h200 & 784915 & 0.0005 & 0.38 & 0.69 & 1.30 & 3.17 & 4.37 & 5.23 & 5.57 \\
Qwen3-8B & correct\_tp4 & a100 & 788186 & 0.0006 & 0.38 & 0.70 & 1.30 & 3.17 & 4.36 & 5.15 & 5.52 \\
Qwen3-8B & correct\_tp4 & h200 & 785547 & 0.0005 & 0.37 & 0.69 & 1.29 & 3.17 & 4.33 & 5.15 & 5.52 \\
Qwen3-8B & correct\_hf & h200 & 783799 & 0.0006 & 0.38 & 0.70 & 1.30 & 3.17 & 4.36 & 5.23 & 5.54 \\
Qwen3-8B & incorrect\_adv\_kv\_fp8\_temp\_0\_96 & h200 & 774505 & 0.0018 & 0.37 & 0.68 & 1.26 & 3.08 & 4.26 & 5.05 & 5.41 \\
Qwen3-8B & incorrect\_adv\_model\_4bit\_temp\_0\_89 & h200 & 708037 & 0.0051 & 0.37 & 0.67 & 1.26 & 3.05 & 4.20 & 5.05 & 5.55 \\
Qwen3-8B & incorrect\_config\_random\_bug\_k\_2 & h200 & 783008 & 0.0029 & 0.38 & 0.70 & 1.31 & 3.17 & 4.37 & 5.23 & 5.57 \\
Qwen3-8B & incorrect\_config\_random\_bug\_k\_32 & h200 & 777720 & 0.0097 & 0.38 & 0.70 & 1.30 & 3.18 & 4.39 & 5.23 & 5.57 \\
Qwen3-8B & incorrect\_kv\_cache\_quant\_fp8 & h200 & 764520 & 0.0028 & 0.38 & 0.70 & 1.31 & 3.17 & 4.33 & 5.18 & 5.53 \\
Qwen3-8B & incorrect\_model\_quant\_4bit & h200 & 710621 & 0.0085 & 0.40 & 0.72 & 1.37 & 3.24 & 4.42 & 5.17 & 5.55 \\
Qwen3-8B & incorrect\_model\_quant\_fp8 & h200 & 778203 & 0.0021 & 0.38 & 0.69 & 1.30 & 3.14 & 4.34 & 5.22 & 5.51 \\
Qwen3-8B & incorrect\_seed\_43 & h200 & 738399 & 0.0005 & 0.38 & 0.70 & 1.31 & 3.20 & 4.44 & 5.25 & 5.58 \\
Qwen3-8B & incorrect\_temperature\_1\_1 & h200 & 781435 & 0.0059 & 0.40 & 0.74 & 1.41 & 3.32 & 4.50 & 5.25 & 5.59 \\
Qwen3-8B & incorrect\_top\_p\_0\_85 & h200 & 783913 & 0.0000 & 0.31 & 0.55 & 1.09 & 2.43 & 3.38 & 4.20 & 4.63 \\
Llama-3.1-8B & reference & a100 & 398030 & 0.0004 & 0.51 & 0.93 & 1.80 & 4.22 & 5.26 & 5.60 & 5.74 \\
Llama-3.1-8B & reference & h200 & 400206 & 0.0004 & 0.50 & 0.92 & 1.77 & 4.22 & 5.28 & 5.61 & 5.76 \\
Llama-3.1-8B & correct\_tp4 & a100 & 401402 & 0.0005 & 0.51 & 0.93 & 1.80 & 4.23 & 5.28 & 5.61 & 5.76 \\
Llama-3.1-8B & correct\_tp4 & h200 & 402934 & 0.0005 & 0.51 & 0.93 & 1.80 & 4.24 & 5.27 & 5.61 & 5.79 \\
Llama-3.1-8B & correct\_hf & h200 & 398700 & 0.0005 & 0.51 & 0.93 & 1.79 & 4.23 & 5.29 & 5.61 & 5.81 \\
Llama-3.1-8B & incorrect\_adv\_kv\_fp8\_temp\_0\_96 & h200 & 399515 & 0.0016 & 0.48 & 0.89 & 1.70 & 4.05 & 5.15 & 5.54 & 5.75 \\
Llama-3.1-8B & incorrect\_adv\_model\_4bit\_temp\_0\_89 & h200 & 382343 & 0.0039 & 0.46 & 0.85 & 1.61 & 3.86 & 5.06 & 5.50 & 5.68 \\
Llama-3.1-8B & incorrect\_config\_random\_bug\_k\_2 & h200 & 399210 & 0.0028 & 0.50 & 0.93 & 1.78 & 4.21 & 5.28 & 5.61 & 5.76 \\
Llama-3.1-8B & incorrect\_config\_random\_bug\_k\_32 & h200 & 396691 & 0.0091 & 0.51 & 0.93 & 1.78 & 4.25 & 5.30 & 5.62 & 5.76 \\
Llama-3.1-8B & incorrect\_kv\_cache\_quant\_fp8 & h200 & 399409 & 0.0029 & 0.53 & 0.95 & 1.86 & 4.27 & 5.28 & 5.62 & 5.79 \\
Llama-3.1-8B & incorrect\_model\_quant\_4bit & h200 & 381999 & 0.0092 & 0.56 & 0.98 & 1.98 & 4.34 & 5.29 & 5.61 & 5.81 \\
Llama-3.1-8B & incorrect\_model\_quant\_fp8 & h200 & 399564 & 0.0025 & 0.52 & 0.94 & 1.82 & 4.24 & 5.26 & 5.61 & 5.76 \\
Llama-3.1-8B & incorrect\_seed\_43 & h200 & 392316 & 0.0004 & 0.51 & 0.94 & 1.80 & 4.28 & 5.29 & 5.63 & 5.83 \\
Llama-3.1-8B & incorrect\_temperature\_1\_1 & h200 & 388448 & 0.0085 & 0.62 & 1.07 & 2.21 & 4.61 & 5.42 & 5.70 & 5.89 \\
Llama-3.1-8B & incorrect\_top\_p\_0\_85 & h200 & 408071 & 0.0000 & 0.37 & 0.70 & 1.34 & 3.21 & 4.25 & 4.64 & 4.82 \\
\bottomrule
\end{tabular}
\caption{Summary statistics for the negative log-likelihood of the claimed token under the verifier's softmax distribution. The $\infty$ column indicates the fraction of tokens where either the probability underflowed to 0 before taking the log, or the token was filtered out by top-$k$ or top-$p$ sampling.}
\label{tab:raw_scores_cross_entropy}
\end{table*}

\clearpage

\begin{table*}[p]
\centering
{\large\textbf{Token-DiFR (exact match) raw score distributions}}

\vspace{0.5em}

\scriptsize
\setlength{\tabcolsep}{3pt}
\begin{tabular}{llllrrrrrrrr}
\toprule
Model & Config & GPU & Count & Mean & Std & p90 & p99 & p99.9 & p99.99 & p99.999 \\
\midrule
Qwen3-30B-A3B & reference & a100 & 630001 & 0.97 & 0.16 & 1.00 & 1.00 & 1.00 & 1.00 & 1.00 \\
Qwen3-30B-A3B & reference & h200 & 627824 & 1.00 & 0.03 & 1.00 & 1.00 & 1.00 & 1.00 & 1.00 \\
Qwen3-30B-A3B & correct\_tp4 & a100 & 629846 & 0.98 & 0.13 & 1.00 & 1.00 & 1.00 & 1.00 & 1.00 \\
Qwen3-30B-A3B & correct\_tp4 & h200 & 630916 & 0.98 & 0.13 & 1.00 & 1.00 & 1.00 & 1.00 & 1.00 \\
Qwen3-30B-A3B & correct\_hf & h200 & 808241 & 0.95 & 0.21 & 1.00 & 1.00 & 1.00 & 1.00 & 1.00 \\
Qwen3-30B-A3B & incorrect\_adv\_kv\_fp8\_temp\_0\_96 & h200 & 606314 & 0.96 & 0.19 & 1.00 & 1.00 & 1.00 & 1.00 & 1.00 \\
Qwen3-30B-A3B & incorrect\_config\_random\_bug\_k\_2 & h200 & 627824 & 0.99 & 0.08 & 1.00 & 1.00 & 1.00 & 1.00 & 1.00 \\
Qwen3-30B-A3B & incorrect\_config\_random\_bug\_k\_32 & h200 & 627824 & 0.99 & 0.10 & 1.00 & 1.00 & 1.00 & 1.00 & 1.00 \\
Qwen3-30B-A3B & incorrect\_kv\_cache\_quant\_fp8 & h200 & 606023 & 0.96 & 0.19 & 1.00 & 1.00 & 1.00 & 1.00 & 1.00 \\
Qwen3-30B-A3B & incorrect\_model\_quant\_4bit & h200 & 566434 & 0.93 & 0.26 & 1.00 & 1.00 & 1.00 & 1.00 & 1.00 \\
Qwen3-30B-A3B & incorrect\_model\_quant\_fp8 & h200 & 663284 & 0.97 & 0.17 & 1.00 & 1.00 & 1.00 & 1.00 & 1.00 \\
Qwen3-30B-A3B & incorrect\_seed\_43 & h200 & 665315 & 0.85 & 0.36 & 1.00 & 1.00 & 1.00 & 1.00 & 1.00 \\
Qwen3-30B-A3B & incorrect\_temperature\_1\_1 & h200 & 618265 & 0.99 & 0.12 & 1.00 & 1.00 & 1.00 & 1.00 & 1.00 \\
Qwen3-30B-A3B & incorrect\_top\_p\_0\_85 & h200 & 628170 & 0.98 & 0.14 & 1.00 & 1.00 & 1.00 & 1.00 & 1.00 \\
Qwen3-8B & reference & a100 & 787234 & 0.99 & 0.09 & 1.00 & 1.00 & 1.00 & 1.00 & 1.00 \\
Qwen3-8B & reference & h200 & 785304 & 0.99 & 0.09 & 1.00 & 1.00 & 1.00 & 1.00 & 1.00 \\
Qwen3-8B & correct\_tp4 & a100 & 788648 & 0.99 & 0.09 & 1.00 & 1.00 & 1.00 & 1.00 & 1.00 \\
Qwen3-8B & correct\_tp4 & h200 & 785979 & 0.99 & 0.09 & 1.00 & 1.00 & 1.00 & 1.00 & 1.00 \\
Qwen3-8B & correct\_hf & h200 & 784280 & 0.99 & 0.09 & 1.00 & 1.00 & 1.00 & 1.00 & 1.00 \\
Qwen3-8B & incorrect\_adv\_kv\_fp8\_temp\_0\_96 & h200 & 775897 & 0.97 & 0.18 & 1.00 & 1.00 & 1.00 & 1.00 & 1.00 \\
Qwen3-8B & incorrect\_adv\_model\_4bit\_temp\_0\_89 & h200 & 711666 & 0.92 & 0.28 & 1.00 & 1.00 & 1.00 & 1.00 & 1.00 \\
Qwen3-8B & incorrect\_config\_random\_bug\_k\_2 & h200 & 785304 & 0.99 & 0.11 & 1.00 & 1.00 & 1.00 & 1.00 & 1.00 \\
Qwen3-8B & incorrect\_config\_random\_bug\_k\_32 & h200 & 785304 & 0.98 & 0.13 & 1.00 & 1.00 & 1.00 & 1.00 & 1.00 \\
Qwen3-8B & incorrect\_kv\_cache\_quant\_fp8 & h200 & 766667 & 0.97 & 0.18 & 1.00 & 1.00 & 1.00 & 1.00 & 1.00 \\
Qwen3-8B & incorrect\_model\_quant\_4bit & h200 & 716701 & 0.92 & 0.27 & 1.00 & 1.00 & 1.00 & 1.00 & 1.00 \\
Qwen3-8B & incorrect\_model\_quant\_fp8 & h200 & 779878 & 0.97 & 0.17 & 1.00 & 1.00 & 1.00 & 1.00 & 1.00 \\
Qwen3-8B & incorrect\_seed\_43 & h200 & 738774 & 0.79 & 0.40 & 1.00 & 1.00 & 1.00 & 1.00 & 1.00 \\
Qwen3-8B & incorrect\_temperature\_1\_1 & h200 & 786035 & 0.98 & 0.14 & 1.00 & 1.00 & 1.00 & 1.00 & 1.00 \\
Qwen3-8B & incorrect\_top\_p\_0\_85 & h200 & 783913 & 0.97 & 0.18 & 1.00 & 1.00 & 1.00 & 1.00 & 1.00 \\
Llama-3.1-8B & reference & a100 & 398185 & 0.99 & 0.08 & 1.00 & 1.00 & 1.00 & 1.00 & 1.00 \\
Llama-3.1-8B & reference & h200 & 400348 & 0.99 & 0.08 & 1.00 & 1.00 & 1.00 & 1.00 & 1.00 \\
Llama-3.1-8B & correct\_tp4 & a100 & 401598 & 0.99 & 0.09 & 1.00 & 1.00 & 1.00 & 1.00 & 1.00 \\
Llama-3.1-8B & correct\_tp4 & h200 & 403141 & 0.99 & 0.09 & 1.00 & 1.00 & 1.00 & 1.00 & 1.00 \\
Llama-3.1-8B & correct\_hf & h200 & 398902 & 0.99 & 0.08 & 1.00 & 1.00 & 1.00 & 1.00 & 1.00 \\
Llama-3.1-8B & incorrect\_adv\_kv\_fp8\_temp\_0\_96 & h200 & 400168 & 0.97 & 0.18 & 1.00 & 1.00 & 1.00 & 1.00 & 1.00 \\
Llama-3.1-8B & incorrect\_adv\_model\_4bit\_temp\_0\_89 & h200 & 383836 & 0.93 & 0.26 & 1.00 & 1.00 & 1.00 & 1.00 & 1.00 \\
Llama-3.1-8B & incorrect\_config\_random\_bug\_k\_2 & h200 & 400348 & 0.99 & 0.11 & 1.00 & 1.00 & 1.00 & 1.00 & 1.00 \\
Llama-3.1-8B & incorrect\_config\_random\_bug\_k\_32 & h200 & 400348 & 0.98 & 0.13 & 1.00 & 1.00 & 1.00 & 1.00 & 1.00 \\
Llama-3.1-8B & incorrect\_kv\_cache\_quant\_fp8 & h200 & 400573 & 0.97 & 0.17 & 1.00 & 1.00 & 1.00 & 1.00 & 1.00 \\
Llama-3.1-8B & incorrect\_model\_quant\_4bit & h200 & 385544 & 0.92 & 0.26 & 1.00 & 1.00 & 1.00 & 1.00 & 1.00 \\
Llama-3.1-8B & incorrect\_model\_quant\_fp8 & h200 & 400549 & 0.97 & 0.17 & 1.00 & 1.00 & 1.00 & 1.00 & 1.00 \\
Llama-3.1-8B & incorrect\_seed\_43 & h200 & 392476 & 0.76 & 0.43 & 1.00 & 1.00 & 1.00 & 1.00 & 1.00 \\
Llama-3.1-8B & incorrect\_temperature\_1\_1 & h200 & 391774 & 0.97 & 0.16 & 1.00 & 1.00 & 1.00 & 1.00 & 1.00 \\
Llama-3.1-8B & incorrect\_top\_p\_0\_85 & h200 & 408071 & 0.97 & 0.18 & 1.00 & 1.00 & 1.00 & 1.00 & 1.00 \\
\bottomrule
\end{tabular}
\caption{Summary statistics for the binary exact-match rate: the fraction of tokens where the provider and verifier sampled the same token under a shared seed. Since this is a binary metric (0 or 1), there are no infinite values.}
\label{tab:raw_scores_exact_match}
\end{table*}

\clearpage

\begin{table*}[p]
\centering
{\large\textbf{Token-DiFR (likelihood, $\sigma=0.02$) raw score distributions}}

\vspace{0.5em}

\scriptsize
\setlength{\tabcolsep}{3pt}
\begin{tabular}{lllllrrrrrrr}
\toprule
Model & Config & GPU & Count & $\infty$ & Mean & Std & p90 & p99 & p99.9 & p99.99 & p99.999 \\
\midrule
Qwen3-30B-A3B & reference & a100 & 629294 & 0.0011 & 40.96 & 710.36 & 0.00 & 237.01 & 10127.86 & 26636.31 & 52152.83 \\
Qwen3-30B-A3B & reference & h200 & 627756 & 0.0001 & 0.27 & 38.23 & 0.00 & 0.00 & 0.99 & 522.25 & 3672.64 \\
Qwen3-30B-A3B & correct\_tp4 & a100 & 629072 & 0.0012 & 2.48 & 102.66 & 0.00 & 8.91 & 356.07 & 4155.61 & 15459.88 \\
Qwen3-30B-A3B & correct\_tp4 & h200 & 630195 & 0.0011 & 2.28 & 105.55 & 0.00 & 8.39 & 327.01 & 3809.94 & 14106.97 \\
Qwen3-30B-A3B & correct\_hf & h200 & 806271 & 0.0024 & 109.36 & 1133.59 & 0.00 & 3203.67 & 17191.44 & 35016.00 & 56656.67 \\
Qwen3-30B-A3B & incorrect\_adv\_kv\_fp8\_temp\_0\_96 & h200 & 604449 & 0.0031 & 12.55 & 232.32 & 0.00 & 198.55 & 2420.33 & 9693.36 & 21700.69 \\
Qwen3-30B-A3B & incorrect\_config\_random\_bug\_k\_2 & h200 & 625908 & 0.0031 & 8.76 & 329.24 & 0.00 & 0.00 & 2159.98 & 15818.20 & 36520.42 \\
Qwen3-30B-A3B & incorrect\_config\_random\_bug\_k\_32 & h200 & 621929 & 0.0094 & 2.20 & 195.42 & 0.00 & 0.00 & 10.77 & 6225.48 & 28781.38 \\
Qwen3-30B-A3B & incorrect\_kv\_cache\_quant\_fp8 & h200 & 603591 & 0.0040 & 10.87 & 220.33 & 0.00 & 172.01 & 2039.49 & 8209.02 & 22831.76 \\
Qwen3-30B-A3B & incorrect\_model\_quant\_4bit & h200 & 559981 & 0.0114 & 43.55 & 447.41 & 0.00 & 1146.94 & 5897.47 & 14895.77 & 38808.91 \\
Qwen3-30B-A3B & incorrect\_model\_quant\_fp8 & h200 & 661475 & 0.0027 & 8.04 & 168.83 & 0.00 & 108.13 & 1564.10 & 7550.36 & 16418.96 \\
Qwen3-30B-A3B & incorrect\_seed\_43 & h200 & 665261 & 0.0001 & 498.82 & 2353.89 & 547.69 & 11376.17 & 28051.18 & 51245.66 & 78249.20 \\
Qwen3-30B-A3B & incorrect\_temperature\_1\_1 & h200 & 615796 & 0.0040 & 0.19 & 12.25 & 0.00 & 0.00 & 30.15 & 168.89 & 1268.61 \\
Qwen3-30B-A3B & incorrect\_top\_p\_0\_85 & h200 & 628161 & 0.0000 & 27.66 & 468.32 & 0.00 & 362.01 & 6146.39 & 18599.42 & 41019.43 \\
Qwen3-8B & reference & a100 & 786831 & 0.0005 & 1.05 & 94.81 & 0.00 & 0.00 & 34.12 & 2628.20 & 12586.91 \\
Qwen3-8B & reference & h200 & 784915 & 0.0005 & 0.92 & 73.40 & 0.00 & 0.00 & 30.30 & 2233.97 & 11444.36 \\
Qwen3-8B & correct\_tp4 & a100 & 788186 & 0.0006 & 0.99 & 75.30 & 0.00 & 0.00 & 35.06 & 2358.73 & 11468.82 \\
Qwen3-8B & correct\_tp4 & h200 & 785547 & 0.0005 & 1.10 & 87.74 & 0.00 & 0.00 & 35.17 & 2440.59 & 12695.14 \\
Qwen3-8B & correct\_hf & h200 & 783799 & 0.0006 & 0.85 & 76.01 & 0.00 & 0.00 & 30.88 & 2077.44 & 11973.31 \\
Qwen3-8B & incorrect\_adv\_kv\_fp8\_temp\_0\_96 & h200 & 774505 & 0.0018 & 8.32 & 203.76 & 0.00 & 89.68 & 1663.93 & 9181.77 & 20706.42 \\
Qwen3-8B & incorrect\_adv\_model\_4bit\_temp\_0\_89 & h200 & 708037 & 0.0051 & 59.88 & 554.57 & 0.00 & 1500.82 & 7794.26 & 19175.61 & 36001.17 \\
Qwen3-8B & incorrect\_config\_random\_bug\_k\_2 & h200 & 783008 & 0.0029 & 11.68 & 382.62 & 0.00 & 0.97 & 3519.33 & 16425.68 & 34912.87 \\
Qwen3-8B & incorrect\_config\_random\_bug\_k\_32 & h200 & 777720 & 0.0097 & 4.01 & 249.74 & 0.00 & 0.00 & 50.61 & 11466.60 & 29261.41 \\
Qwen3-8B & incorrect\_kv\_cache\_quant\_fp8 & h200 & 764520 & 0.0028 & 6.42 & 174.56 & 0.00 & 74.54 & 1117.25 & 8024.11 & 21036.45 \\
Qwen3-8B & incorrect\_model\_quant\_4bit & h200 & 710621 & 0.0085 & 46.14 & 462.96 & 0.00 & 1155.74 & 6061.66 & 17207.19 & 31459.71 \\
Qwen3-8B & incorrect\_model\_quant\_fp8 & h200 & 778203 & 0.0021 & 5.66 & 164.58 & 0.00 & 60.84 & 962.45 & 7164.61 & 19103.23 \\
Qwen3-8B & incorrect\_seed\_43 & h200 & 738399 & 0.0005 & 727.44 & 2930.33 & 1519.65 & 14255.76 & 33201.36 & 58969.98 & 88083.83 \\
Qwen3-8B & incorrect\_temperature\_1\_1 & h200 & 781435 & 0.0059 & 0.29 & 21.52 & 0.00 & 3.71 & 51.78 & 117.61 & 1029.91 \\
Qwen3-8B & incorrect\_top\_p\_0\_85 & h200 & 783913 & 0.0000 & 36.04 & 531.94 & 0.00 & 683.94 & 7069.84 & 19913.24 & 43582.39 \\
Llama-3.1-8B & reference & a100 & 398030 & 0.0004 & 0.89 & 81.33 & 0.00 & 0.00 & 11.80 & 1956.56 & 11729.23 \\
Llama-3.1-8B & reference & h200 & 400206 & 0.0004 & 0.82 & 78.04 & 0.00 & 0.00 & 11.48 & 2343.55 & 11459.91 \\
Llama-3.1-8B & correct\_tp4 & a100 & 401402 & 0.0005 & 0.94 & 73.49 & 0.00 & 0.00 & 17.54 & 2283.42 & 12815.66 \\
Llama-3.1-8B & correct\_tp4 & h200 & 402934 & 0.0005 & 0.84 & 62.90 & 0.00 & 0.00 & 15.22 & 2108.15 & 9845.70 \\
Llama-3.1-8B & correct\_hf & h200 & 398700 & 0.0005 & 0.60 & 57.33 & 0.00 & 0.00 & 11.61 & 1687.17 & 7990.34 \\
Llama-3.1-8B & incorrect\_adv\_kv\_fp8\_temp\_0\_96 & h200 & 399515 & 0.0016 & 6.46 & 159.49 & 0.00 & 53.33 & 1376.86 & 7515.03 & 15938.51 \\
Llama-3.1-8B & incorrect\_adv\_model\_4bit\_temp\_0\_89 & h200 & 382343 & 0.0039 & 41.20 & 459.77 & 0.00 & 943.38 & 6086.98 & 17721.77 & 33091.78 \\
Llama-3.1-8B & incorrect\_config\_random\_bug\_k\_2 & h200 & 399210 & 0.0028 & 11.82 & 347.68 & 0.00 & 0.00 & 3769.77 & 16308.28 & 27385.34 \\
Llama-3.1-8B & incorrect\_config\_random\_bug\_k\_32 & h200 & 396691 & 0.0091 & 11.24 & 458.01 & 0.00 & 0.00 & 1217.70 & 24609.92 & 35827.70 \\
Llama-3.1-8B & incorrect\_kv\_cache\_quant\_fp8 & h200 & 399409 & 0.0029 & 4.38 & 128.38 & 0.00 & 39.21 & 772.30 & 5650.39 & 15063.85 \\
Llama-3.1-8B & incorrect\_model\_quant\_4bit & h200 & 381999 & 0.0092 & 29.73 & 338.04 & 0.00 & 683.93 & 4522.18 & 12930.92 & 21827.17 \\
Llama-3.1-8B & incorrect\_model\_quant\_fp8 & h200 & 399564 & 0.0025 & 4.34 & 146.48 & 0.00 & 38.17 & 673.27 & 6565.22 & 17794.56 \\
Llama-3.1-8B & incorrect\_seed\_43 & h200 & 392316 & 0.0004 & 1106.86 & 3870.31 & 2927.18 & 19447.98 & 40243.01 & 65425.25 & 87587.88 \\
Llama-3.1-8B & incorrect\_temperature\_1\_1 & h200 & 388448 & 0.0085 & 0.34 & 10.24 & 0.00 & 7.01 & 59.65 & 131.60 & 1324.05 \\
Llama-3.1-8B & incorrect\_top\_p\_0\_85 & h200 & 408071 & 0.0000 & 39.27 & 561.00 & 0.00 & 794.06 & 7564.55 & 20686.86 & 42736.77 \\
\bottomrule
\end{tabular}
\caption{Summary statistics for the likelihood-based score derived from the sampled token margin using a Gaussian noise model with standard deviation $\sigma=0.02$. The $\infty$ column indicates the fraction of tokens where the claimed token was filtered out by top-$k$ or top-$p$ sampling.}
\label{tab:raw_scores_likelihood_sigma0p02}
\end{table*}

\clearpage

\begin{table*}[p]
\centering
{\large\textbf{Token-DiFR (likelihood, $\sigma=0.08$) raw score distributions}}

\vspace{0.5em}

\scriptsize
\setlength{\tabcolsep}{3pt}
\begin{tabular}{lllllrrrrrrr}
\toprule
Model & Config & GPU & Count & $\infty$ & Mean & Std & p90 & p99 & p99.9 & p99.99 & p99.999 \\
\midrule
Qwen3-30B-A3B & reference & a100 & 629294 & 0.0011 & 2.61 & 44.62 & 0.00 & 17.20 & 637.12 & 1669.35 & 3264.44 \\
Qwen3-30B-A3B & reference & h200 & 627756 & 0.0001 & 0.02 & 2.41 & 0.00 & 0.00 & 0.76 & 35.39 & 233.19 \\
Qwen3-30B-A3B & correct\_tp4 & a100 & 629072 & 0.0012 & 0.18 & 6.50 & 0.00 & 1.71 & 24.82 & 263.43 & 970.57 \\
Qwen3-30B-A3B & correct\_tp4 & h200 & 630195 & 0.0011 & 0.16 & 6.67 & 0.00 & 1.66 & 22.97 & 241.79 & 885.97 \\
Qwen3-30B-A3B & correct\_hf & h200 & 806271 & 0.0024 & 6.95 & 71.23 & 0.00 & 203.82 & 1078.84 & 2193.21 & 3545.97 \\
Qwen3-30B-A3B & incorrect\_adv\_kv\_fp8\_temp\_0\_96 & h200 & 604449 & 0.0031 & 0.85 & 14.70 & 0.00 & 14.72 & 154.73 & 609.94 & 1360.77 \\
Qwen3-30B-A3B & incorrect\_config\_random\_bug\_k\_2 & h200 & 625908 & 0.0031 & 0.56 & 20.68 & 0.00 & 0.00 & 138.40 & 992.97 & 2287.25 \\
Qwen3-30B-A3B & incorrect\_config\_random\_bug\_k\_32 & h200 & 621929 & 0.0094 & 0.14 & 12.26 & 0.00 & 0.00 & 1.88 & 392.99 & 1803.45 \\
Qwen3-30B-A3B & incorrect\_kv\_cache\_quant\_fp8 & h200 & 603591 & 0.0040 & 0.74 & 13.93 & 0.00 & 13.00 & 130.84 & 517.09 & 1431.49 \\
Qwen3-30B-A3B & incorrect\_model\_quant\_4bit & h200 & 559981 & 0.0114 & 2.86 & 28.29 & 0.00 & 74.79 & 372.46 & 935.29 & 2430.31 \\
Qwen3-30B-A3B & incorrect\_model\_quant\_fp8 & h200 & 661475 & 0.0027 & 0.55 & 10.71 & 0.00 & 8.80 & 101.01 & 475.88 & 1030.54 \\
Qwen3-30B-A3B & incorrect\_seed\_43 & h200 & 665261 & 0.0001 & 31.64 & 147.87 & 37.00 & 715.19 & 1757.80 & 3207.74 & 4895.66 \\
Qwen3-30B-A3B & incorrect\_temperature\_1\_1 & h200 & 615796 & 0.0040 & 0.02 & 0.81 & 0.00 & 0.00 & 3.42 & 12.79 & 82.44 \\
Qwen3-30B-A3B & incorrect\_top\_p\_0\_85 & h200 & 628161 & 0.0000 & 1.78 & 29.49 & 0.00 & 25.20 & 388.04 & 1166.87 & 2568.49 \\
Qwen3-8B & reference & a100 & 786831 & 0.0005 & 0.08 & 5.97 & 0.00 & 0.00 & 3.72 & 167.76 & 790.91 \\
Qwen3-8B & reference & h200 & 784915 & 0.0005 & 0.07 & 4.63 & 0.00 & 0.00 & 3.44 & 143.04 & 719.45 \\
Qwen3-8B & correct\_tp4 & a100 & 788186 & 0.0006 & 0.07 & 4.75 & 0.00 & 0.00 & 3.79 & 150.86 & 720.98 \\
Qwen3-8B & correct\_tp4 & h200 & 785547 & 0.0005 & 0.08 & 5.53 & 0.00 & 0.00 & 3.79 & 156.00 & 797.68 \\
Qwen3-8B & correct\_hf & h200 & 783799 & 0.0006 & 0.06 & 4.79 & 0.00 & 0.00 & 3.48 & 133.22 & 752.53 \\
Qwen3-8B & incorrect\_adv\_kv\_fp8\_temp\_0\_96 & h200 & 774505 & 0.0018 & 0.57 & 12.87 & 0.00 & 7.57 & 107.28 & 577.94 & 1298.61 \\
Qwen3-8B & incorrect\_adv\_model\_4bit\_temp\_0\_89 & h200 & 708037 & 0.0051 & 3.92 & 35.02 & 0.00 & 97.03 & 491.14 & 1202.90 & 2254.79 \\
Qwen3-8B & incorrect\_config\_random\_bug\_k\_2 & h200 & 783008 & 0.0029 & 0.75 & 24.04 & 0.00 & 0.76 & 223.59 & 1030.96 & 2186.76 \\
Qwen3-8B & incorrect\_config\_random\_bug\_k\_32 & h200 & 777720 & 0.0097 & 0.26 & 15.67 & 0.00 & 0.00 & 4.90 & 720.84 & 1833.46 \\
Qwen3-8B & incorrect\_kv\_cache\_quant\_fp8 & h200 & 764520 & 0.0028 & 0.45 & 11.03 & 0.00 & 6.55 & 72.92 & 505.52 & 1319.25 \\
Qwen3-8B & incorrect\_model\_quant\_4bit & h200 & 710621 & 0.0085 & 3.04 & 29.26 & 0.00 & 75.34 & 382.74 & 1079.82 & 1970.89 \\
Qwen3-8B & incorrect\_model\_quant\_fp8 & h200 & 778203 & 0.0021 & 0.40 & 10.40 & 0.00 & 5.61 & 63.18 & 451.75 & 1198.37 \\
Qwen3-8B & incorrect\_seed\_43 & h200 & 738399 & 0.0005 & 46.09 & 183.99 & 98.22 & 895.27 & 2079.77 & 3690.57 & 5510.38 \\
Qwen3-8B & incorrect\_temperature\_1\_1 & h200 & 781435 & 0.0059 & 0.04 & 1.38 & 0.00 & 1.17 & 4.98 & 9.43 & 67.43 \\
Qwen3-8B & incorrect\_top\_p\_0\_85 & h200 & 783913 & 0.0000 & 2.33 & 33.49 & 0.00 & 45.61 & 445.82 & 1249.02 & 2728.71 \\
Llama-3.1-8B & reference & a100 & 398030 & 0.0004 & 0.06 & 5.12 & 0.00 & 0.00 & 1.97 & 125.64 & 737.27 \\
Llama-3.1-8B & reference & h200 & 400206 & 0.0004 & 0.06 & 4.92 & 0.00 & 0.00 & 1.94 & 149.91 & 720.43 \\
Llama-3.1-8B & correct\_tp4 & a100 & 401402 & 0.0005 & 0.07 & 4.64 & 0.00 & 0.00 & 2.45 & 146.14 & 805.21 \\
Llama-3.1-8B & correct\_tp4 & h200 & 402934 & 0.0005 & 0.06 & 3.98 & 0.00 & 0.00 & 2.26 & 135.15 & 619.47 \\
Llama-3.1-8B & correct\_hf & h200 & 398700 & 0.0005 & 0.04 & 3.62 & 0.00 & 0.00 & 1.95 & 108.74 & 503.41 \\
Llama-3.1-8B & incorrect\_adv\_kv\_fp8\_temp\_0\_96 & h200 & 399515 & 0.0016 & 0.45 & 10.10 & 0.00 & 5.09 & 89.25 & 473.67 & 1000.49 \\
Llama-3.1-8B & incorrect\_adv\_model\_4bit\_temp\_0\_89 & h200 & 382343 & 0.0039 & 2.72 & 29.03 & 0.00 & 61.98 & 384.32 & 1112.00 & 2072.92 \\
Llama-3.1-8B & incorrect\_config\_random\_bug\_k\_2 & h200 & 399210 & 0.0028 & 0.75 & 21.86 & 0.00 & 0.00 & 239.27 & 1023.62 & 1716.17 \\
Llama-3.1-8B & incorrect\_config\_random\_bug\_k\_32 & h200 & 396691 & 0.0091 & 0.71 & 28.73 & 0.00 & 0.00 & 79.24 & 1542.66 & 2243.95 \\
Llama-3.1-8B & incorrect\_kv\_cache\_quant\_fp8 & h200 & 399409 & 0.0029 & 0.32 & 8.13 & 0.00 & 4.09 & 51.19 & 357.00 & 945.80 \\
Llama-3.1-8B & incorrect\_model\_quant\_4bit & h200 & 381999 & 0.0092 & 1.99 & 21.41 & 0.00 & 45.61 & 286.38 & 812.42 & 1368.68 \\
Llama-3.1-8B & incorrect\_model\_quant\_fp8 & h200 & 399564 & 0.0025 & 0.31 & 9.25 & 0.00 & 4.01 & 44.94 & 414.25 & 1116.55 \\
Llama-3.1-8B & incorrect\_seed\_43 & h200 & 392316 & 0.0004 & 69.94 & 242.85 & 186.49 & 1219.93 & 2519.96 & 4094.08 & 5479.38 \\
Llama-3.1-8B & incorrect\_temperature\_1\_1 & h200 & 388448 & 0.0085 & 0.04 & 0.72 & 0.00 & 1.52 & 5.53 & 10.35 & 85.93 \\
Llama-3.1-8B & incorrect\_top\_p\_0\_85 & h200 & 408071 & 0.0000 & 2.53 & 35.32 & 0.00 & 52.57 & 476.77 & 1297.39 & 2675.85 \\
\bottomrule
\end{tabular}
\caption{Summary statistics for the likelihood-based score derived from the sampled token margin using a Gaussian noise model with standard deviation $\sigma=0.08$. The $\infty$ column indicates the fraction of tokens where the claimed token was filtered out by top-$k$ or top-$p$ sampling.}
\label{tab:raw_scores_likelihood_sigma0p08}
\end{table*}

%% file: sections/appendix-additional-results.tex
\clearpage

\section{All Detection Results Across Models and Configurations}
\label{app:comprehensive_results}

This appendix presents comprehensive batch scaling results across all three evaluated model architectures (Llama-3.1-8B-Instruct, Qwen3-8B, and Qwen3-30B-A3B) and all misconfiguration types. These results demonstrate the robustness of our verification methods across different model architectures and scales.

\clearpage

% Put this once in your preamble or before the figure:
\newcommand{\llamaPlotWidth}{0.8\textwidth}
% You can change 0.8 to 0.7, 0.9, etc to make all three plots smaller or larger.

\begin{figure*}[t]
    \centering

    \begin{subfigure}[t]{\llamaPlotWidth}
        \centering
        \includegraphics[width=\llamaPlotWidth]{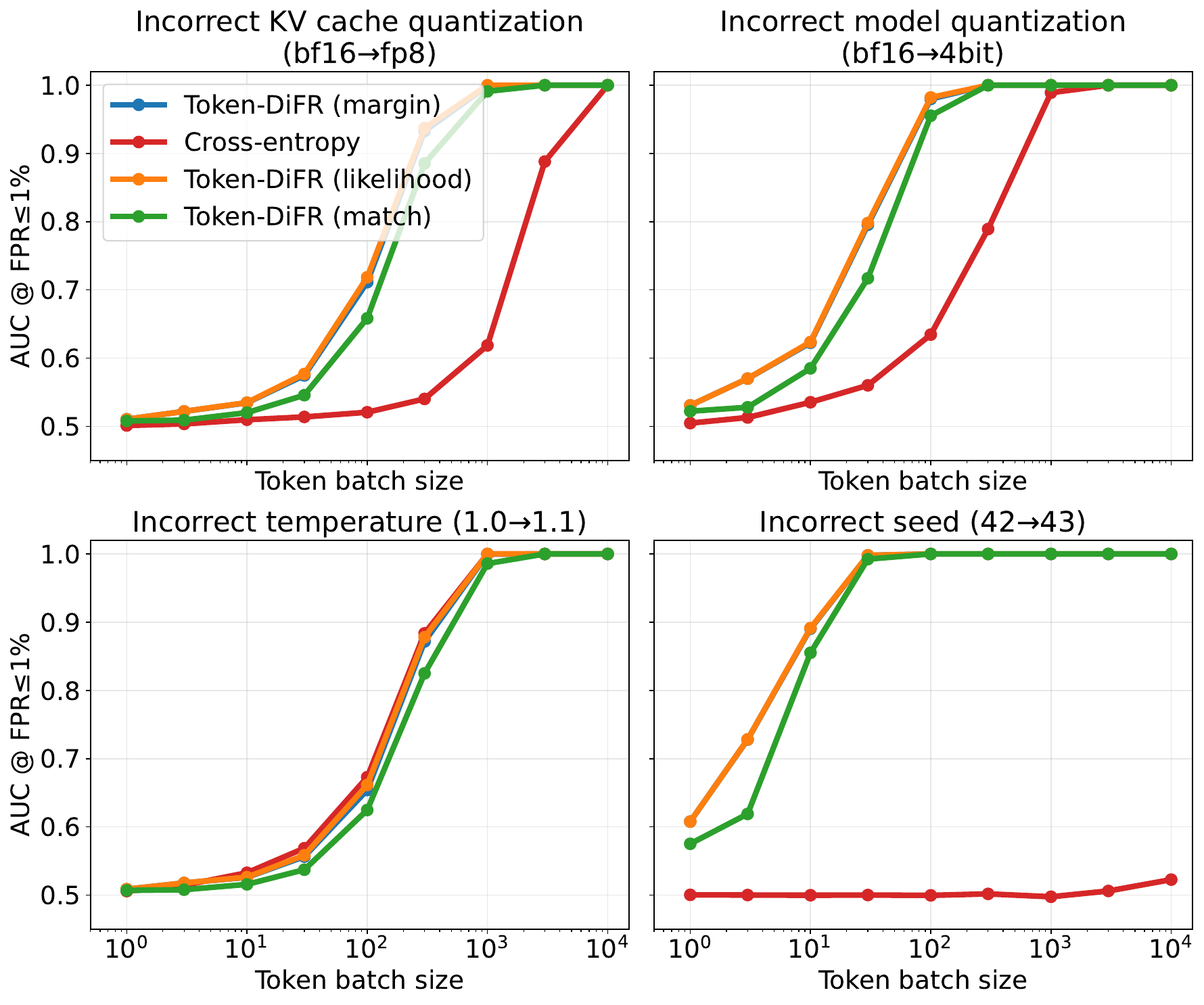}
        \caption{Standard misconfigurations.}
        \label{fig:llama_all_results_standard}
    \end{subfigure}

    \vspace{0.75em}

    \begin{subfigure}[t]{\llamaPlotWidth}
        \centering
        \includegraphics[width=\llamaPlotWidth]{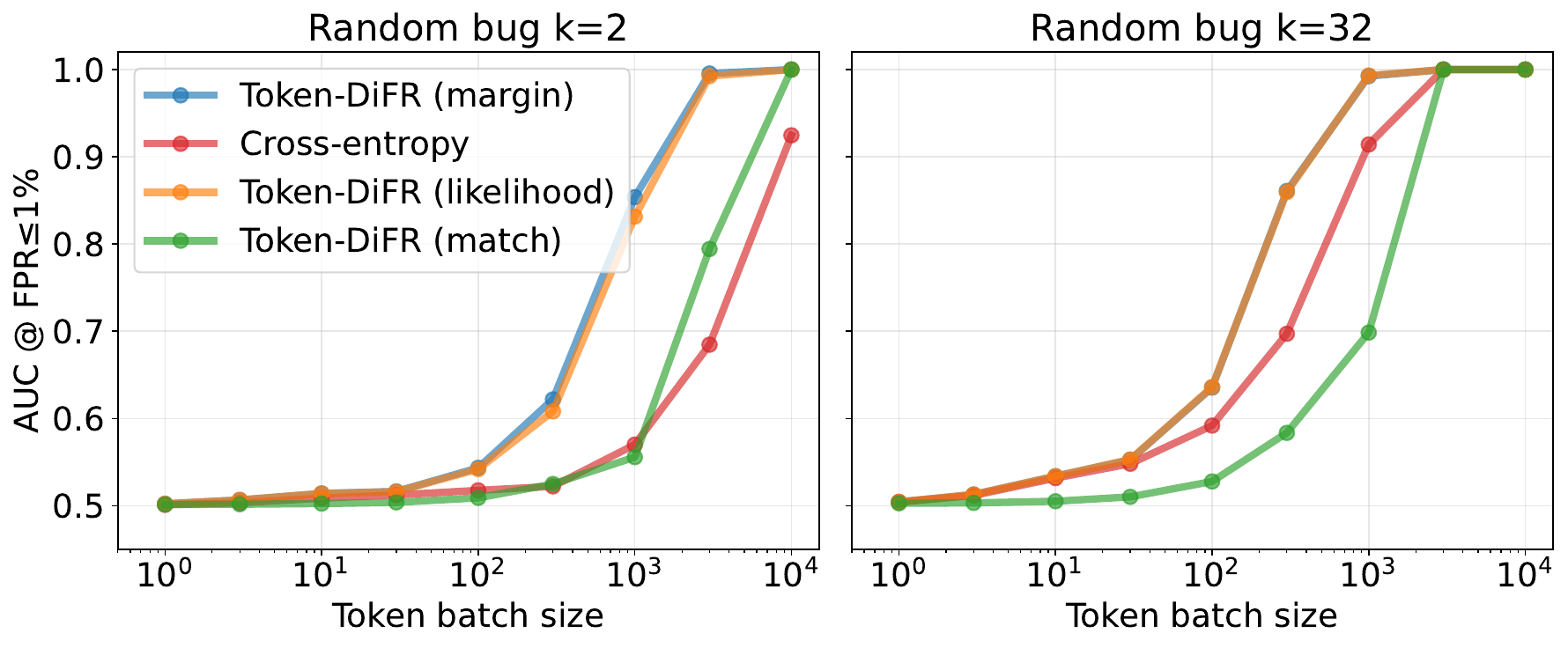}
        \caption{Bug simulations.}
        \label{fig:llama_all_results_bugs}
    \end{subfigure}

    \vspace{0.75em}

    \begin{subfigure}[t]{\llamaPlotWidth}
        \centering
        \includegraphics[width=\llamaPlotWidth]{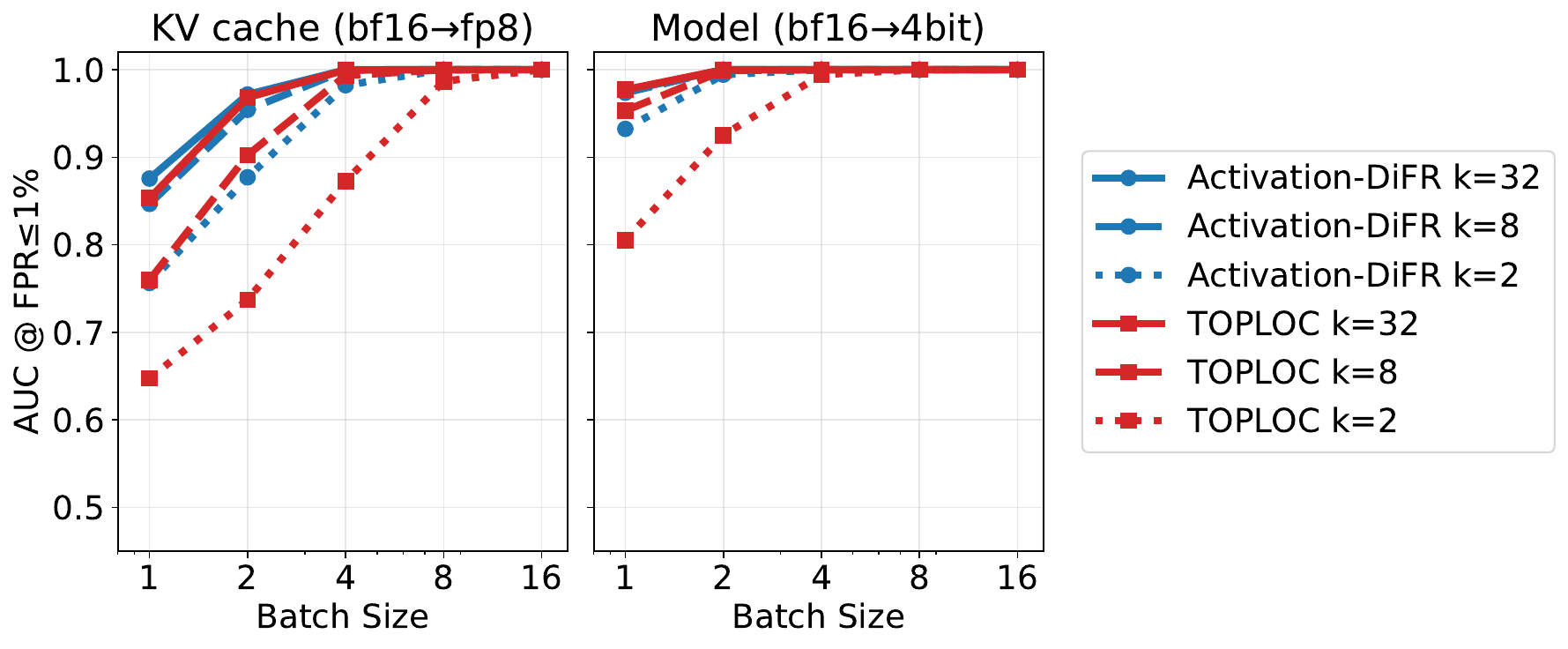}
        \caption{Activation comparisons (AUC).}
        \label{fig:llama_all_results_activation}
    \end{subfigure}

    \caption{Token-DiFR and Activation-DiFR results for Llama 3.1 8B with the four metric comparison (Token-DiFR, cross entropy, Token-DiFR likelihood, and exact token match).}
    \label{fig:llama_all_results}
\end{figure*}

\clearpage

\begin{figure*}[t]
    \centering

    \begin{subfigure}[t]{\llamaPlotWidth}
        \centering
        \includegraphics[width=\llamaPlotWidth]{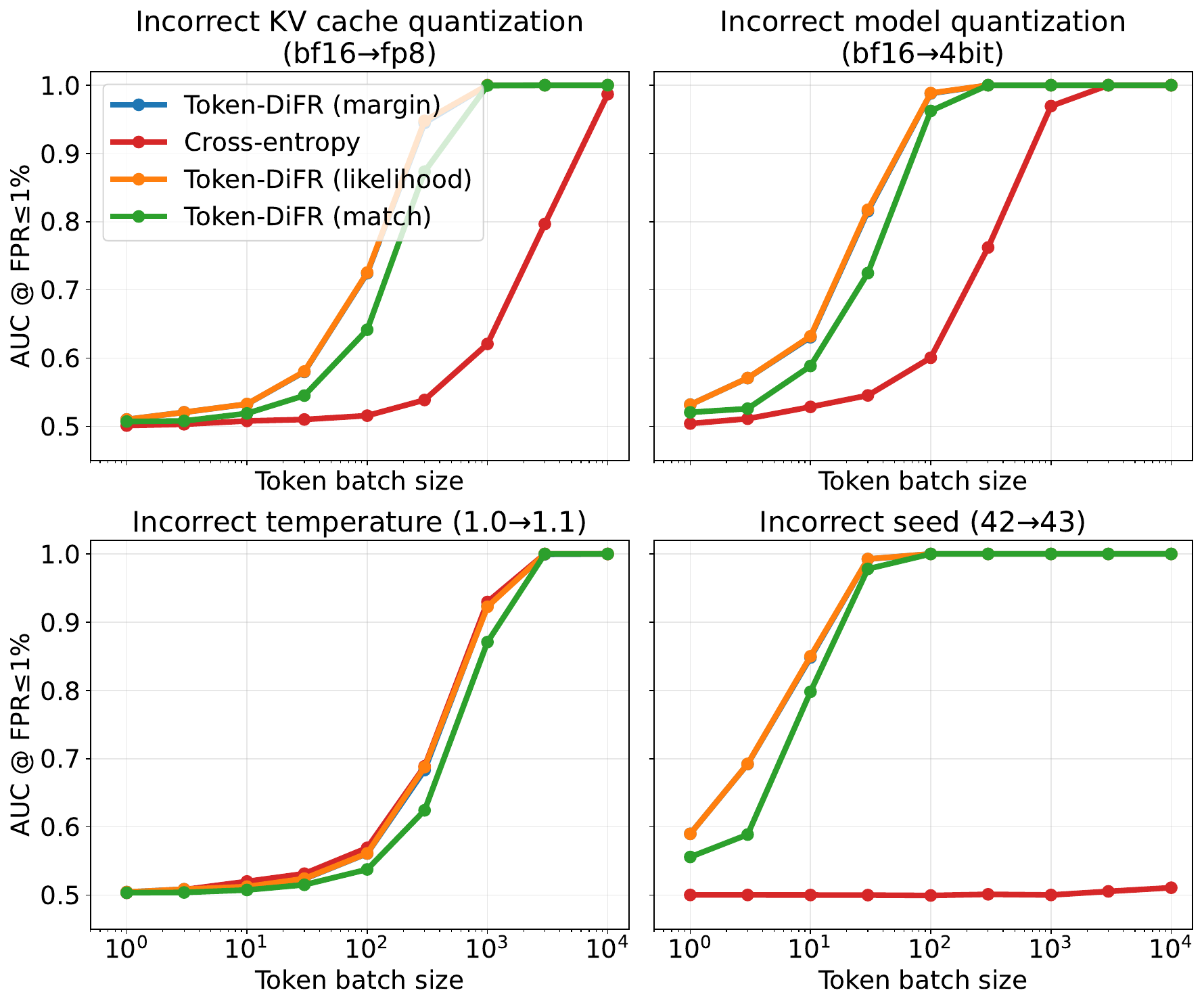}
        \caption{Standard misconfigurations.}
    \end{subfigure}

    \vspace{0.75em}

    \begin{subfigure}[t]{\llamaPlotWidth}
        \centering
        \includegraphics[width=\llamaPlotWidth]{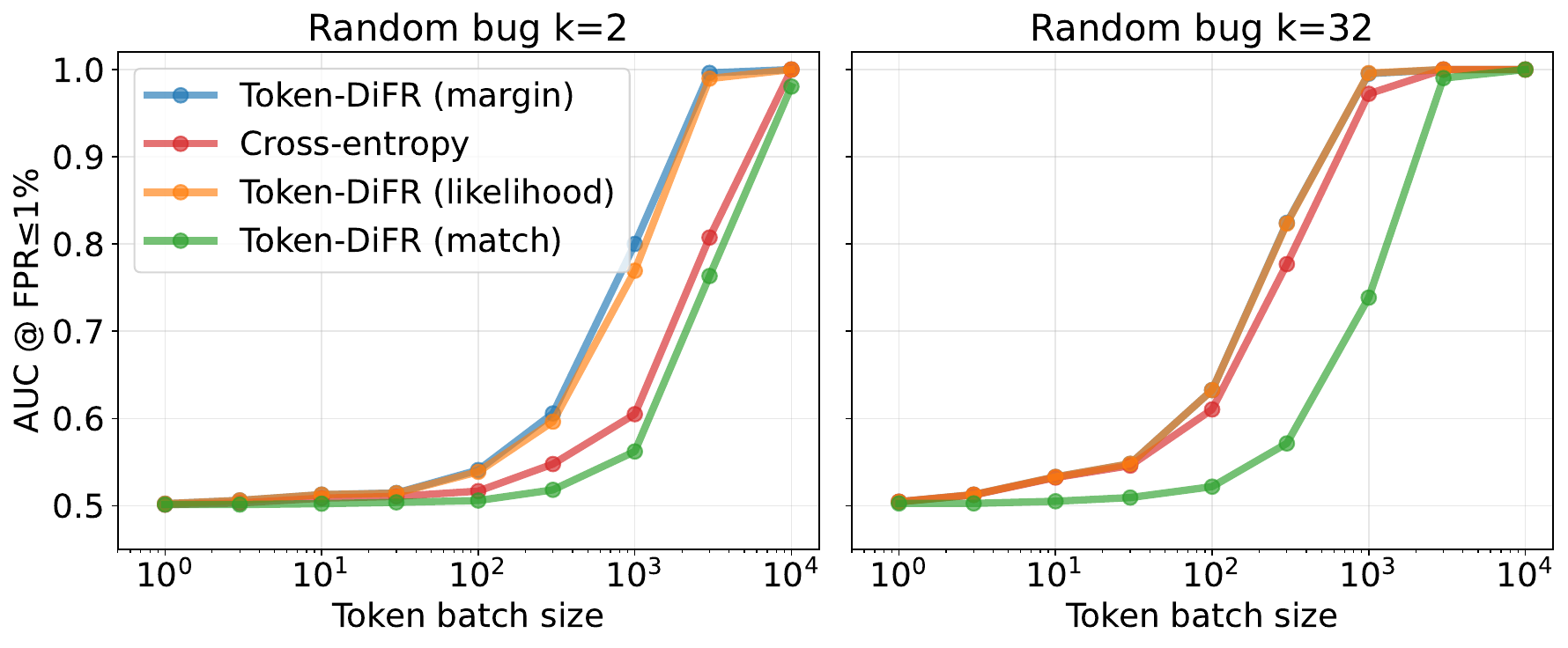}
        \caption{Bug simulations.}
    \end{subfigure}

    \vspace{0.75em}

    \begin{subfigure}[t]{\llamaPlotWidth}
        \centering
        \includegraphics[width=\llamaPlotWidth]{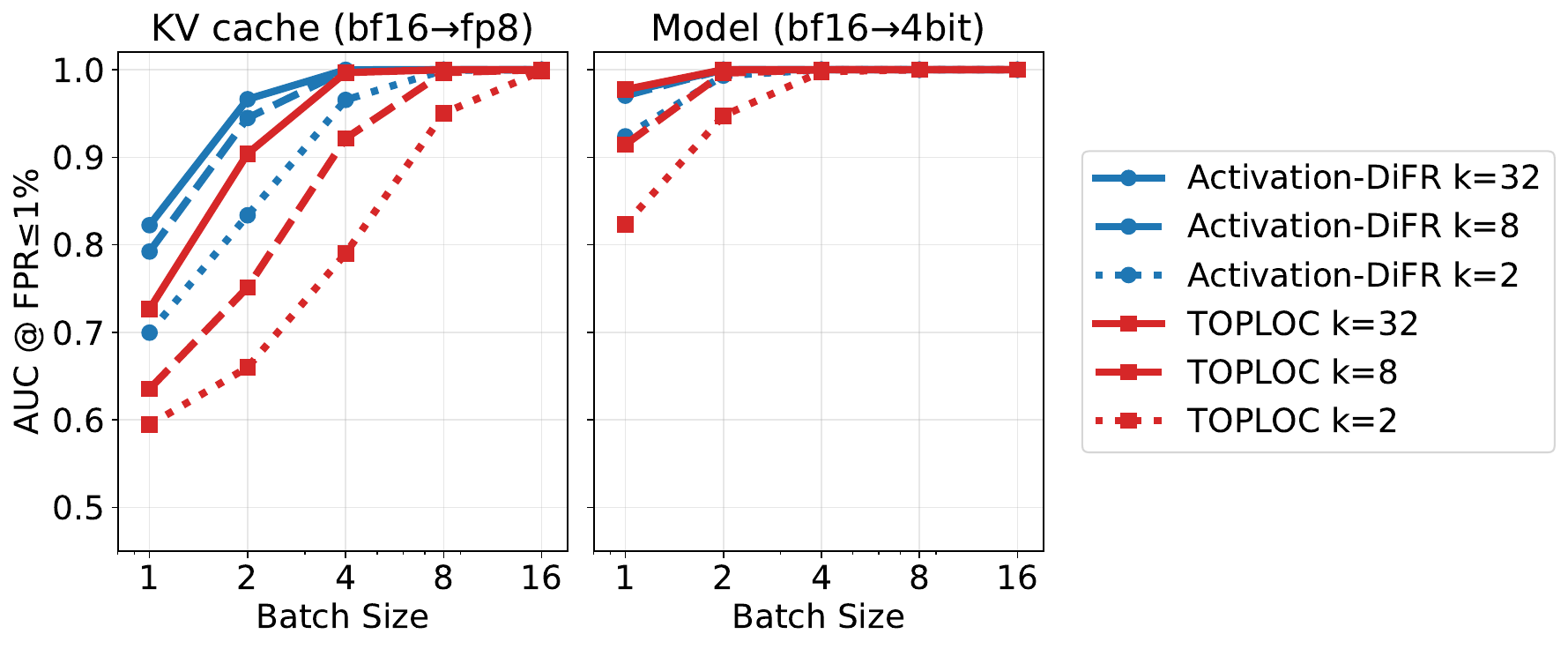}
        \caption{Activation comparisons (AUC).}
    \end{subfigure}

    \caption{Token-DiFR and Activation-DiFR results for Qwen3-8B with the four metric comparison (Token-DiFR, cross entropy, Token-DiFR likelihood, and exact token match).}
    \label{fig:qwen3_8B_all_results}
\end{figure*}

\clearpage
\section{Qwen3-30B-A3B Results}
\label{app:qwen3-30b-a3b-results}

For Llama-3.1-8B and Qwen3-8B, we find that detection performance is highly stable across different environments. Varying hardware (A100 vs.\ H200), tensor-parallel configurations, or inference implementations (vLLM versus HuggingFace) produces only minor changes in the score distributions, and even the most subtle misconfigurations remain reliably detectable using our default Token-DiFR configuration.

The Qwen3-30B-A3B results exhibit substantially more variability. The largest shifts in behavior arise when switching between vLLM and HuggingFace, although we also observe noticeable changes when moving from H200 to A100 within vLLM. It is unclear whether these differences are driven primarily by the mixture-of-experts architecture or by implementation details in the current inference stacks. This is a particularly challenging setting: when verifier and provider both use our reference H200/vLLM configuration, Qwen3-30B-A3B achieves an unusually high exact-match rate of around 99.9\%, while the HuggingFace configuration has a much lower exact-match rate of roughly 95\%.

When the verifier matches the provider’s setup exactly (reference H200-only), the 30B model behaves much like the smaller models: minor misconfigurations, including small temperature shifts and FP8 KV-cache quantization, are detected accurately and consistently without tuning. As we expand the calibration set to the trusted vLLM pool, which includes multiple parallelism setups and both H100 and A100 runs, performance for the smallest deviations degrades: in particular, we can no longer reliably detect the temperature 1.1 vs.\ 1.0 misconfiguration at our target false positive rate. Adding HuggingFace runs in the pooled-with-HF configuration broadens the null distribution further, and FP8 KV-cache quantization also becomes difficult to distinguish from the pooled honest baseline. Despite these substantial implementation-driven differences, all configurations still cleanly detect major problems such as 4-bit model quantization and incorrect seeds with modest batch sizes.

In the following pages, we present Qwen3-30B-A3B results for all three verifier–provider configurations we consider: \textbf{reference H200 only}, \textbf{trusted vLLM pool}, and \textbf{trusted vLLM and HuggingFace pool}.

\clearpage

\begin{figure*}[t]
    \centering

    \begin{subfigure}[t]{\llamaPlotWidth}
        \centering
        \includegraphics[width=\llamaPlotWidth]{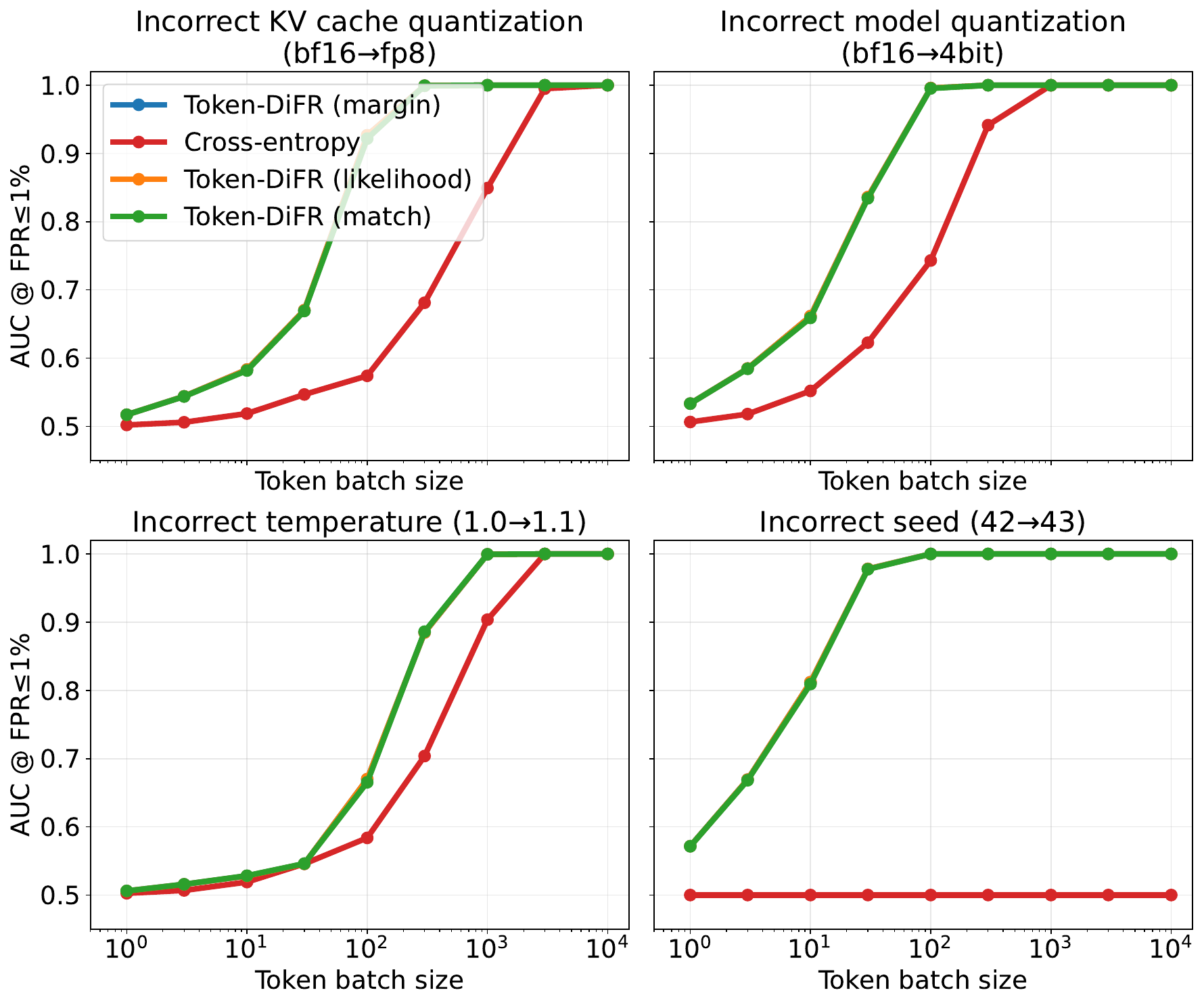}
        \caption{Standard misconfigurations.}
    \end{subfigure}

    \vspace{0.75em}

    \begin{subfigure}[t]{\llamaPlotWidth}
        \centering
        \includegraphics[width=\llamaPlotWidth]{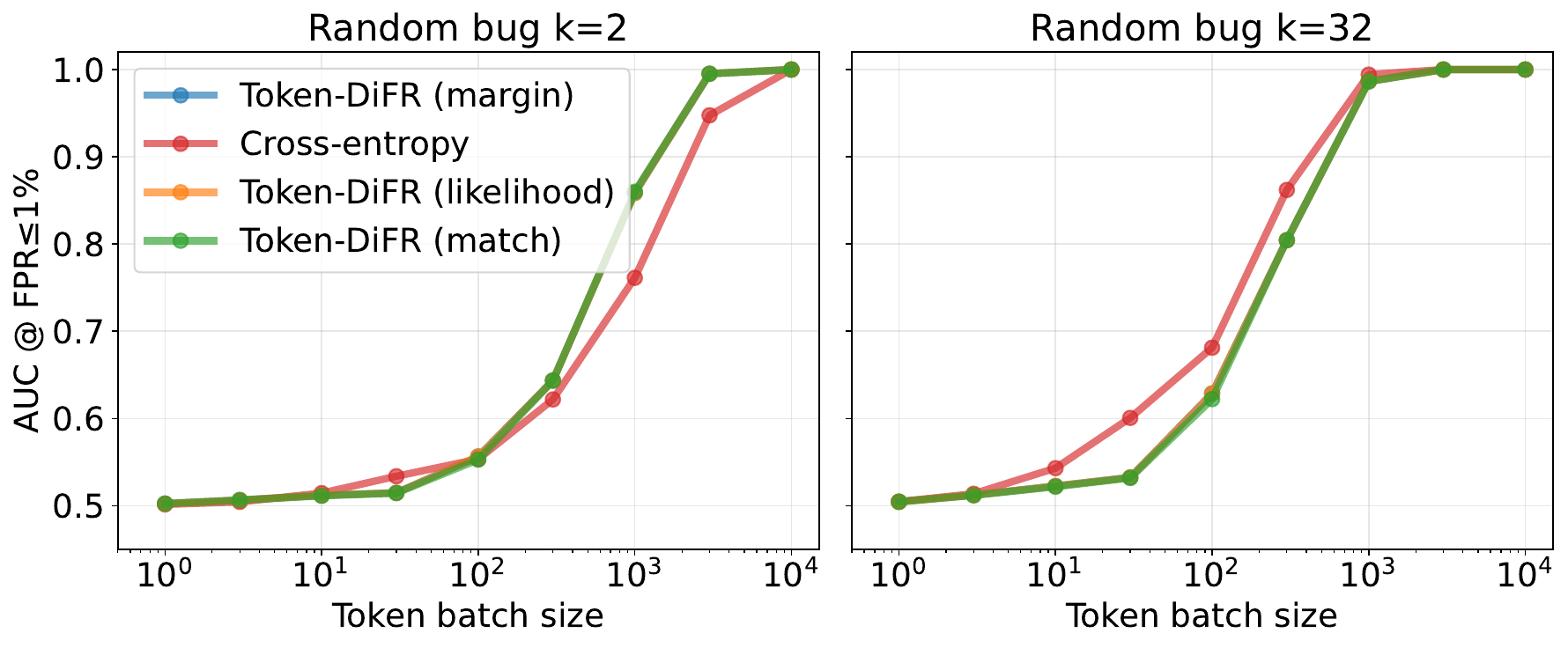}
        \caption{Bug simulations.}
    \end{subfigure}

    \vspace{0.75em}

    \begin{subfigure}[t]{\llamaPlotWidth}
        \centering
        \includegraphics[width=\llamaPlotWidth]{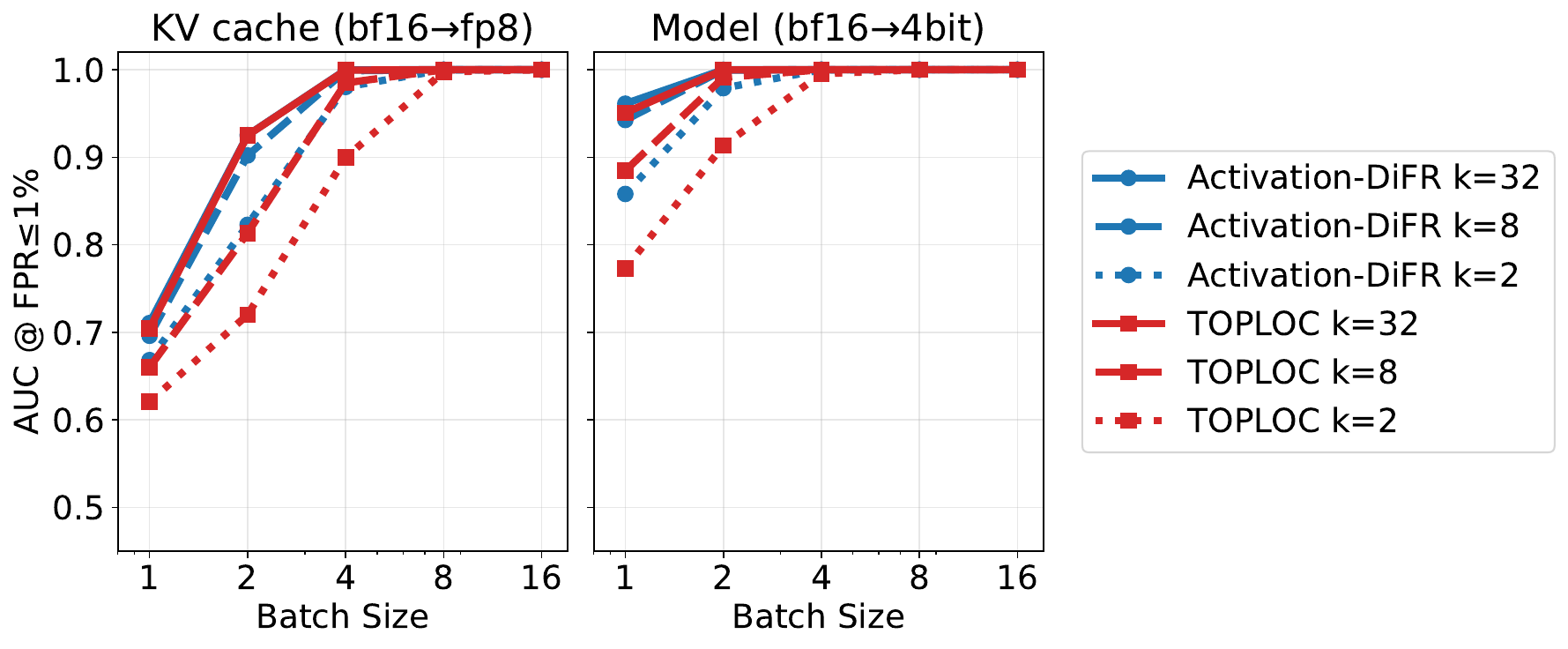}
        \caption{Activation comparisons (AUC).}
    \end{subfigure}

    \caption{Token-DiFR and Activation-DiFR results for Qwen3-30B-A3B with the reference H200 only setup with the four metric comparison (Token-DiFR, cross entropy, Token-DiFR likelihood, and exact token match).}
    \label{fig:qwen3_30B_h200_all_results}
\end{figure*}

\clearpage

\begin{figure*}[t]
    \centering

    \begin{subfigure}[t]{\llamaPlotWidth}
        \centering
        \includegraphics[width=\llamaPlotWidth]{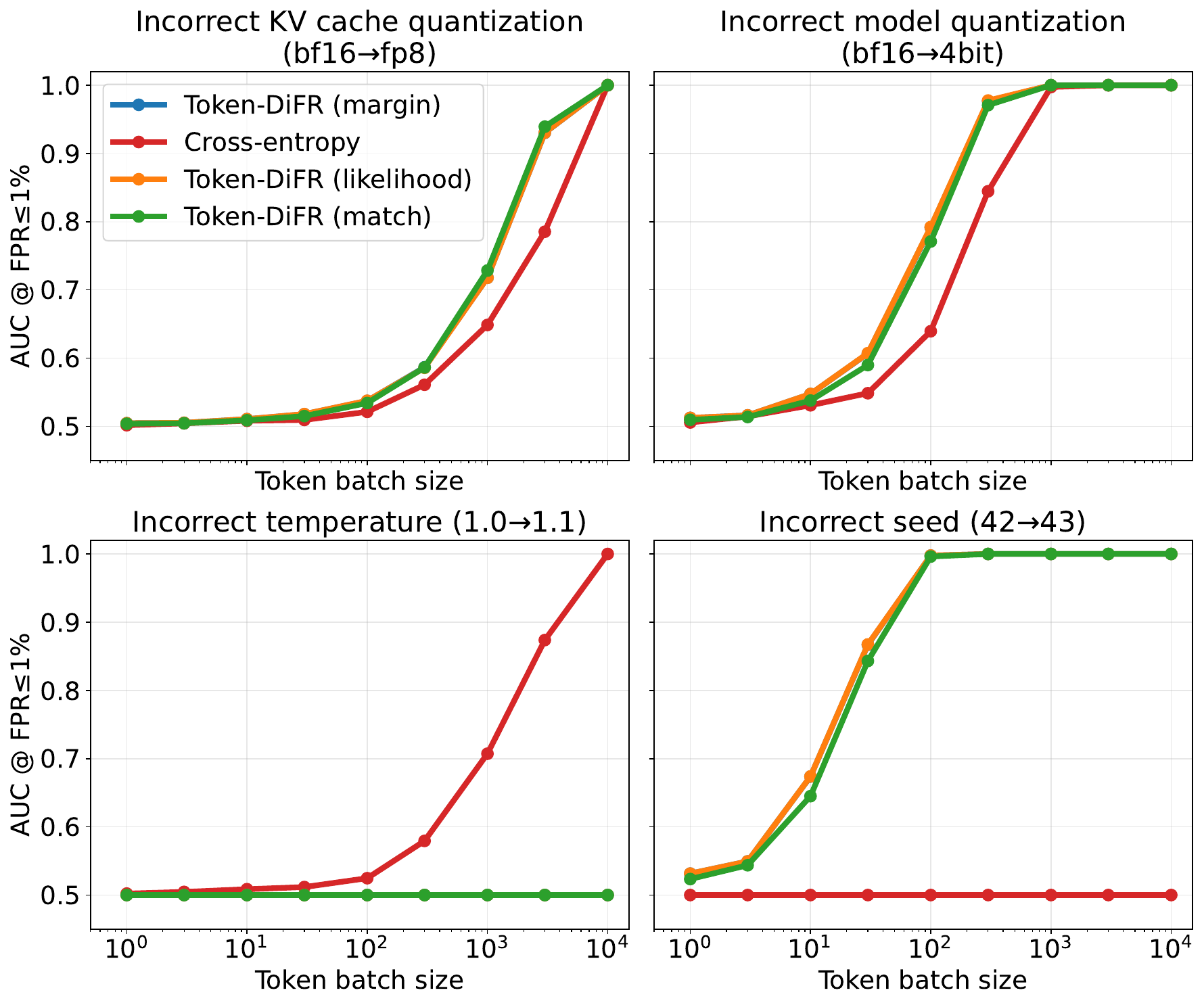}
        \caption{Standard misconfigurations.}
    \end{subfigure}

    \vspace{0.75em}

    \begin{subfigure}[t]{\llamaPlotWidth}
        \centering
        \includegraphics[width=\llamaPlotWidth]{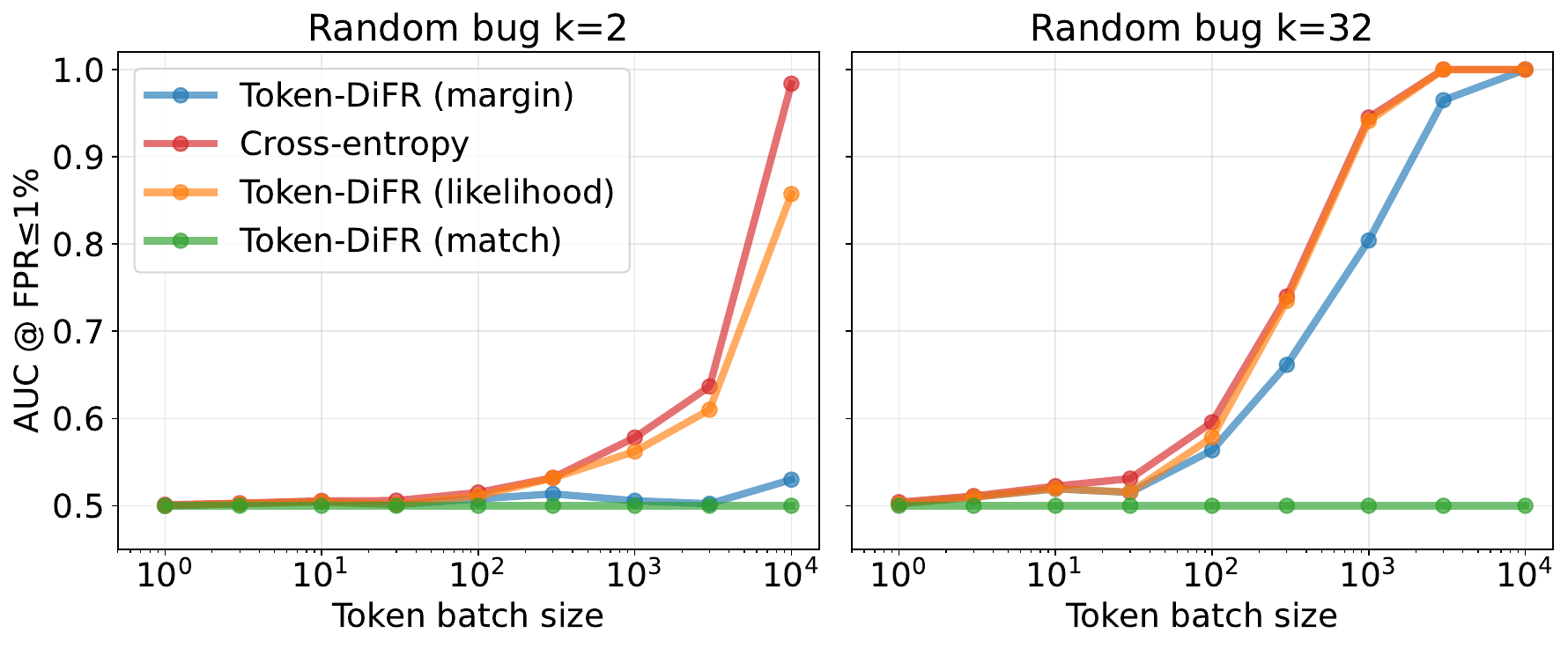}
        \caption{Bug simulations.}
    \end{subfigure}

    \vspace{0.75em}

    \begin{subfigure}[t]{\llamaPlotWidth}
        \centering
        \includegraphics[width=\llamaPlotWidth]{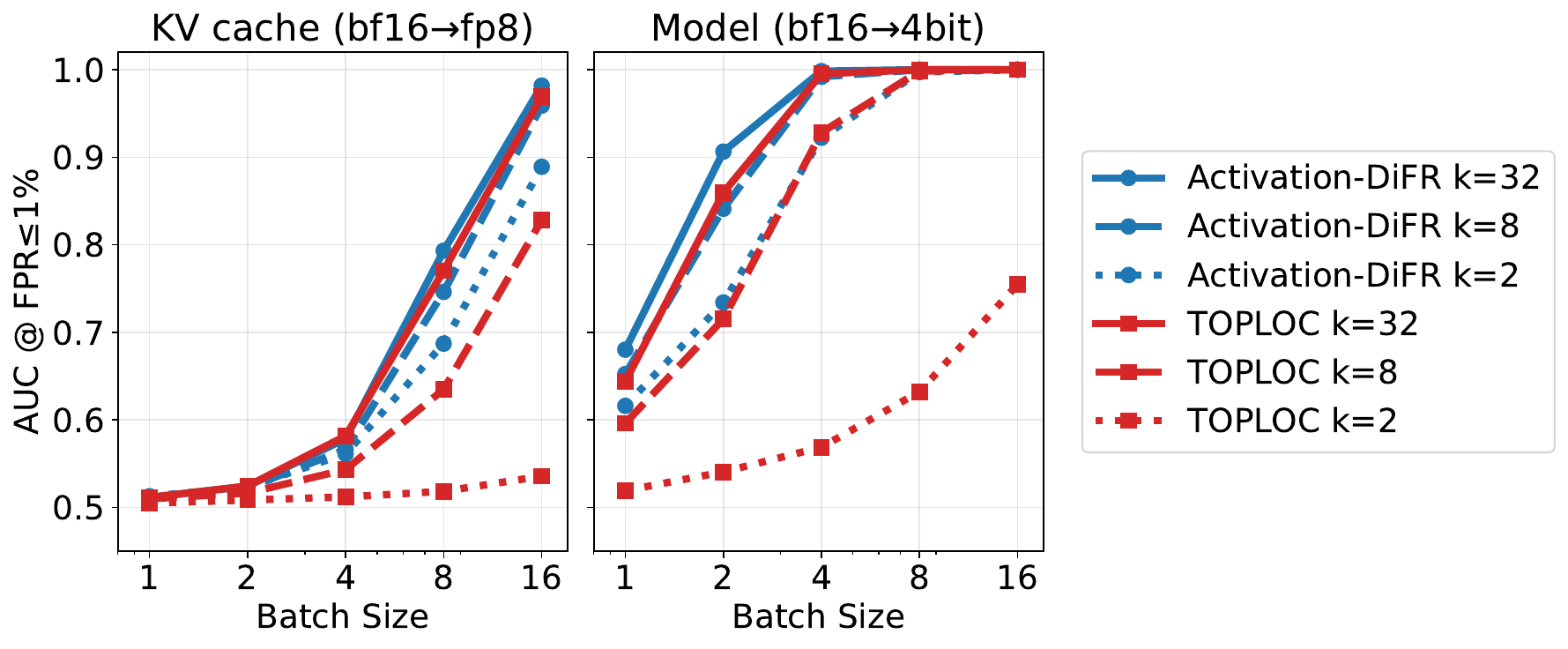}
        \caption{Activation comparisons (AUC).}
    \end{subfigure}

    \caption{Token-DiFR and Activation-DiFR results for Qwen3-30B-A3B with the vLLM pooled setup with the four metric comparison (Token-DiFR, cross entropy, Token-DiFR likelihood, and exact token match).}
    \label{fig:qwen3_30B_pooled_all_results}
\end{figure*}

\clearpage

\begin{figure*}[t]
    \centering

    \begin{subfigure}[t]{\llamaPlotWidth}
        \centering
        \includegraphics[width=\llamaPlotWidth]{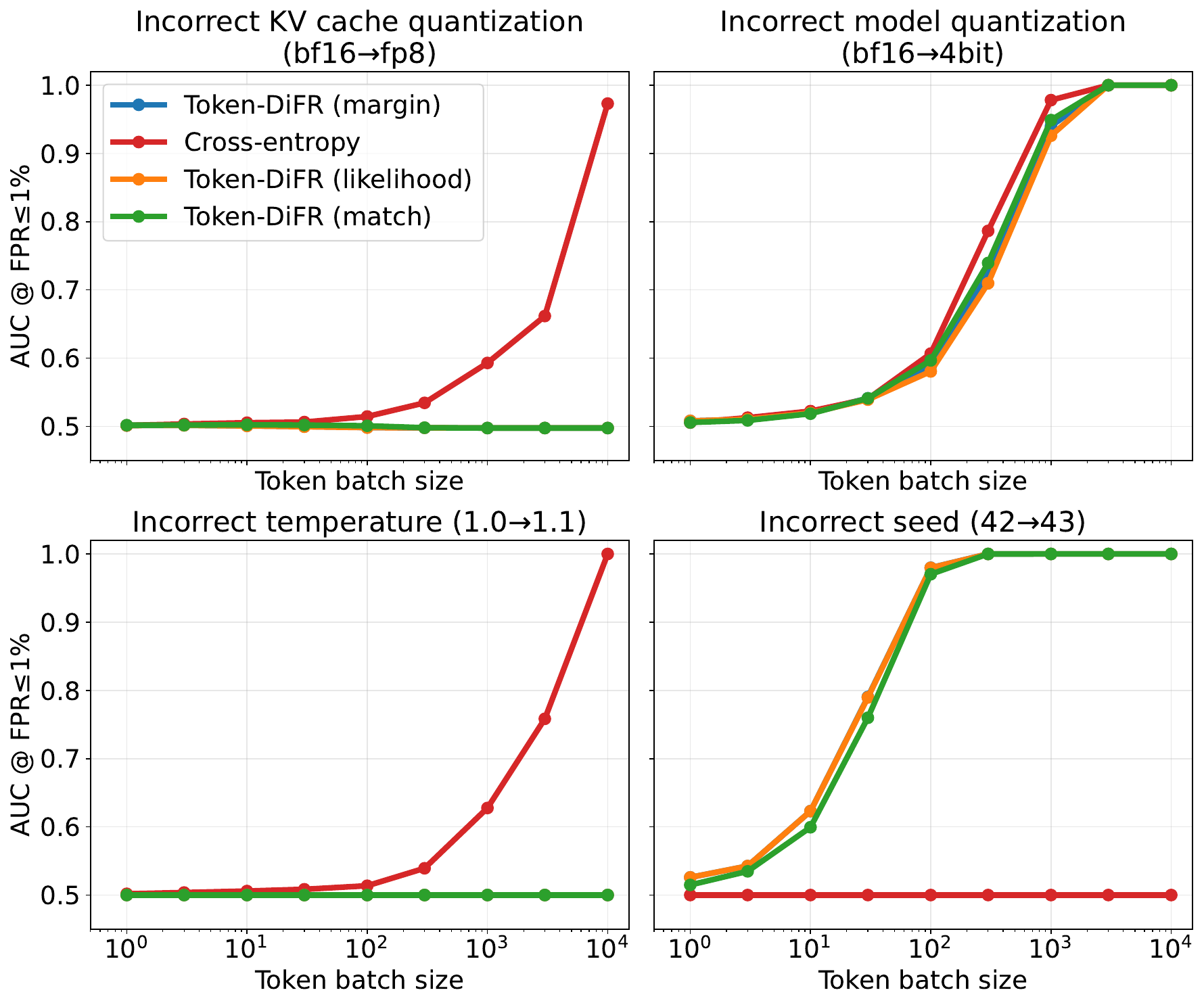}
        \caption{Standard misconfigurations.}
    \end{subfigure}

    \vspace{0.75em}

    \begin{subfigure}[t]{\llamaPlotWidth}
        \centering
        \includegraphics[width=\llamaPlotWidth]{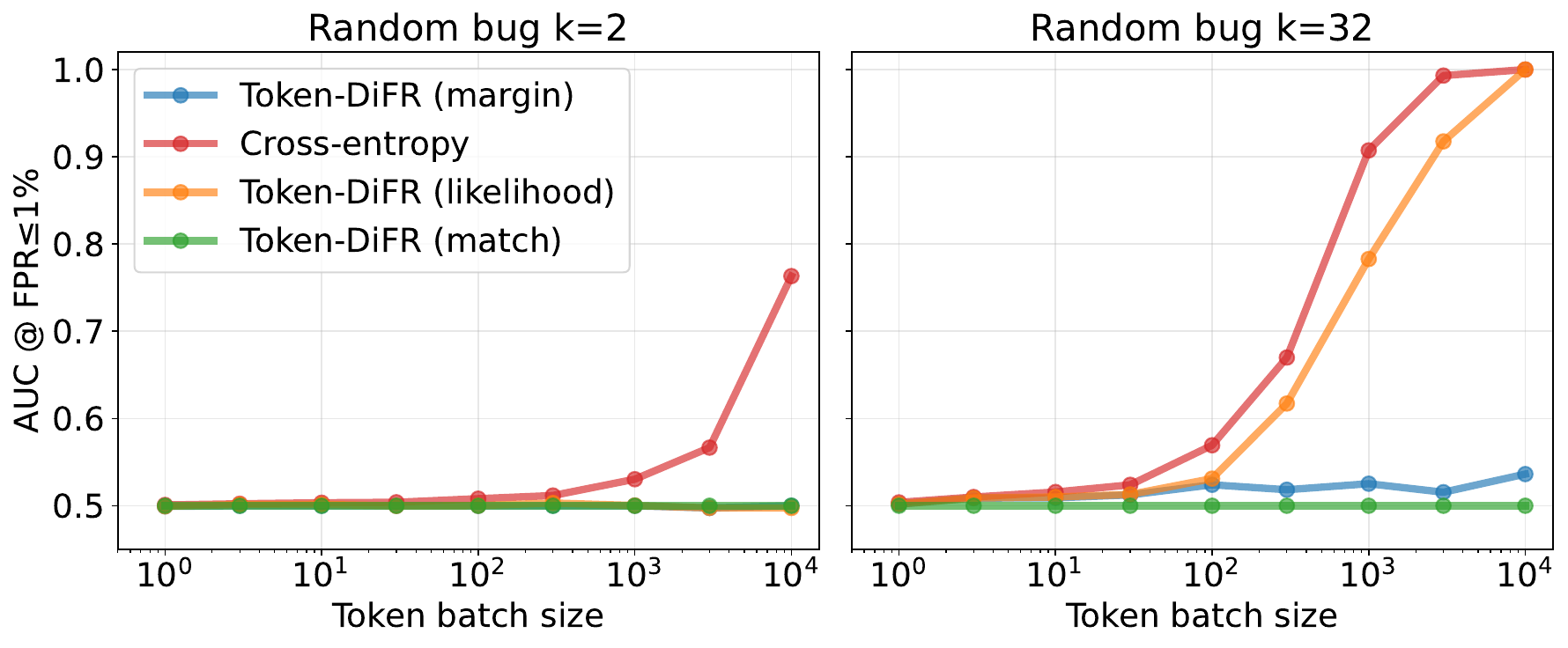}
        \caption{Bug simulations.}
    \end{subfigure}

    \vspace{0.75em}

    \begin{subfigure}[t]{\llamaPlotWidth}
        \centering
        \includegraphics[width=\llamaPlotWidth]{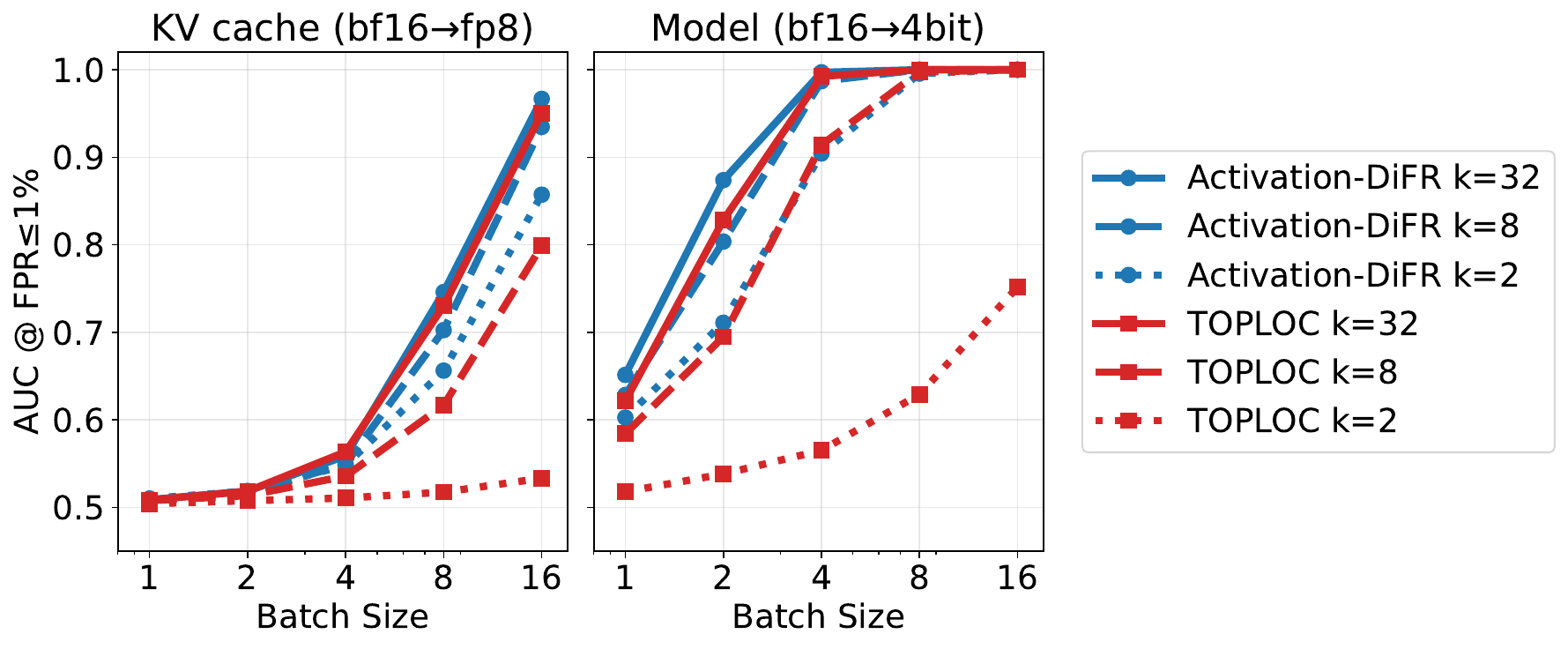}
        \caption{Activation comparisons (AUC).}
    \end{subfigure}

    \caption{Token-DiFR and Activation-DiFR results for Qwen3-30B-A3B with the vLLM + HuggingFace pooled setup with the four metric comparison (Token-DiFR, cross entropy, Token-DiFR likelihood, and exact token match).}
    \label{fig:qwen3_30B_pooled_with_hf_all_results}
\end{figure*}

%% file: citations.bib
@misc{ong2025toploclocalitysensitivehashing,
      title={TOPLOC: A Locality Sensitive Hashing Scheme for Trustless Verifiable Inference},
      author={Jack Min Ong and Matthew Di Ferrante and Aaron Pazdera and Ryan Garner and Sami Jaghouar and Manveer Basra and Max Ryabinin and Johannes Hagemann},
      year={2025},
      eprint={2501.16007},
      archivePrefix={arXiv},
      primaryClass={cs.CR},
      url={https://arxiv.org/abs/2501.16007},
}

@misc{wolf2020huggingfacestransformersstateoftheartnatural,
      title={HuggingFace's Transformers: State-of-the-art Natural Language Processing}, 
      author={Thomas Wolf and Lysandre Debut and Victor Sanh and Julien Chaumond and Clement Delangue and Anthony Moi and Pierric Cistac and Tim Rault and Rémi Louf and Morgan Funtowicz and Joe Davison and Sam Shleifer and Patrick von Platen and Clara Ma and Yacine Jernite and Julien Plu and Canwen Xu and Teven Le Scao and Sylvain Gugger and Mariama Drame and Quentin Lhoest and Alexander M. Rush},
      year={2020},
      eprint={1910.03771},
      archivePrefix={arXiv},
      primaryClass={cs.CL},
      url={https://arxiv.org/abs/1910.03771}, 
}

@misc{zhu2025auditingblackboxllmapis,
      title={Auditing Black-Box LLM APIs with a Rank-Based Uniformity Test}, 
      author={Xiaoyuan Zhu and Yaowen Ye and Tianyi Qiu and Hanlin Zhu and Sijun Tan and Ajraf Mannan and Jonathan Michala and Raluca Ada Popa and Willie Neiswanger},
      year={2025},
      eprint={2506.06975},
      archivePrefix={arXiv},
      primaryClass={cs.CR},
      url={https://arxiv.org/abs/2506.06975}, 
}

@misc{verifyingModelWeightExfil,
      title={Verifying LLM Inference to Prevent Model Weight Exfiltration}, 
      author={Roy Rinberg and Adam Karvonen and Alex Hoover and Daniel Reuter and Keri Warr},
      year={2025},
      eprint={2511.02620},
      archivePrefix={arXiv},
      primaryClass={cs.CR},
      url={https://arxiv.org/abs/2511.02620}, 
}

@book{gumbel1954,
  author    = {E. J. Gumbel},
  title     = {Statistical Theory of Extreme Values and Some Practical Applications: A Series of Lectures},
  year      = {1954},
  publisher = {U.S. Department of Commerce, National Bureau of Standards},
  series    = {Applied Mathematics Series},
  volume    = {33},
  address   = {Washington, D.C.}
}

@incollection{johnson1984extensions,
  title={Extensions of {L}ipschitz mappings into a {H}ilbert space},
  author={Johnson, William B and Lindenstrauss, Joram},
  booktitle={Contemporary Mathematics},
  volume={26},
  pages={189--206},
  year={1984},
  publisher={American Mathematical Society}
}

@misc{vllm_gumbel_sampler,
  title = {{vLLM} Gumbel-Max Sampler Implementation},
  author = {{vLLM Contributors}},
  howpublished = {\url{https://github.com/vllm-project/vllm/blob/10d765482d19abfab6c66b5f815720a66aa9de42/vllm/v1/sample/ops/topk_topp_sampler.py#L252}},
  year = {2024},
  note = {Line 252}
}

@article{he2025nondeterminism,
  author = {Horace He and Thinking Machines Lab},
  title = {Defeating Nondeterminism in {LLM} Inference},
  journal = {Thinking Machines Lab: Connectionism},
  year = {2025},
  note = {\url{https://thinkingmachines.ai/blog/defeating-nondeterminism-in-llm-inference/}},
  doi = {10.64434/tml.20250910}
}

@misc{moonshot2025k2verifier,
  author = {{Moonshot AI}},
  title = {K2 Vendor Verifier},
  year = {2025},
  howpublished = {\url{https://github.com/MoonshotAI/K2-Vendor-Verfier}},
  note = {Accessed: 2025-10-04}
}

@misc{anthropic2025postmortem,
  author = {{Anthropic}},
  title = {A Postmortem of Three Recent Issues},
  year = {2025},
  howpublished = {\url{https://www.anthropic.com/engineering/a-postmortem-of-three-recent-issues}},
  note = {Accessed: 2025-10-04}
}

@misc{chen2025zktorch,
      title={zkTorch: Privacy-Preserving Deep Learning with Zero-Knowledge Proofs},
      author={Boyuan Chen and Muqing Zheng and Tianyi Liu and Ruzica Piskac and Zhong Shao},
      year={2025},
      eprint={2502.00226},
      archivePrefix={arXiv},
      primaryClass={cs.CR},
      url={https://arxiv.org/abs/2502.00226},
}

@misc{sun2024zkllm,
      title={zkLLM: Zero Knowledge Proofs for Large Language Models},
      author={Haochen Sun and Jason Li and Hongyang Zhang},
      year={2024},
      eprint={2404.16109},
      archivePrefix={arXiv},
      primaryClass={cs.CR},
      url={https://arxiv.org/abs/2404.16109},
}

@misc{kwon2023vllm,
      title={Efficient Memory Management for Large Language Model Serving with PagedAttention},
      author={Woosuk Kwon and Zhuohan Li and Siyuan Zhuang and Ying Sheng and Lianmin Zheng and Cody Hao Yu and Joseph Gonzalez and Hao Zhang and Ion Stoica},
      year={2023},
      eprint={2309.06180},
      archivePrefix={arXiv},
      primaryClass={cs.LG},
      url={https://arxiv.org/abs/2309.06180},
}

@inproceedings{grattafiori2024llama3,
      title={The Llama 3 Herd of Models},
      author={Aaron Grattafiori and Abhimanyu Dubey and Abhinav Jauhri and Abhinav Pandey and others},
      year={2024},
      url={https://arxiv.org/abs/2407.21783},
}

@misc{schmalbach2024temperature,
  title={Does Temperature 0 Guarantee Deterministic LLM Outputs?},
  author={Schmalbach, Vincent},
  year={2024},
  howpublished={\url{https://www.vincentschmalbach.com/does-temperature-0-guarantee-deterministic-llm-outputs/}},
  note={Accessed: 2025-10-06}
}

@misc{vieira2014gumbel,
  title={The Gumbel-Max Trick},
  author={Vieira, Tim},
  year={2014},
  howpublished={\url{https://timvieira.github.io/blog/post/2014/07/31/gumbel-max-trick/}},
  note={Accessed: 2025-10-06}
}

@misc{ultrachat_200k,
  title={UltraChat 200k},
  author={HuggingFace H4},
  year={2023},
  howpublished={\url{https://huggingface.co/datasets/HuggingFaceH4/ultrachat_200k}},
  note={Filtered subset of 1.4M dialogue dataset}
}

@article{yang2025qwen3,
  title={Qwen3 Technical Report},
  author={Yang, An and others},
  journal={arXiv preprint arXiv:2505.09388},
  year={2025}
}

@article{huijben2021review,
  author = {Huijben, Ivo and Kool, Wouter and Mooij, Joris M. and de Vries, Bert and van Sloun, Wouter J. J.},
  title = {A Review of the Gumbel-Max Trick and Its Extensions for Discrete Stochasticity in Machine Learning},
  journal = {arXiv preprint arXiv:2110.01515},
  year = {2021}
}

@article{puigcerver2024sparse,
  title={From Sparse to Soft Mixtures of Experts},
  author={Puigcerver, Joan and Riquelme, Carlos and Mustafa, Basil and Houlsby, Neil},
  journal={arXiv preprint arXiv:2308.00951},
  year={2023},
  eprint={2308.00951},
  archivePrefix={arXiv},
  url={https://arxiv.org/abs/2308.00951}
}

@misc{chen2023acceleratinglargelanguagemodel,
      title={Accelerating Large Language Model Decoding with Speculative Sampling}, 
      author={Charlie Chen and Sebastian Borgeaud and Geoffrey Irving and Jean-Baptiste Lespiau and Laurent Sifre and John Jumper},
      year={2023},
      eprint={2302.01318},
      archivePrefix={arXiv},
      primaryClass={cs.CL},
      url={https://arxiv.org/abs/2302.01318}, 
}

@misc{leviathan2023fastinferencetransformersspeculative,
      title={Fast Inference from Transformers via Speculative Decoding}, 
      author={Yaniv Leviathan and Matan Kalman and Yossi Matias},
      year={2023},
      eprint={2211.17192},
      archivePrefix={arXiv},
      primaryClass={cs.LG},
      url={https://arxiv.org/abs/2211.17192}, 
}

@misc{li2025eagle3scalinginferenceacceleration,
      title={EAGLE-3: Scaling up Inference Acceleration of Large Language Models via Training-Time Test}, 
      author={Yuhui Li and Fangyun Wei and Chao Zhang and Hongyang Zhang},
      year={2025},
      eprint={2503.01840},
      archivePrefix={arXiv},
      primaryClass={cs.CL},
      url={https://arxiv.org/abs/2503.01840}, 
}

@misc{zhang2025learningharmonizedrepresentationsspeculative,
      title={Learning Harmonized Representations for Speculative Sampling}, 
      author={Lefan Zhang and Xiaodan Wang and Yanhua Huang and Ruiwen Xu},
      year={2025},
      eprint={2408.15766},
      archivePrefix={arXiv},
      primaryClass={cs.LG},
      url={https://arxiv.org/abs/2408.15766}, 
}

@misc{singh2025logic,
      title={{LOGIC: Trustless Inference through Log-Probability Verification}},
      author={Amarjot Singh and Francesco Virga and Sean Smith and Sam Hogan},
      year={2025},
      howpublished={\url{https://inference.net/blog/logic}},
      note={Context Labs Blog, November 5, 2025}
}

@misc{erdil2025inferenceeconomicslanguagemodels,
      title={Inference economics of language models}, 
      author={Ege Erdil},
      year={2025},
      eprint={2506.04645},
      archivePrefix={arXiv},
      primaryClass={cs.LG},
      url={https://arxiv.org/abs/2506.04645}, 
}

@misc{gao2025modelequalitytestingmodel,
      title={Model Equality Testing: Which Model Is This API Serving?}, 
      author={Irena Gao and Percy Liang and Carlos Guestrin},
      year={2025},
      eprint={2410.20247},
      archivePrefix={arXiv},
      primaryClass={cs.LG},
      url={https://arxiv.org/abs/2410.20247}, 
}
